\documentclass{article}

\usepackage{graphicx}
\usepackage{algorithm}
\usepackage{algpseudocode}
\usepackage{amsmath} 
\usepackage{booktabs}
\usepackage{multirow}
\usepackage{amsmath}
\usepackage{amssymb}
\usepackage{booktabs}
\usepackage{tabularx}

\newcommand{\RM}{SARM2}
\usepackage[preprint]{corl_2026} 

\title{SARM2: Multi-Task Stage Aware Reward Modeling for Self Improving Robotic Manipulation}

\vspace{-2.5em}

\author{
  Qianzhong Chen$^{1}$ \hspace{0.12em}
  Hau Zheng$^{1}$ \hspace{0.12em}
  Justin Yu$^{2}$ \hspace{0.12em}
  Suning Huang$^{1}$ \hspace{0.12em}
  Jiankai Sun$^{1}$ \hspace{0.12em} \\[1ex]
  \textbf{Ken Goldberg$^{2}$ \hspace{0.12em}
  Chuan Wen$^{3}$ \hspace{0.12em}
  Pieter Abbeel$^{2}$ \hspace{0.12em}
  Yide Shentu$^{2, 4}$ \hspace{0.12em}
  Philipp Wu$^{4}$ \hspace{0.12em}
  Mac Schwager$^{1}$ \hspace{0.12em}} \\[1ex]
  $^{1}$Stanford University, $^{2}$UC Berkeley, $^{3}$Shanghai Jiao Tong University $^{4}$xdof.ai \\
}

\begin{document}
\maketitle

\makeatletter
\renewcommand{\@makefntext}[1]{\noindent #1}
\makeatother
\footnotetext{For any questions, please contact: \texttt{qchen23@stanford.edu}}


\vspace{-2.5em}

\begin{figure}[h!]
    \centering
    \includegraphics[width=\linewidth]{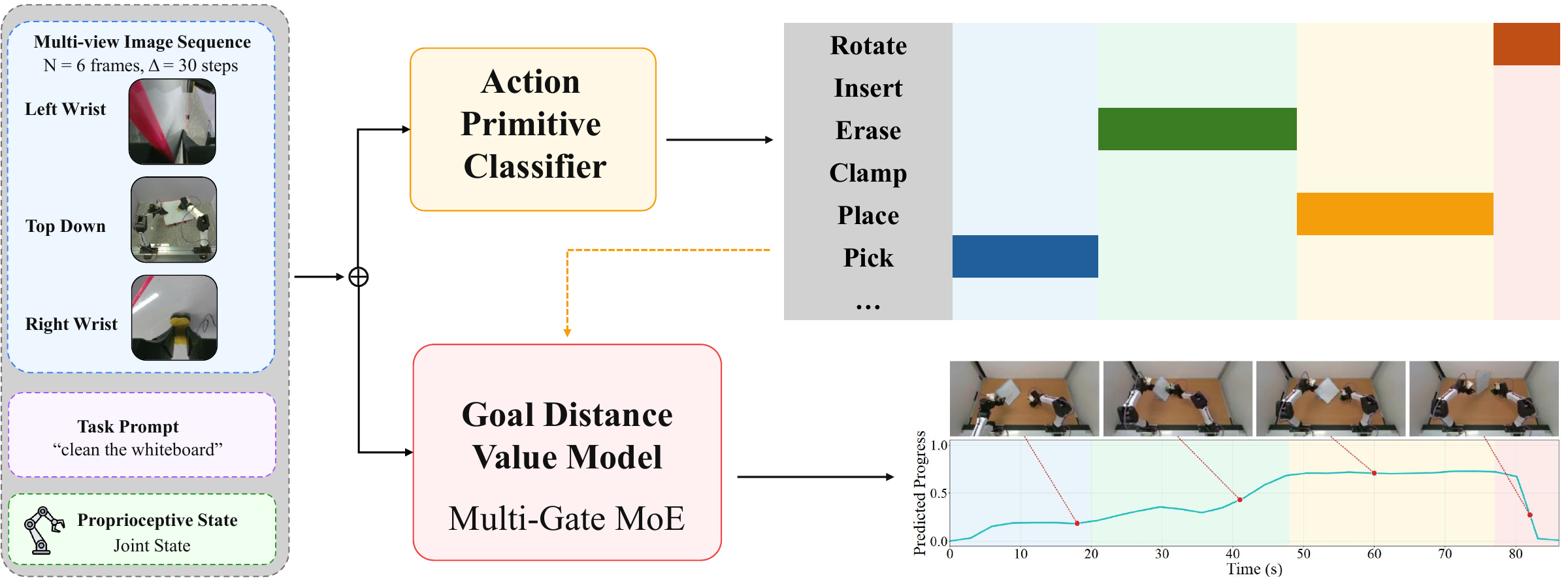}
    \vspace{-1.5em}
    \caption{\textbf{Overview of \RM.} \RM{} achieves multi-task stage aware reward modeling by leveraging a general stage estimator, which classifies the current segment over $K{+}1{=}22$ action primitives. The stage information is used by a downstream multi-gate Mixture of Experts (MMoE) value head, achieving dense, accurate, and general value estimation for manipulation tasks.}
    \label{fig:hero}
    \vspace{-1.0em}
\end{figure}

\begin{abstract}
Fine-tuning vision-language-action (VLA) policies for long-horizon manipulation still relies heavily on behavior cloning, which requires costly high-quality demonstrations and keeps policies near the demonstration distribution. Reward models can reduce this dependence by reweighting demonstrations and providing dense supervision for on-robot reinforcement learning (RL), but they must be dense, accurate, and general. Existing methods fall short: task-specific stage-aware models are accurate but require per-task annotations, while general vision-language-model (VLM) reward models are broadly applicable but too coarse for fine-grained long-horizon progress. We introduce \textbf{\RM}, a multi-task stage-aware reward model that combines an action-primitive-based stage estimator with a multi-gate Mixture-of-Experts (MMoE) value head to produce dense per-step rewards across manipulation tasks. Building on \RM, we further propose \textbf{SPIRAL} (Self-Policy Improvement via Reward-Aligned Learning), an on-policy reward-guided framework that improves VLA policies from cheap autonomous rollouts. On a 10-task benchmark, \RM{} reduces value-estimation MSE by $80\%$ over the strongest baselines; when used in SPIRAL, it improves task success from around $50\%$ to near-perfect performance on Folding Shorts ($58\% \to 100\%$) and Cleaning Whiteboard ($50\% \to 90\%$), showing that high-quality dense rewards are key to a stable robot data flywheel. Project website: \url{https://qianzhong-chen.github.io/sarm2.github.io/}.
\end{abstract}

\vspace{-1.5em}


\section{Introduction}

\vspace{-1.0em}

Vision-Language-Action (VLA) models \citep{zitkovich2023rt, kim2024openvla, team2024octo, black2024pi_0, intelligence2025pi_} have established end-to-end policy learning as a dominant paradigm in robotic manipulation. Yet most VLAs struggle beyond short-horizon tasks out of the box; extending them to long-horizon tasks typically requires fine-tuning on in-domain data, and supervised fine-tuning (SFT) demands large volumes of human demonstrations and remains prone to out-of-distribution (OOD) failures.

Recent methods leverage reward models to filter training samples and extract more value from high-quality data~\citep{chen2025sarm}, but these offline approaches are bounded by the demonstration distribution and cannot self-improve from rollouts. Methods like $\pi^{*}_{0.6}$~\citep{intelligence2025pi} and ConRFT~\citep{chen2025conrft} continually refine the policy via on-policy DAgger, but require costly human-in-the-loop supervision from operators proficient in both teleoperation and policy training. A separate line of work~\citep{guo2025improving, sun2026prior, lei2025rl} uses online interaction data with sparse rewards, falling short on long-horizon tasks. Across these directions, scaling VLA-RL relies on an accurate reward model that can faithfully judge whether each segment of a long-horizon task is executed correctly.

Existing reward models for manipulation fall into two groups: task-specific reward models and general-purpose ones. Early task-specific designs~\citep{ma2023liv, alakuijala2024video, hung2024victor, kim2025subtask, zhang2025rewind} are typically trained on a narrow set of demonstrations and struggle on long-horizon, OOD conditions, where subtle stage transitions and compounding errors are difficult to capture; SARM~\citep{chen2025sarm} adds a stage-conditioned hierarchy that improves accuracy on long-horizon tasks, but it remains single-task and relies on task-specific stage annotations, which limits its applicability as task diversity grows. General-purpose VLM-based reward models~\citep{ma2024vision, tan2025robo, chen2026topreward, lee2026roboreward, liang2026robometer} instead leverage the broad priors of pretrained VLMs and generalize across tasks, yet their predictions tend to be coarse-grained and noisy at the step level, making them ill-suited as a dense supervisory signal for long-horizon manipulation. To support VLA-RL at scale, a reward model must satisfy three requirements simultaneously, being \emph{dense}, \emph{accurate}, and \emph{general} enough to supervise long-horizon tasks across domains, yet none of the existing reward models meets all three.

We propose \textbf{\RM}, a multi-task stage-aware reward model pairing a general stage estimator with a multi-gate Mixture-of-Experts (MMoE) value head. The stage estimator transfers across tasks through a shared action-primitive vocabulary, and its predicted primitive selects the corresponding MMoE gate, guiding the value head to activate the most relevant domain- and action-specific experts for accurate dense value estimation. We further introduce \textbf{SPIRAL} (Self-Policy Improvement via Reward-Aligned Learning), an on-policy real-robot RL framework that leverges \RM's dense rewards and cheap autonomous rollouts into a self-improving data flywheel. Our contributions are:

\vspace{-0.8em}

\begin{itemize}
    \item \textbf{\RM}, a multi-task stage-aware reward model that tightly couples a generalizable action-primitive stage estimator with a MMoE value head: the predicted primitive directly selects the corresponding MMoE gate, producing accurate dense rewards across diverse long-horizon tasks within one model.
    \item \textbf{SPIRAL}, a reward-aligned on-policy residual RL framework that leverages \RM's dense rewards. SPIRAL closes an  autonomous robot data flywheel, enabling continual policy refinement on long-horizon tasks with minimal human intervention.
    \item We validate our method on two real-world long-horizon manipulation tasks, \textit{Folding Shorts} and \textit{Cleaning Whiteboard}: plugged into SPIRAL, \RM{} outperforms sparse-reward and large VLM-based reward-model baselines, driving substantial gains in downstream policy success.
\end{itemize}


\vspace{-1em}

\section{Related Works}
\label{sec:related_works}

\vspace{-0.8em}

\paragraph{Reward Models for Robotic Manipulation.} Task-specific reward models learn from visual observations using mid-sized vision-language encoders such as CLIP~\citep{radford2021learning} or SigLIP~\citep{tschannen2025siglip}~\citep{ma2023liv, alakuijala2024video, hung2024victor, kim2025subtask, zhang2025rewind, intelligence2025pi}. To handle long-horizon, contact-rich tasks, \textit{stage-aware} variants~\citep{mu2024drs, kim2025subtask, chen2025sarm} decompose a task into sub-stages and produce stage-conditioned dense rewards. These methods are fundamentally single-task: each new task requires retraining, and SARM~\citep{chen2025sarm} further requires dense per-frame stage annotations. We compare against ReWiND~\citep{zhang2025rewind}, the strongest single-task baseline without dense annotation. A complementary line builds task-agnostic reward models on top of pretrained VLMs~\citep{ma2024vision, lee2026roboreward, tan2025robo, chen2026topreward, liang2026robometer}; their out-of-the-box rewards are too coarse for dense per-step signals, and fine-tuning their large parameter count is expensive. We benchmark against TOPReward~\citep{chen2026topreward} and Robometer~\citep{liang2026robometer} in Section~\ref{sec:experiment_rm}. \textbf{\RM} combines both paradigms: a reusable stage estimator built on a shared action-primitive vocabulary, paired with a multi-task value head that adapts to multiple tasks in a single model.

\vspace{-0.8em}

\paragraph{Real-Robot Reinforcement Learning.} RL on real-robot data is hindered by two challenges: obtaining dense rewards on long-horizon tasks, and backpropagation through iterative diffusion/flow sampling. End-to-end RL trains compact single-task policies from scratch~\citep{huang2024mentor, luo2024serl, luo2025precise, zhao2025real, kalashnikov2018scalable, seo2024continuous, wu2022daydreamer}, sacrificing expressiveness. Offline-to-online methods initialize via BC then refine with RL: simple MLP heads~\citep{hu2023imitation, yang2024robot, chen2025conrft} are easy to fine-tune but limited; AWR-style updates on expressive policies~\citep{wu2025robocopilot, intelligence2025pi, peng2019advantage} depend on human DAgger interventions. RL100~\citep{lei2025rl} backpropagates PPO~\citep{schulman2017proximal} through denoising at high compute cost, while steering~\citep{wagenmaker2025steering, niu2024xted} and residual~\citep{xiao2025self, ankile2025residual} methods modify base-policy outputs without updating weights. Closest to ours is DICE-RL~\citep{sun2026prior}, which unifies value-guided action selection with residual learning. All of the above rely on \textit{sparse} rewards. MoE in robotics has been explored for policy capacity scaling~\citep{huang2024mentor, hao2026abstracting, cheng2025moe, zhang2026language, shen2025expertise, du2025himoe, yang2025drivemoe, huang2025moe, seyde2022strength, chen2025grad}, but to our knowledge \textbf{\RM} is the first to apply MoE to a multi-task robotics reward model.

\vspace{-1em}

\section{Method}
\label{sec:method}

\vspace{-0.8em}

We present our method in two parts: Section~\ref{sec:rm} introduces \RM, our multi-task stage-aware reward model that produces dense progress estimates across long-horizon tasks, and Section~\ref{sec:spiral} introduces SPIRAL, a self-improvement framework that turns \RM's dense rewards into an autonomous on-policy data flywheel.

\vspace{-0.8em}

\subsection{\RM: Multi-Task Stage-Aware Reward Model}
\label{sec:rm}

\begin{figure} [h!]
    \centering
    \includegraphics[width=0.9\linewidth]{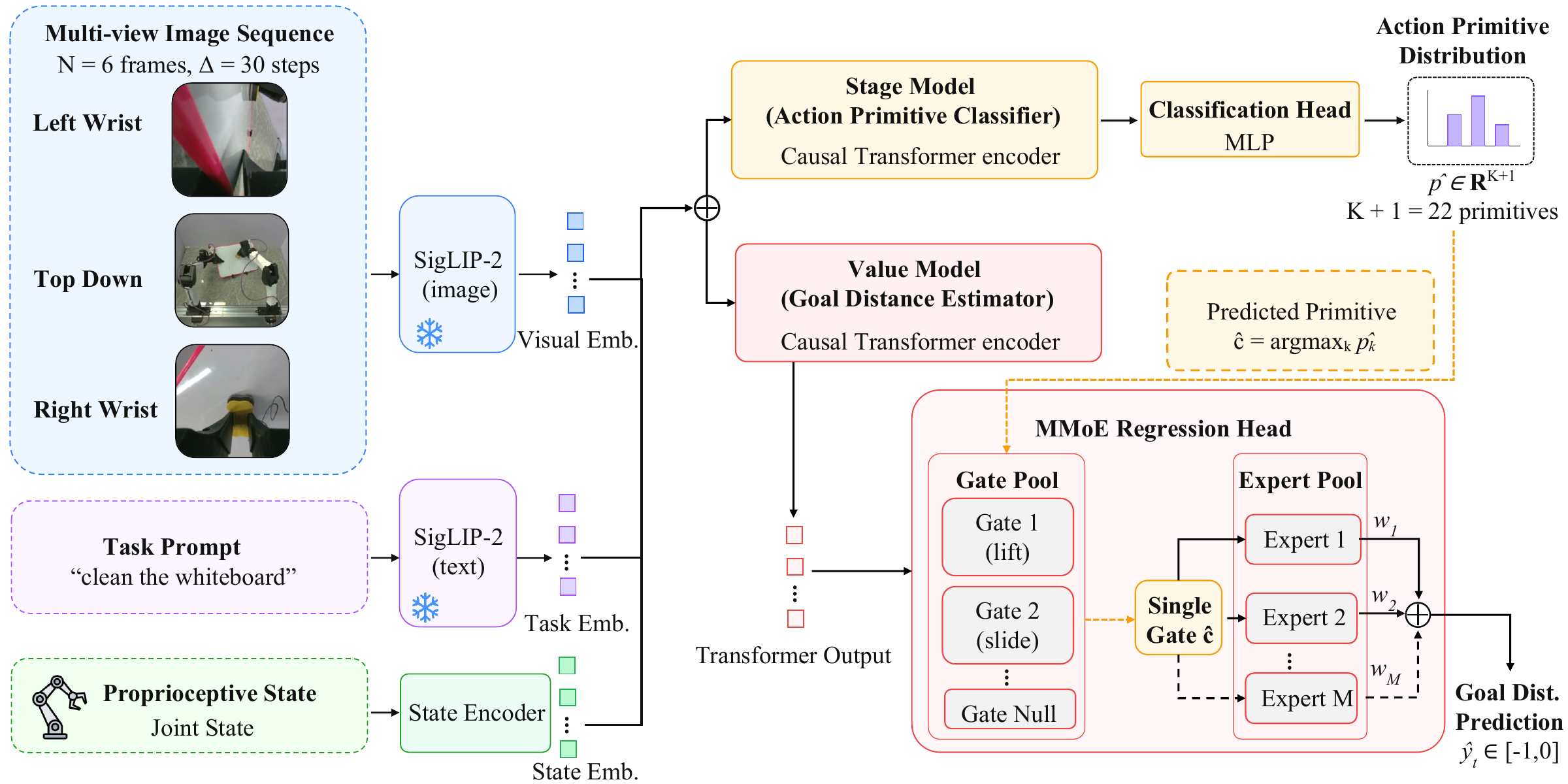}
    \vspace{-1.0em}
    \caption{\textbf{Overview of \RM.} Three camera views plus proprioceptive state are encoded by a shared frozen SigLIP-2 backbone, whose cached frame embeddings feed two \emph{separately trained} causal Transformers: (i) a task-agnostic \emph{stage estimator} that classifies the current segment over $K{+}1{=}22$ candidates ($K{=}21$ action primitives and a null class used as a fallback when the model is uncertain), and (ii) a \emph{multi-gate MoE value decoder} whose gate is selected by the predicted primitive group and routes the fused token through top-$k$ shared experts to produce a dense progress estimate.}
    \label{fig:\RM}
    \vspace{-1.5em}
\end{figure}


\subsubsection{Action-Primitive Stage Estimator}
\label{sec:moe_reward_model}

SARM~\citep{chen2025sarm} showed that stage-aware reward modeling enables accurate progress estimation on long-horizon tasks, but its stage estimator depends on task-specific annotations and must be retrained for each new task. We instead use \textit{action primitives} as a task-agnostic intermediate representation. Two observations from an internal corpus of $10{,}000+$ hours of annotated manipulation data motivate this design: (1) demonstrations across nearly all tasks decompose into sequences of action primitives that recur across tasks, despite differences in task names, scenes, and objects; (2) the primitive vocabulary is comparatively small, and many primitives are either semantically similar (e.g., \texttt{pour}/\texttt{dump}) or form dual pairs rarely co-occurring within a task (e.g., \texttt{pull}/\texttt{push}), allowing further compression.

\vspace{-0.8em}

\paragraph{Action Primitive Dataset Construction.}
We curated 200 hours of real-world manipulation data spanning 100 tasks. After consolidation we identified $K = 21$ action primitives that cover $>90\%$ of total task duration. We partitioned the source dataset along annotation boundaries and constructed a balanced dataset $\mathcal{D}_{\text{AP}}$ with (i) each primitive contributing $t_k = 3$ hours, and (ii) maximum source-task diversity per primitive, yielding $(K + 1) \cdot t_k = 66$ hours of segments. The additional null class is formed from all remaining long-tail samples, including reset behaviors and action primitives with low occurrence frequency. Construction details are in Appendix~\ref{sec:action_primitive_groups}.

\vspace{-0.8em}

\paragraph{Model Architecture.}
We frame stage estimation as temporal classification over the primitive vocabulary plus a null class ($K + 1 = 22$ classes). At time $t$, the model receives $N = 6$ recent frames sampled at stride $\Delta = 30$ ($\approx 6$\,s of context at 30\,Hz). We use three views $v \in \{\text{wrist}_L, \text{wrist}_R, \text{top}\}$, denoted $\mathbf{I}_{t}^{(v)}$, plus proprioceptive state $\mathbf{s}_t$. Each frame is encoded by a frozen SigLIP-2~\citep{tschannen2025siglip} encoder $\phi$; $\mathbf{s}_t$ is projected by an MLP $\psi$. All tokens are processed by a 4-layer causal Transformer $f_\theta$ followed by a linear head $g_\theta$ producing class logits $\mathbf{p}_t = \mathrm{softmax}(g_\theta(\mathbf{h}_t))$. The model is trained with cross-entropy on $\mathcal{D}_{\text{AP}}$; full details on hyperparameter are in Appendix~\ref{appendix:training}.


\vspace{-0.5em}

\subsubsection{MMoE Value Decoder}
\label{sec:stage_estimation}

\vspace{-0.5em}

Building on the stage estimator, we design a multi-task value model that estimates dense, stage-conditioned goal-distance. Different primitives exhibit markedly different visual dynamics and progress signatures, so a monolithic value head suffers from interference and tail underfitting. We address this with a \textit{multi-gate Mixture-of-Experts (MMoE)} decoder that specializes along primitive groups while sharing a common expert pool.

The value model uses its own 6-layer causal Transformer backbone, trained separately from the stage estimator's 4-layer Transformer; the two share only the frozen SigLIP-2 encoder and its cached six-frame embeddings, which eliminates redundant visual encoding at inference. At step $t$, the value model receives: (1) three-camera frames; (2) proprioceptive state $\mathbf{s}_t$; (3) task-name text embedding $\mathbf{e}_{\text{task}}$; and (4) the predicted-primitive embedding $\mathbf{e}_{\text{prim}}$ of $\tilde{y}_t$ from the stage estimator. Following SARM~\citep{chen2025sarm}, the input frame sequence concatenates: the first episode frame (visual anchor), the six recent frames at stride $\Delta=30$, and up to three rewinding frames as implicit negatives~\citep{zhang2025rewind} that prevent the model from collapsing to a monotonic time-index predictor.

We partition the $K+1$ primitives into $M+1$ semantic groups by shared visual/motion patterns (see Appendix~\ref{sec:action_primitive_groups}). Following MMoE~\citep{ma2018modeling, lei2024m3}, each group has a dedicated gate $G_m$ over a shared pool of $E$ experts (each a 3-layer MLP). The active gate at step $t$ is selected by $m(\tilde{y}_t)$, producing top-$k$ routing weights and MoE output

\vspace{-1.5em}

\begin{equation}
    \mathbf{o}_t = \sum_{e=1}^{E} g_{t,e}^{(m(\tilde{y}_t))} \cdot \mathcal{E}_e(\mathbf{h}_t), \quad \mathbf{g}_t^{(m)} = \mathrm{TopK}(\mathrm{softmax}(W_m \mathbf{h}_t)).
\end{equation}
Group-conditioned routing provides an explicit inductive bias: primitives from different groups use separate gates, while shared experts remain accessible. To prevent router collapse, we add per-gate balance and entropy auxiliary losses ($\mathcal{L}_{\text{balance}}$, $\mathcal{L}_{\text{entropy}}$); full formulae and ablations are in Appendix~\ref{sec:moe_dense_budget} and~\ref{sec:decoder_ffn}, full details on hyperparameter are in Appendix~\ref{appendix:training}.

Following $\pi^{*}_{0.6}$~\cite{intelligence2025pi}, the value model predicts normalized \textit{remaining steps to completion}, $r_t^{\star} = -(T-t)/T \in [-1, 0]$, which we find easier to fine-tune on rollouts with variable horizons. The full objective combines an MSE term with the auxiliary losses:
\begin{equation}
    \mathcal{L} = \mathbb{E}_{\mathcal{D}}[(\hat{r}_t - r_t^{\star})^2] + \lambda_{\text{bal}} \mathcal{L}_{\text{balance}} + \lambda_{\text{ent}} \mathcal{L}_{\text{entropy}}.
\end{equation}

\vspace{-0.8em}

\subsection{SPIRAL: Dense-Reward-Enabled Robot Self-Improvement Framework}
\label{sec:spiral}

\begin{figure}
    \centering
    \includegraphics[width=\linewidth]{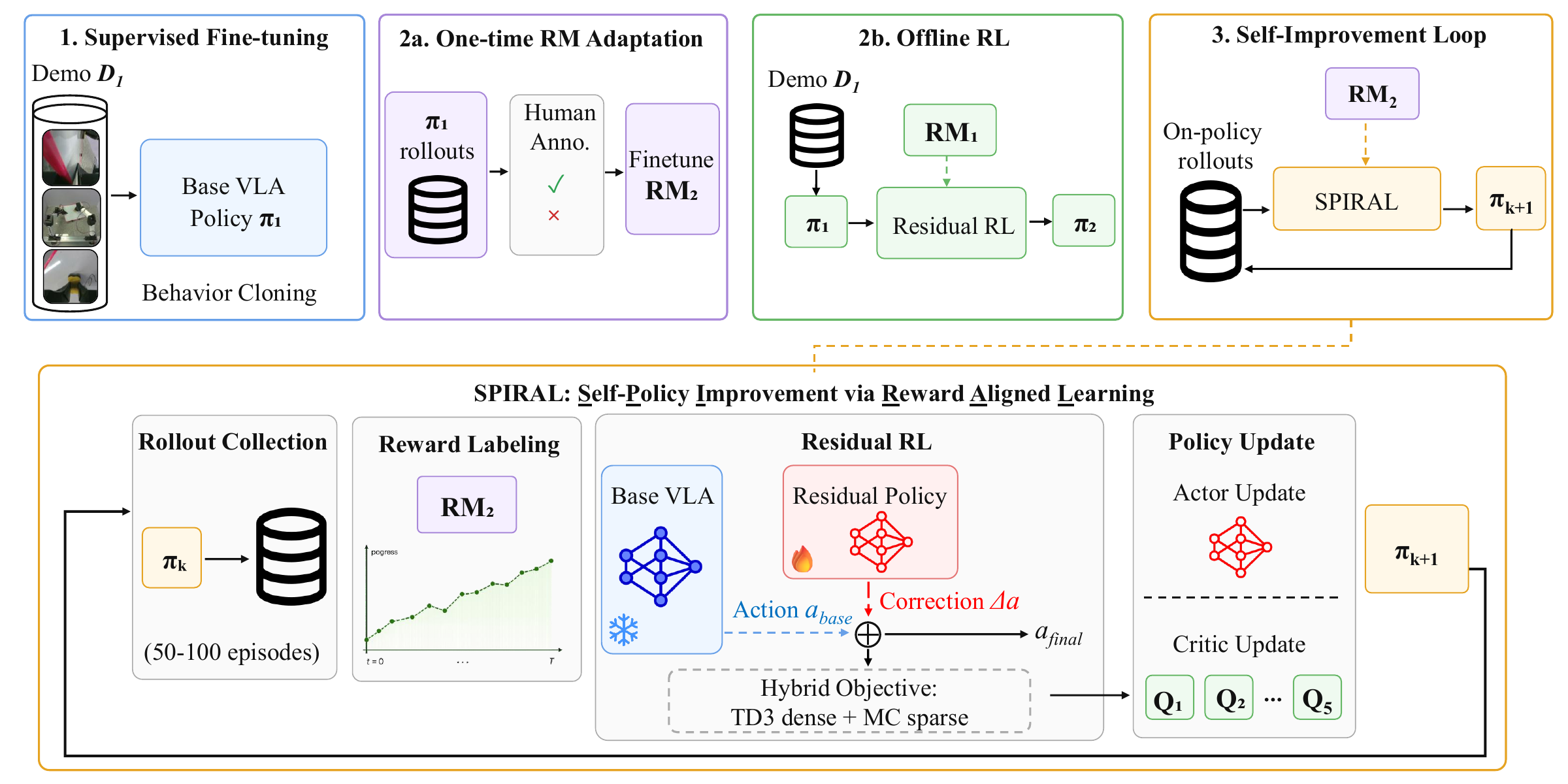}
    \vspace{-1.5em}
    \caption{\textbf{SPIRAL: \RM-powered self-improvement framework.} (1) BC fine-tunes $\pi_{\text{VLA}}$ on demos to obtain $\pi_1$. (2) In parallel, (2a) a one-time human annotation of $\sim$100 rollouts from $\pi_1$ adapts $\mathrm{RM}_1 \to \mathrm{RM}_2$ to cover the rollout distribution, while (2b) an offline SPIRAL update with the pretrained $\mathrm{RM}_1$ trains $\pi_2$. (3) An autonomous loop then alternates rollout collection, $\mathrm{RM}_2$ relabeling, and SPIRAL updates with no further supervision.}
    \label{fig:real_rl_framework}
    \vspace{-1.2em}
\end{figure}

We integrate the reward model from Section~\ref{sec:moe_reward_model} into \textbf{SPIRAL} (\emph{Self-Policy Improvement via Reward-Aligned Learning}), a closed-loop framework in which a VLA policy is iteratively refined through on-policy rollouts and reward-guided residual RL. As a residual-RL substrate, SPIRAL adopts the distribution contractive RL finetuning scheme of DICE-RL~\citep{sun2026prior}, which samples multiple latent noises for the flow policy and residual policy to generate diverse actions from the same observation, inducing a controlled exploration space for stable, sample-efficient real-robot learning. Aligning this substrate with a dense, stage-aware reward and making it long-horizon-ready. However, there are two key modifications that define SPIRAL: (i) the TD3~\citep{fujimoto2018addressing} objective uses dense per-step rewards from our multi-task reward model rather than sparse terminal rewards, and (ii) we combine this dense-reward TD3 objective with a Monte Carlo (MC) objective driven by sparse episode-level rewards, with more details in~\ref{sec:rl_design_choice}. Together with the autonomous rollout-relabel-update loop described below, these modifications turn a sparse-reward residual-RL recipe into a self-improving robot data flywheel.

Following DICE-RL~\citep{sun2026prior}, we instantiate the critic as an ensemble of $N=5$ chunk-based $Q$-functions $\{Q_{\phi_n}\}_{n=1}^{N}$ to mitigate value overestimation, with target networks and target-policy smoothing as in standard TD3. The dense reward $r_t^{\text{dense}}$ is provided by our reward model, and the bootstrap action in $y_t^{\text{TD}}$ is generated by $\pi_{\theta'}(s_{t+1}) + \epsilon$ with $\epsilon \sim \mathrm{clip}(\mathcal{N}(0, \sigma^2), -c, c)$; the critic loss is summed over all $N$ ensemble members. The residual actor is then trained with a TD3+BC-style objective,

\vspace{-1.5em}

\begin{equation}
    \min_{\theta} \; \mathbb{E}_{\substack{s \sim \mathcal{D} \\ z \sim \mathcal{N}(0, I)}} \Big[ -Q_{\phi}(s, a) + \beta \, \| s_{\theta}(s, z) \|_2^2 \Big],
\end{equation}
where the BC regularizer with weight $\beta = 30$ keeps the residual close to the pretrained policy while $Q_{\phi}$ encourages improving edits. Following the multi-sample expectation training of~\citet{sun2026prior}, at each visited state we draw $\kappa$ latent samples $\{z_j\}_{j=1}^{\kappa} \sim \mathcal{N}(0, I)$ from the flow prior, form candidate action chunks $a^{(j)} = \pi_{\text{pre}}(s, z_j) + s_\theta(s, z_j)$, and average both the critic TD target and the actor objective over the $\kappa$ candidates rather than over a single draw; at deployment we apply best-of-$\kappa$ action selection, executing $a^{(j^\star)}$ with $j^\star = \arg\max_j Q_\phi(s, a^{(j)})$. This reuses each visited state across $\kappa$ latent draws and provides a low-variance signal aligned with the full latent-induced action distribution rather than a single sample.

The full SPIRAL pipeline alternates policy rollouts, reward model adaptation, and residual RL across four stages: (1) BC fine-tuning of the VLA backbone on demonstrations to obtain $\pi_1$; (2) an initial offline SPIRAL update with pretrained reward model $\mathrm{RM}_1$, producing $\pi_2$; (3) a \emph{one-time} reward model adaptation that human-annotates rollouts from the weakest policy $\pi_1$ and fine-tunes $\mathrm{RM}_1 \to \mathrm{RM}_2$ to cover the rollout distribution; (4) an autonomous on-policy loop that iteratively labels new rollouts with $\mathrm{RM}_2$ and refines the policy through SPIRAL updates, requiring no further human labels. See Figure~\ref{fig:real_rl_framework}, Algorithm~\ref{alg:self_improvement}, Appendix~\ref{appendix:training} and Appendix~\ref{sec:rl_design_choice} for framework visualization, algorithm, detailed hyperparameter, design choices and practical considerations.

\vspace{-1.0em}

\section{Experimental Results}
\label{sec:result}

\vspace{-0.8em}

We empirically validate \textbf{\RM} and its self-improvement framework around three questions: \textbf{Q1.} Do the action-primitive stage estimator and MMoE improve value modeling accuracy on long-horizon and compositional tasks? \textbf{Q2.} Does \textbf{SPIRAL} make the policy improve through rollouts and surpass BC and offline RL baselines? \textbf{Q3.} Why is reward-model quality critical for the manipulation data flywheel?

\vspace{-0.8em}

\subsection{Reward Model Evaluation}
\label{sec:experiment_rm}

\vspace{-0.8em}

\paragraph{Task Suite, Baselines, and Ablations.}
We evaluate on 10 manipulation tasks split into two subsets: $\mathcal{S}_1$ (5 \textit{classic} tasks --- pick-and-place, cloth folding etc., which dominate general reward-model training data) and $\mathcal{S}_2$ (5 \textit{unconventional} tasks requiring tool use or multi-stage compositional execution); details in Appendix~\ref{sec:task_description}. Baselines: \textbf{ReWiND}~\citep{zhang2025rewind} (\textbf{RW}), trained from scratch on the same 10 tasks, no stage estimator and no MoE; \textbf{TOPReward}~\citep{chen2026topreward} (\textbf{TR}) and \textbf{Robometer}~\citep{liang2026robometer} (\textbf{RM}), general-purpose VLM-based reward models. For TOPReward, we use \textit{Qwen3-VL-8B-Instruct} variant. For Robometer we also test a LoRA fine-tuned variant (\textbf{RM-FT}) on our 10-task data for 2000 steps following the official recipe (full fine-tuning is impractical; LoRA alone needs $\sim 2\times$ the VRAM of our model). Ablations: \textbf{w/o SE} (single-gate MoE on ReWiND, no stage estimation); \textbf{w/o MG} (single-gate \RM{} removing the multi-gate inductive bias); $\mathcal{S}_1$ \textbf{Var.} (\RM{} trained only on $\mathcal{S}_1$).

\vspace{-0.8em}

\paragraph{Evaluation and Analysis.}
We evaluate on (i) held-out human demonstrations across all 10 tasks, and (ii) policy rollouts on the 2 tasks for which a downstream policy was trained, and we additionally report MoE routing diagnostics for every model with an MoE structure; the full protocol is in Appendix~\ref{sec:rm_protocol}. Table~\ref{tab:rm_result} reveals several consistent patterns. \textbf{\RM} achieves the lowest demonstration MSE on both task subsets and overall ($0.006$ on $\mathcal{S}_1$, $0.031$ on $\mathcal{S}_2$, $0.020$ combined), as well as the highest rollout classification score on the harder T2 ($\rho = 0.667$). We highlight four observations.

\vspace{-0.5em}

\textit{(1) General-purpose VLM-based reward models underperform on dense value estimation.} Despite their strong semantic generalization, TOPReward and Robometer yield demo losses $4$-$6\times$ higher than \textbf{\RM} on $\mathcal{S}_1$ and perform poorly on rollout classification, primarily due to overoptimistic estimation; see Figure~\ref{fig:rm_est} for visualizations. LoRA fine-tuning Robometer on our 10-task dataset substantially improves both demo loss and rollout score, bringing it close to from-scratch \textbf{ReWiND}, but despite being an order of magnitude larger ($\sim 4$B vs.\ $\sim 200$M parameters) it still underperforms \emph{every} from-scratch in-domain baseline. We attribute this to two factors: LoRA constrains the updatable parameter subspace and limits adaptation of fine-grained temporal patterns; and more fundamentally, VLM pretraining optimizes for coarse semantic alignment rather than the motion-level temporal discrimination that dense reward modeling demands.

\vspace{-0.5em}

\textit{(2) Stage estimation and MMoE are individually beneficial and synergistic.} Removing the stage estimator (\textbf{w/o SE}) raises demo loss from $0.020$ to $0.034$; removing the multi-gate design (\textbf{w/o MG}) raises it to $0.026$. The full \textbf{\RM} model improves substantially over both, showing that stage-awareness and multi-gate routing capture complementary structure: the stage estimator localizes the model in the task's primitive sequence, while the MMoE allocates capacity along primitive-group boundaries.

\vspace{-0.5em}

\textit{(3) Training on a broader task distribution improves performance.} The $\mathcal{S}_1$ \textbf{Var.} ablation, trained only on the simple subset, reaches $0.010$ demo loss on $\mathcal{S}_1$, which is better than every baseline yet still $1.7\times$ worse than full \textbf{\RM} ($0.006$), which sees both subsets during training. This indicates positive transfer from $\mathcal{S}_2$ back to $\mathcal{S}_1$, likely because $\mathcal{S}_2$ exposes the model to a richer set of action primitives and visual contexts that disambiguate progress signals shared with $\mathcal{S}_1$.

\vspace{-0.5em}

\textit{(4) MoE routing health correlates with downstream performance.} The \textbf{MoE Density} row reports the normalized top-$k$ routing score $\mathcal{S}_{\text{route}} \in [0,1]$ defined in Equation~\ref{eqn:moe_load_score}, where $0$ corresponds to a perfectly balanced router and $1$ to a fully collapsed one. \textbf{w/o SE} is closest to collapse ($\mathcal{S}_{\text{route}} = 0.87$): without primitive-conditioned gating, the router funnels most tokens through a small subset of experts. \textbf{w/o MG} is markedly healthier ($0.42$) but still concentrated, as a single shared gate cannot disentangle primitive groups. Both \RM{} variants are well-balanced ($0.23$ for $\mathcal{S}_1$\,Var.\ and $0.10$ for full \RM), and the ranking aligns with rollout score. This supports our claim that primitive-conditioned multi-gate routing directly mitigates router collapse.

\vspace{-0.5em}

\begin{table}[h!]
\small
\centering
\caption{\textbf{Reward-model evaluation on the 10-task benchmark.} \textbf{Demo $\mathcal{L}$}: per-frame MSE on held-out demos for the classic ($\mathcal{S}_1$) and unconventional ($\mathcal{S}_2$) subsets and frame-level micro-average. \textbf{Rollout $\rho \in [-1,1]$}, more details refer to Section~\ref{sec:rollout_eval} (T1: Folding Shorts, T2: Cleaning Whiteboard). \textbf{MoE Density}: normalized per-sample max expert utilization (values close to 1 indicate routing collapse, more details in Equation~\ref{eqn:moe_load_score}). Best per row in bold.}
\label{tab:rm_result}
\begin{tabular}{lcccccccc}
\toprule
\textbf{Metrics} & \textbf{RW} & \textbf{TR} & \textbf{RM} & \textbf{RM (FT)} & \textbf{w/o SE} & \textbf{w/o MG} & \textbf{$\mathcal{S}_1$ Var.} & \textbf{\RM} \\
\midrule
\multicolumn{9}{l}{\emph{Demo $\mathcal{L}$ $\downarrow$}} \\
$\mathcal{S}_1$         & 0.026 & 0.090 & 0.067 & 0.032 & 0.025 & 0.010 & 0.010 & \textbf{0.006} \\
$\mathcal{S}_2$         & 0.044 & 0.120 & 0.113 & 0.052 & 0.040 & 0.038 & N/A    & \textbf{0.031} \\
Overall         & 0.036 & 0.107 & 0.093 & 0.043 & 0.034 & 0.026 & N/A    & \textbf{0.020} \\
\midrule
\multicolumn{9}{l}{\emph{Rollout $\rho$ $\uparrow$}} \\
T1              & 0.167    & -0.222    & 0.222    & 0.667    & 0.556    & 0.611    & \textbf{0.889}    & 0.833 \\
T2              & 0.222 & -0.444 & -0.111 & 0.000 & 0.556 & 0.333 & N/A    & \textbf{0.667} \\
\midrule
\multicolumn{9}{l}{\emph{MoE Density}} \\
 & N/A & N/A & N/A & N/A & 0.87 & 0.42 & 0.23 & 0.10 \\
\bottomrule
\end{tabular}
\vspace{-1.2em}
\end{table}

\subsection{Policy Learning with Self Improvement}

\vspace{-0.8em}

\paragraph{Tasks and Evaluation Protocol.}
We select two representative tasks, one classic task from $\mathcal{S}_1$, one unconventional task from $\mathcal{S}_2$ to evaluate \textbf{\RM}-powered self-improvement; full descriptions in Appendix~\ref{sec:task_description}. \textit{Task 1 (classic): Folding shorts}~\citep{black2024pi_0, intelligence2025pi_, chen2025sarm, zheng2025x, wen2025dexvla, kooijmans2026_unfolding}, a long-horizon deformable-object manipulation task. We use 20 hours / 670 episodes (60--180\,s each, no quality filtering) and report success on two subtasks: \textbf{Flat} (2\,min limit) and \textbf{Crumpled} (4\,min limit, harder per~\citep{chen2025sarm}), 12 trials each across 3 shorts colors. \textit{Task 2 (unconventional): Cleaning Whiteboard}: grasp an eraser, stabilize a tilted whiteboard with the other hand, wipe all letters, and stop. Base policy trained on 10\,h / 530 episodes. We draw 5--10 letters covering 50--70\% of the board with a 2\,min limit and score each episode on five tiers (\textbf{0/25/50/75/100\%}) tracking grasping, partial cleaning, full cleaning, and correct termination, see Appendix~\ref{sec:wb_progress} for more details. We evaluate 4 configurations (2 eraser colors $\times$ 2 board frames) with 5 episodes each (20 total).

\vspace{-0.8em}

\paragraph{Baselines.}
Two groups by training-data source. \textit{Demo-only:} \textbf{BC} (vanilla BC policy fintuned from $\pi_{0.5}$~\citep{intelligence2025pi_} ), \textbf{RA-BC}~\citep{chen2025sarm} (weighted BC with \RM{} per-step rewards), \textbf{RL-Sparse} (DICE-RL~\citep{sun2026prior} with terminal reward $=1$ for demos), and \textbf{RL-Dense} (SPIRAL with \RM{} dense rewards). \textit{Rollout-based self-improvement} all start from the RL-Dense demo checkpoint and differ only in reward source: \textbf{Sparse} (human-recorded terminal success), \textbf{RM (FT)} (LoRA-fine-tuned Robometer dense labels. The model is fintuned both on 10-tasks dataset as well as labeled rollouts, exactly the same procedure as \RM{}), and \textbf{\RM{} (ours)} (our dense labels). We report after Round 3 of Algorithm~\ref{alg:self_improvement}; at each round the policy is initialized from the checkpoint that produced the rollouts used for training, ensuring a fair on-policy comparison across reward sources.


\vspace{-0.8em}

\begin{figure}[h!]
    \centering
    \includegraphics[width=\linewidth]{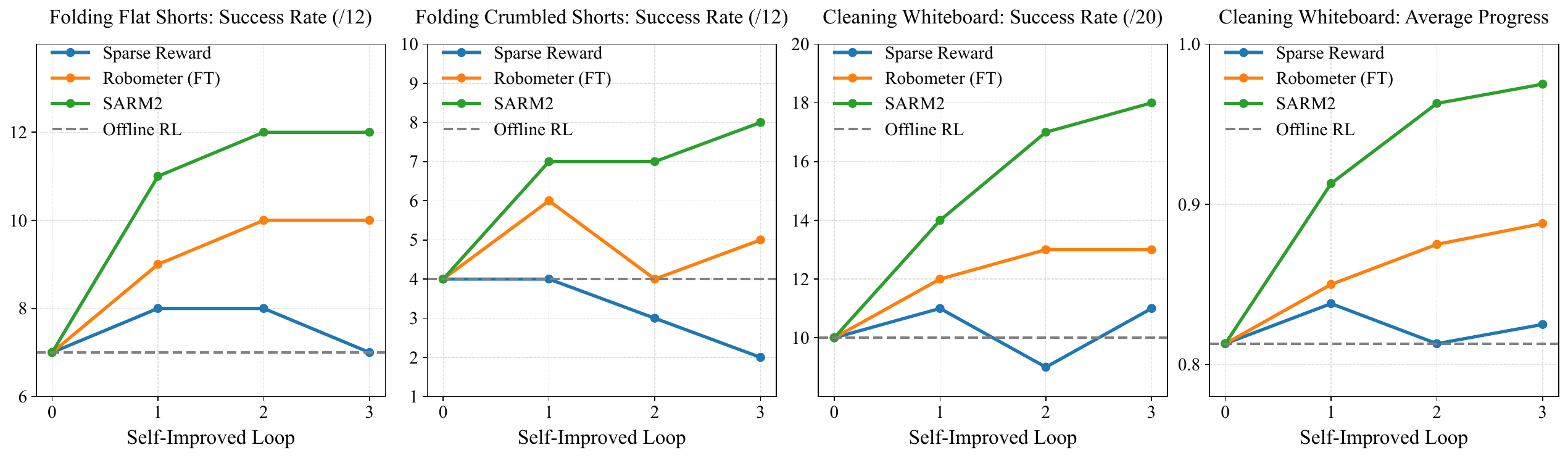}
    \vspace{-1.5em}
    \caption{\textbf{Self-improvement trends across three rounds of Algorithm~\ref{alg:self_improvement}.} Top: Folding Shorts (Flat and Crumpled SR). Bottom: Cleaning Whiteboard (SR and average five-tier progress). All curves start from the same RL-Dense checkpoint but differ in the rollout-labeling reward source and rollout episodes for later iterations. \RM{} improves monotonically on both tasks, RM (FT) plateaus, and Sparse regresses below the offline baseline on Folding Shorts Crumpled.}
    \label{fig:self_improve_trend}
    \vspace{-1.6em}
\end{figure}

\paragraph{Analysis.}
Table~\ref{tab:policy_results} and Figure~\ref{fig:self_improve_trend} reveal three findings.

\vspace{-0.5em}

\textit{(1) SPIRAL is an effective offline RL backbone.}
On demonstration-only training, RL-Dense reaches $7/12$ on Folding-Shorts Flat (vs.\ $1/12$ for BC, $4/12$ for RL-Sparse) and $10/20$ at $81.3\%$ progress on Cleaning Whiteboard (vs.\ $6/20$, $62.5\%$ for RL-Sparse). The gap over RA-BC is small on SR but RL-Dense generalizes better on harder subtasks and yields smoother motions in qualitative inspection, indicating that dense rewards combined with a value bootstrap give a stronger optimization signal than terminal sparse rewards or reward-weighted imitation.

\vspace{-0.5em}

\textit{(2) SPIRAL extends beyond offline baselines via reward-aligned self-improvement.}
All rollout-based methods start from the same RL-Dense checkpoint. After three rounds, SPIRAL with \textbf{\RM} dense rewards beats every offline baseline on every metric: $12/12$ Flat, $8/12$ Crumpled, $18/20$ at $97.5\%$ progress on Whiteboard. Figure~\ref{fig:self_improve_trend} shows the gains accumulate monotonically across rounds. This confirms that on-policy autonomous rollouts can improve the policy substantially above what offline RL alone delivers, validating SPIRAL as a real-robot self-improvement framework.

\vspace{-0.5em}

\textit{(3) The data flywheel is bottlenecked by reward-model quality.}
The hardest setting, Folding Shorts Crumpled, is diagnostic: training with sparse terminal rewards \emph{degrades} the RL-Dense checkpoint from $4/12$ to $2/12$, because a single endpoint signal cannot credit-assign across a $\sim\!3$\,min flattening stage and the loop reinforces flawed rollouts, rotating the flywheel backward. Robometer (FT) dense rewards do better but still trail \textbf{\RM} by wide margins ($10/12$ vs.\ $12/12$ on Flat; $5/12$ vs.\ $8/12$ on Crumpled; $13/20$ vs.\ $18/20$ on Whiteboard), showing that density alone is insufficient: for an effective robot self-improvement loop, the reward model must faithfully reflect the current rollout state, distinguishing true progress, corrective adjustment, and mistakes. Visualizations are provided in Figures~\ref{fig:rm_rollout_shorts} and~\ref{fig:rm_rollout_white}.

\begin{table}[h]
\small
\centering
\caption{\textbf{Policy evaluation after Round~3 of self-improvement.} \textit{Demo-only} methods train on demos; \textit{Rollout-based} methods all start from the RL-Dense checkpoint but differ in rollout episodes and reward source. \textbf{SR}: successes over total trials. \textbf{Avg.\ Prog.}: mean Cleaning Whiteboard score on the five-tier scale (Appendix~\ref{sec:wb_progress}). \textbf{Prog.\ Gain}: absolute progress over BC. Best per column in bold.}
\label{tab:policy_results}
\begin{tabular}{llcccccc}
\toprule
\multicolumn{2}{c}{\multirow{2}{*}{\textbf{Training Methods}}}
& \multicolumn{2}{c}{\textbf{Folding Shorts (SR$\uparrow$)}} 
& \multicolumn{3}{c}{\textbf{Cleaning Whiteboard}} \\
\cmidrule(lr){3-4} \cmidrule(lr){5-7}
\multicolumn{2}{c}{}
& \textbf{Flat} 
& \textbf{Crumble} 
& \textbf{SR$\uparrow$} 
& \textbf{Avg. Prog. $\uparrow$}
& \textbf{Prog. Gain $\uparrow$} \\
\midrule
\multirow{4}{*}{\textit{Demo only}}
& BC         & 1/12  & 0/12 & 4/20  & 0.475 & N/A \\
& RA-BC         & 6/12  & \textbf{5/12} & \textbf{10/20} & 0.775 & 0.300 \\
& RL-Sparse    & 4/12  & 2/12 & 6/20  & 0.625 & 0.150\\
& RL-Dense     & \textbf{7/12}  & 4/12 & \textbf{10/20} & \textbf{0.813} & \textbf{0.338} \\
\midrule
\multirow{3}{*}{\textit{Rollout based}}
& Sparse       & 7/12  & 2/12 & 11/20 & 0.825 & 0.350 \\
& RM (FT)      & 10/12 & 5/12 & 13/20 & 0.888 & 0.413\\
& \RM{} (ours) & \textbf{12/12} & \textbf{8/12} & \textbf{18/20} & \textbf{0.975} & \textbf{0.500} \\
\bottomrule
\end{tabular}
\vspace{-0.8em}
\end{table}

\vspace{-0.8em}

\section{Conclusion}
\label{sec:conclusion}

\vspace{-0.8em}

We presented \textbf{\RM}, a multi-task stage-aware reward model unifying a task-agnostic action-primitive stage estimator with a multi-gate MoE value decoder, producing dense, accurate rewards across long-horizon manipulation tasks within a single model. Built on top of \RM, \textbf{SPIRAL} turns those dense rewards into a reward-aligned on-policy self-improvement loop, enabling continual policy refinement from cheap autonomous rollouts. Across a 10-task benchmark, \RM{} outperforms task-specific and VLM-based baselines on demo MSE and rollout classification; ablations show stage-awareness, multi-gate routing, and training-distribution diversity each contribute. Plugged into SPIRAL, \RM{} drives substantial gains on Folding Shorts and Cleaning Whiteboard, demonstrating that reward quality is the load-bearing factor for a self-sustaining robot data flywheel.

\vspace{-0.8em}

\paragraph{Limitations.} Our approach has three main limitations. \textit{(1) Embodiment scope.} The action-primitive vocabulary is derived from a bimanual table-top dataset, so extending \RM{} to mobile manipulation or other embodiments requires rebuilding the primitive set and re-training the stage estimator. \textit{(2) Residual RL ceiling.} SPIRAL refines policies via residual RL rather than updating the VLA weights, which suffices for action-level and modality-wise correction but cannot fix errors at the intention or task-decomposition level, where mechanisms beyond residual RL (e.g., VLM fine-tuning) are needed. \textit{(3) Sample efficiency.} On-policy rollout collection remains relatively sample-inefficient and still requires a human operator for routine resets between rollouts.


\clearpage


\bibliography{SARM2}  

\clearpage
\section{Appendix}


\subsection{Hardware Setup}
For our real-world experiments, we use a bimanual tabletop robotic platform. The system consists of:

\begin{itemize}
  \item Two 6-DOF YAM robot arms manufactured by I2RT~\citep{i2rt_yam_arm}.
  \item Three RealSense D405 cameras: one mounted on each wrist and one fixed overhead camera for observing the workspace~\citep{d405_camera}.
\end{itemize}

Demonstration data is collected through a leader-follower GELLO teleoperation system~\citep{wu2024gello}. The environment is recorded at 30~fps and includes synchronized streams from three camera views: the left wrist, the right wrist, and the fixed top view. It also records robot joint angles and commanded joint-angle actions.

\begin{figure}[h!]
    \centering
    \includegraphics[width=0.6\linewidth]{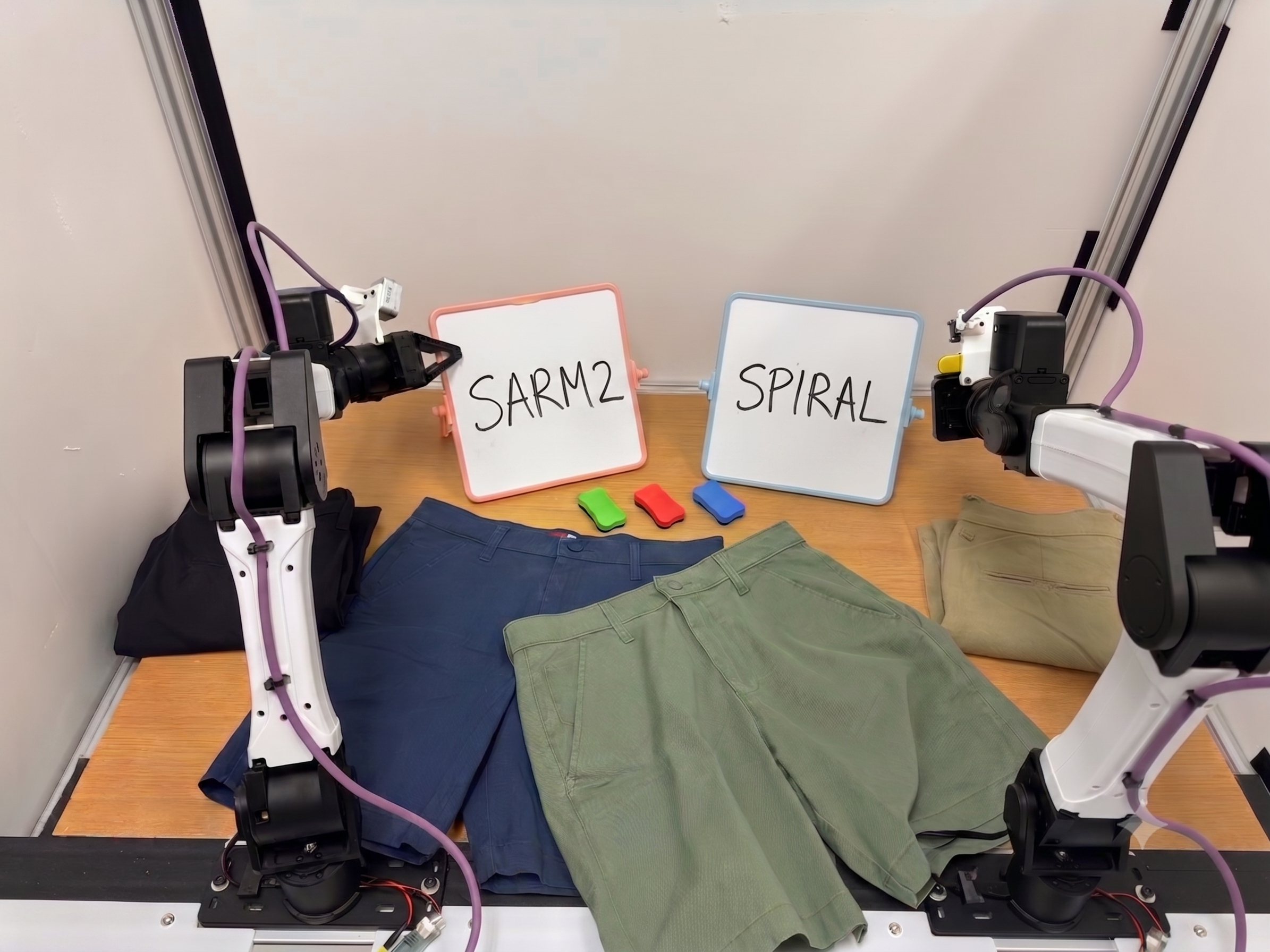}
    \caption{The physical station used for data collection and policy evaluation.}
    \label{fig:placeholder}
\end{figure}

\subsection{Task Description}
\label{sec:task_description}

\begin{figure}[h!]
    \centering
    \includegraphics[width=0.92\linewidth]{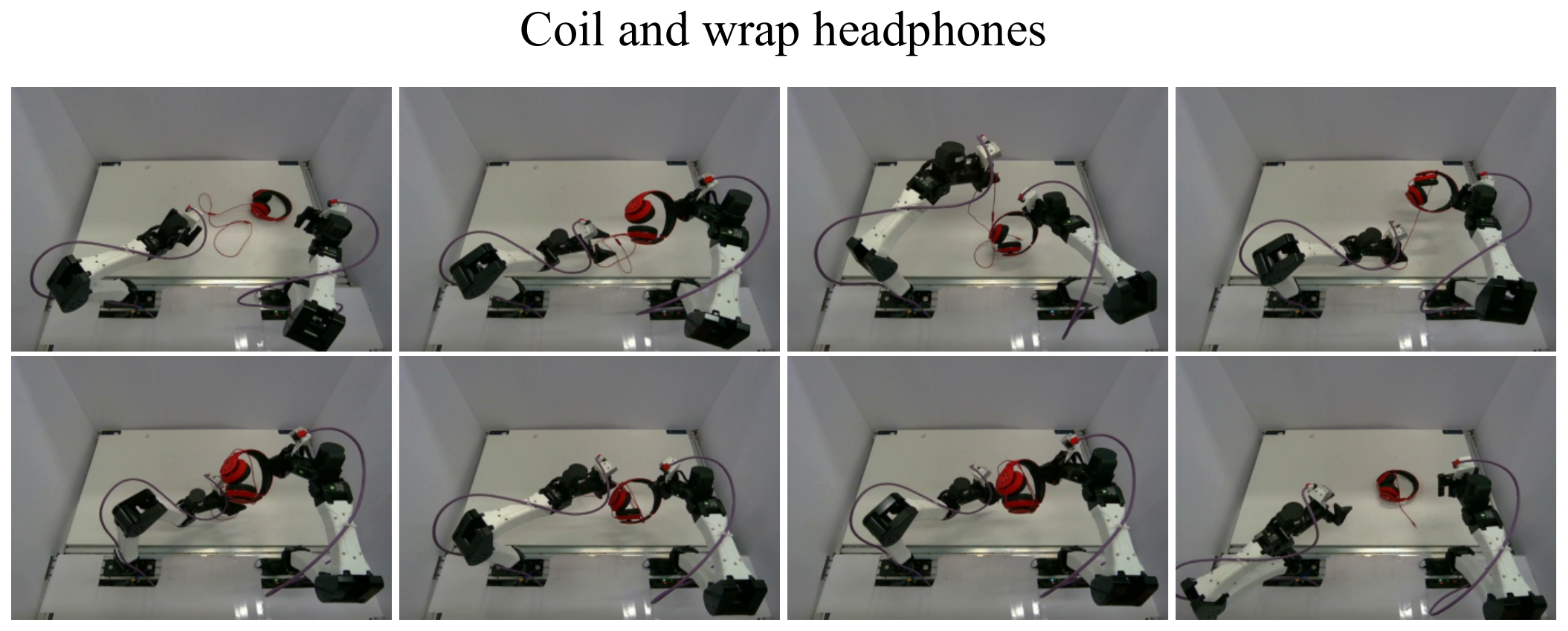}\\[0.6em]
    \includegraphics[width=0.92\linewidth]{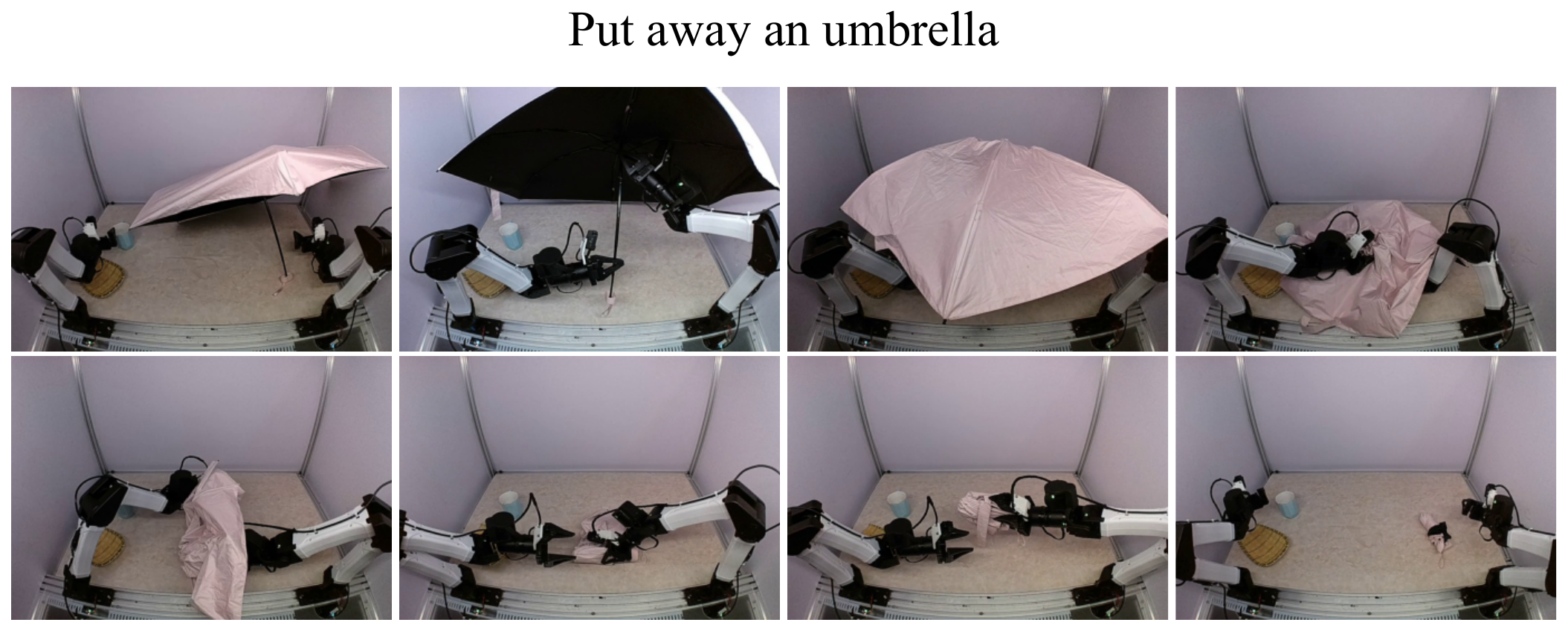}
    \caption{\textbf{Visualization of the 10 evaluation tasks.} Top-to-bottom, page-by-page: $\mathcal{S}_1$: (1) Pick and place plates into bin, (2) Pick and place plates into dish rack, (3) Folding the t-shirt, (4) Folding shorts, (5) Pull plug off the socket; $\mathcal{S}_2$: (6) Clean whiteboard with whiteboard eraser, (7) Set dinner table, (8) Put away an umbrella, (9) Sweep paper scraps with broom, (10) Coil and wrap headphones.}
    \label{fig:task_visualizations}
\end{figure}

We describe all 10 evaluation tasks below, grouped by subset, with the corresponding visualizations consolidated in Figure~\ref{fig:task_visualizations}. $\mathcal{S}_1$ (\textit{classic}) contains the two pick-and-place tasks, two cloth-folding tasks, and pulling plugs, all of which align well with primitives that dominate general reward-model training distributions. $\mathcal{S}_2$ (\textit{unconventional}) contains the remaining five tasks, which require tool use, deformable or articulated object manipulation, and multi-stage compositional execution. For tasks such as pick-and-place in which the same action sequence recurs many times within a single episode, the robot is required to repeat that sequence until the table is cleared.

\subsubsection*{$\mathcal{S}_1$: Classic Tasks}

\paragraph{Pick and place plates into bin.}
Several plates are arranged on the board, and the robot is required to clear them one at a time. In each cycle, the robot first picks up a plate from the center of the board and then places it in the plastic bin. The robot repeats this cycle until no plates remain on the board.

\paragraph{Pick and place plates into dish rack.}
This task is a variant of the bin task in which each plate must instead be inserted vertically into a slot of a dish rack, which requires precise pose alignment between the gripper and the rack opening. In each cycle, the robot first picks up a plate from the board and then inserts it into an available slot of the dish rack. The robot repeats this cycle until the board is cleared, which typically requires 5 to 10 insertion cycles per episode.

\paragraph{Fold the t-shirt.}
This is a long-horizon deformable-object task. The robot first grasps a T-shirt from the pile and transports it to the center of the board. It then flattens the garment and performs two consecutive folds. Finally, the robot places the neatly folded T-shirt in the corner of the board and resets both grippers. The flattening stage is the longest and visually most variable, whereas the folding stages are bimanual and require coordinated pinch grasps.

\paragraph{Fold the shorts.}
The procedure for this task closely mirrors that of folding the T-shirt. The robot first grasps the shorts from the pile and moves them to the center of the board. It then flattens the garment, performs two consecutive folds, and finally places the folded shorts in the corner of the board.

\paragraph{Pull plugs off the socket.}
The robot first repositions a power strip from its initial location on the board to a more convenient working location. It then iteratively removes each plug from the socket and deposits it at the center of the board. After several plugs have been removed, the robot may translate the power strip again to expose the remaining plugs from a more accessible angle, after which the unplug-and-place cycle continues. The episode terminates once all the plugs are removed from power strip and the power strip has been placed back at the center of the board.

\subsubsection*{$\mathcal{S}_2$: Unconventional Tasks}

\paragraph{Clean whiteboard with whiteboard eraser.}
One arm closely holds the small whiteboard to stabilize it throughout the task, while the other arm retrieves the whiteboard eraser placed at the center of the board. The holding arm then tilts the whiteboard to expose its surface, and the other arm wipes the letters from the whiteboard. After wiping, the robot places the eraser back at the center of the board and finally returns the whiteboard to its original position. The task is bimanual throughout, with one arm stabilizing the tilted whiteboard while the other performs the wiping motion.

\paragraph{Set dinner table.}
This is a composite long-horizon task that combines deformable manipulation with multi-object placement. The robot first spreads a tablecloth across the workspace and places a plate at the center of the cloth. It then picks up a glass with the right arm and sets it down at the appropriate location. Finally, the robot arranges the utensils, which involves left-to-right and right-to-left hand-overs to bring each utensil (a knife and a fork) into the appropriate gripper before its final placement on the tablecloth.

\paragraph{Sweep paper scraps with broom.}
The robot first picks up a broom from the center of the board, performing one or more bimanual hand-overs as needed to reach a stable wielding grasp, and then picks up a dustpan with the other arm. The main body of the task is a long repeated cycle: the robot sweeps paper scraps from the center of the board, sweeps them into the dustpan, and dumps the scraps from the dustpan into a trash bin. This sweep-gather-dump cycle is repeated until no scraps remain, after which the dustpan and broom are placed back on the board.

\paragraph{Coil and wrap headphones.}
This is a precision-demanding task. The robot first picks up a pair of headphones from the center of the board. It then performs a long bimanual wrapping motion that coils the cable around the body of the headphones. Once wrapping is complete, the robot places the wrapped headphones back at the center of the board and resets both arms.

\paragraph{Put away an umbrella.}
This is a multi-stage articulated-object task. The robot first picks up an open umbrella from the center of the board and passes it from the right arm to the left. It then presses the snap button against the runner to release the canopy and presses the shaft to collapse the umbrella. Next, the robot adjusts and folds the canopy along its natural creases. Finally, the canopy is rolled around the shaft, the strap is grasped and the velcro is fastened, and the closed umbrella is placed back at the center of the board.

\begin{figure}[p]
    \centering
    \includegraphics[width=0.92\linewidth]{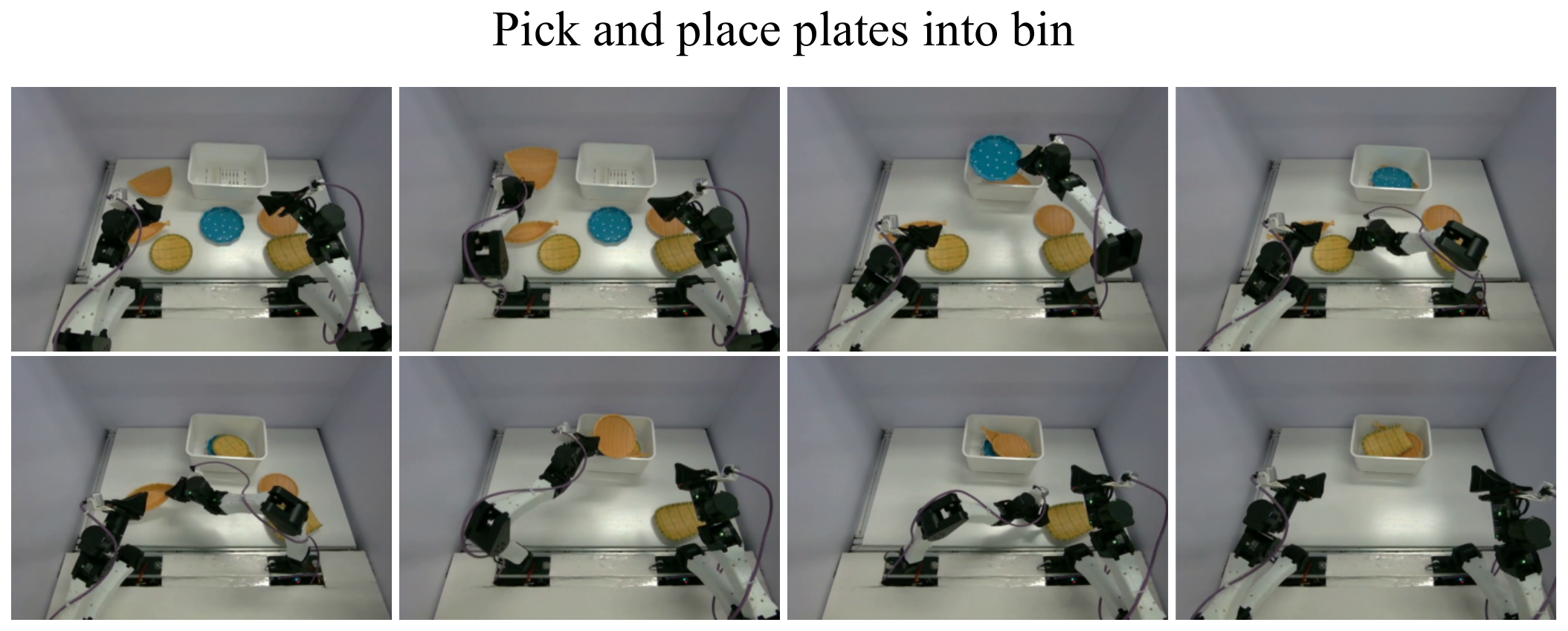}\\[0.6em]
    \includegraphics[width=0.92\linewidth]{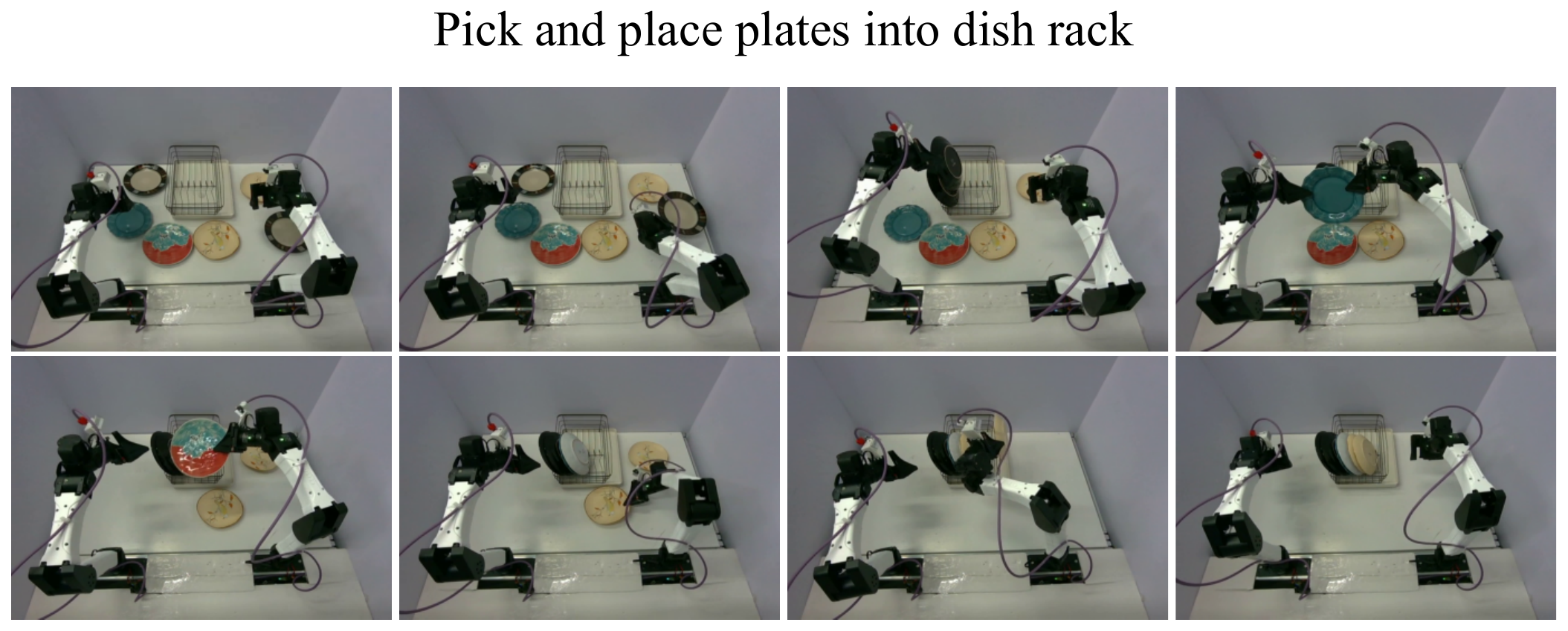}\\[0.6em]
    \includegraphics[width=0.92\linewidth]{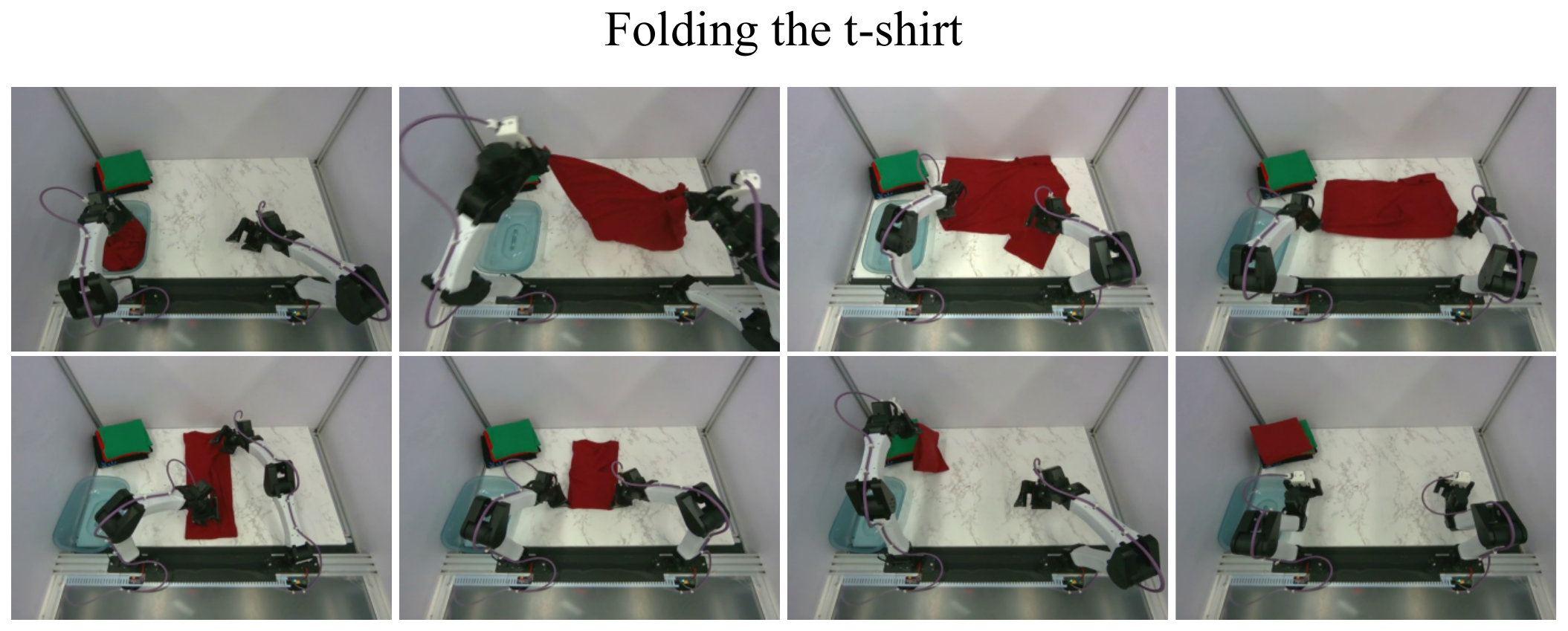}\\[0.6em]
    \includegraphics[width=0.92\linewidth]{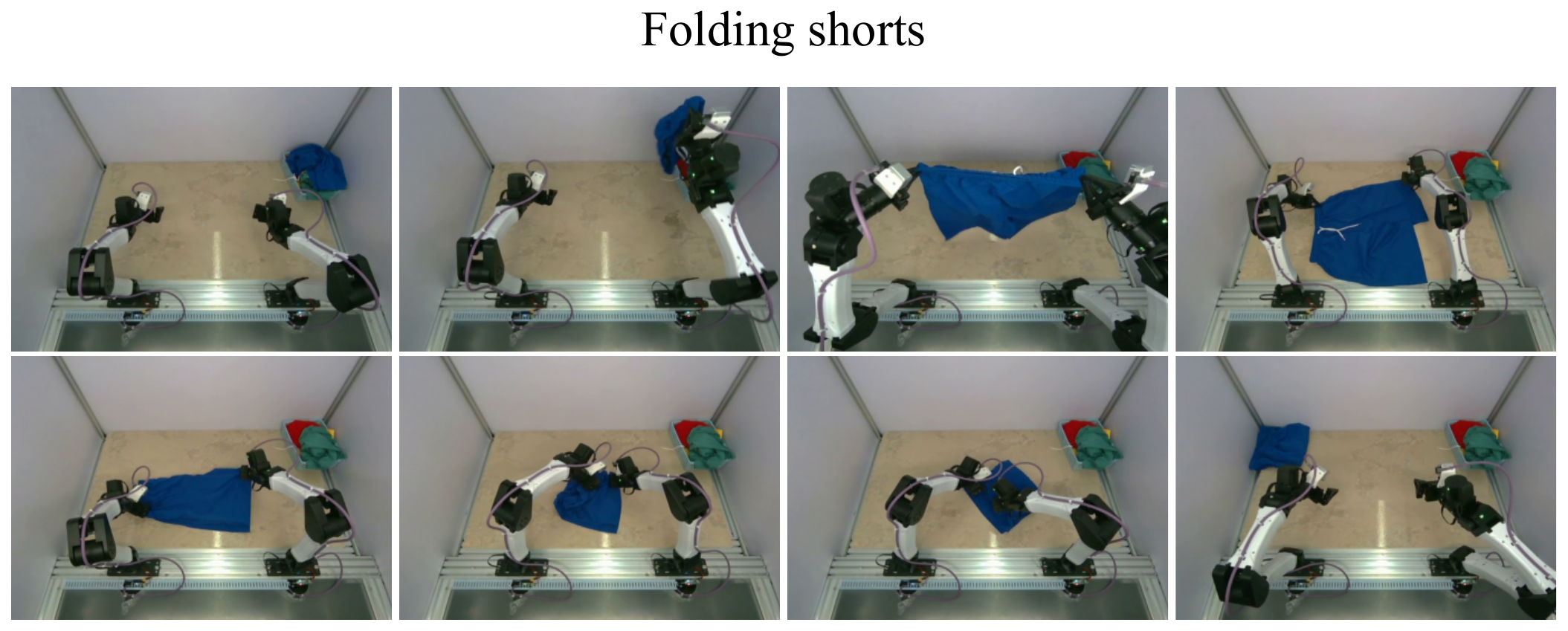}
\end{figure}

\begin{figure}[p]
    \centering
    \includegraphics[width=0.92\linewidth]{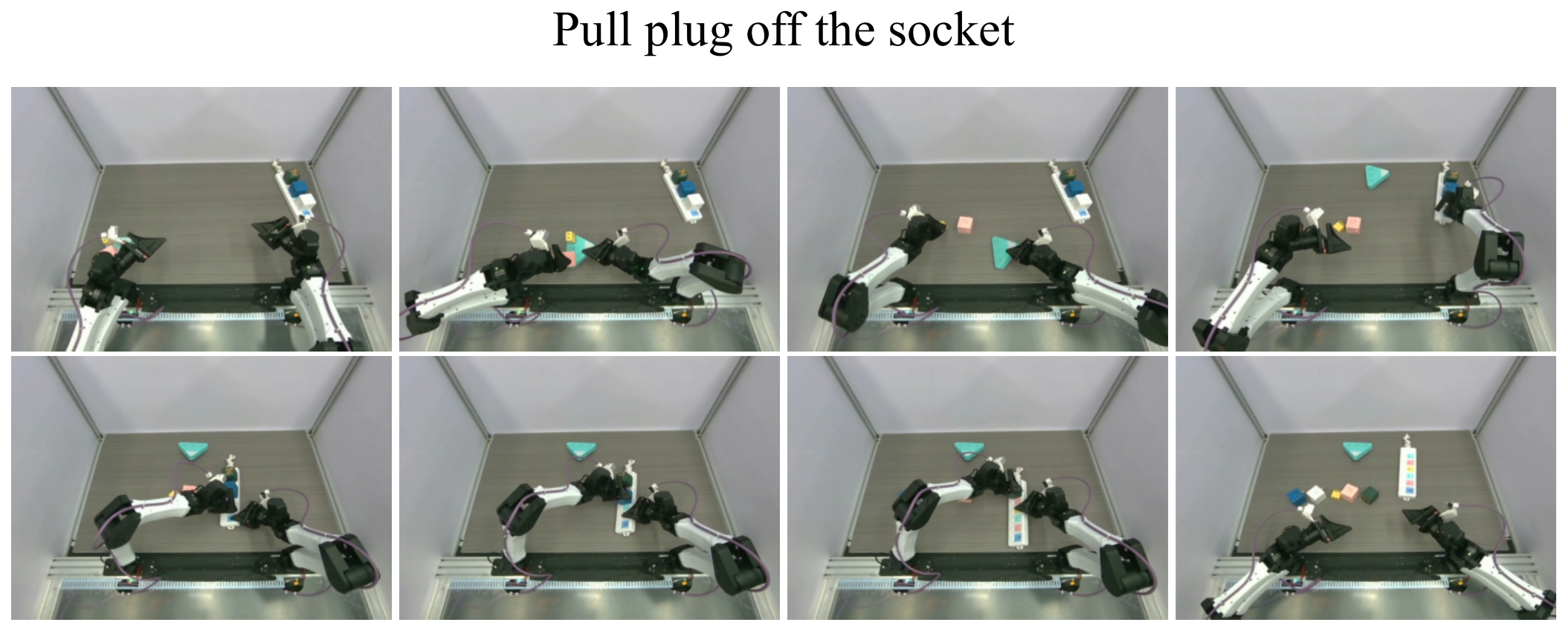}\\[0.6em]
    \includegraphics[width=0.92\linewidth]{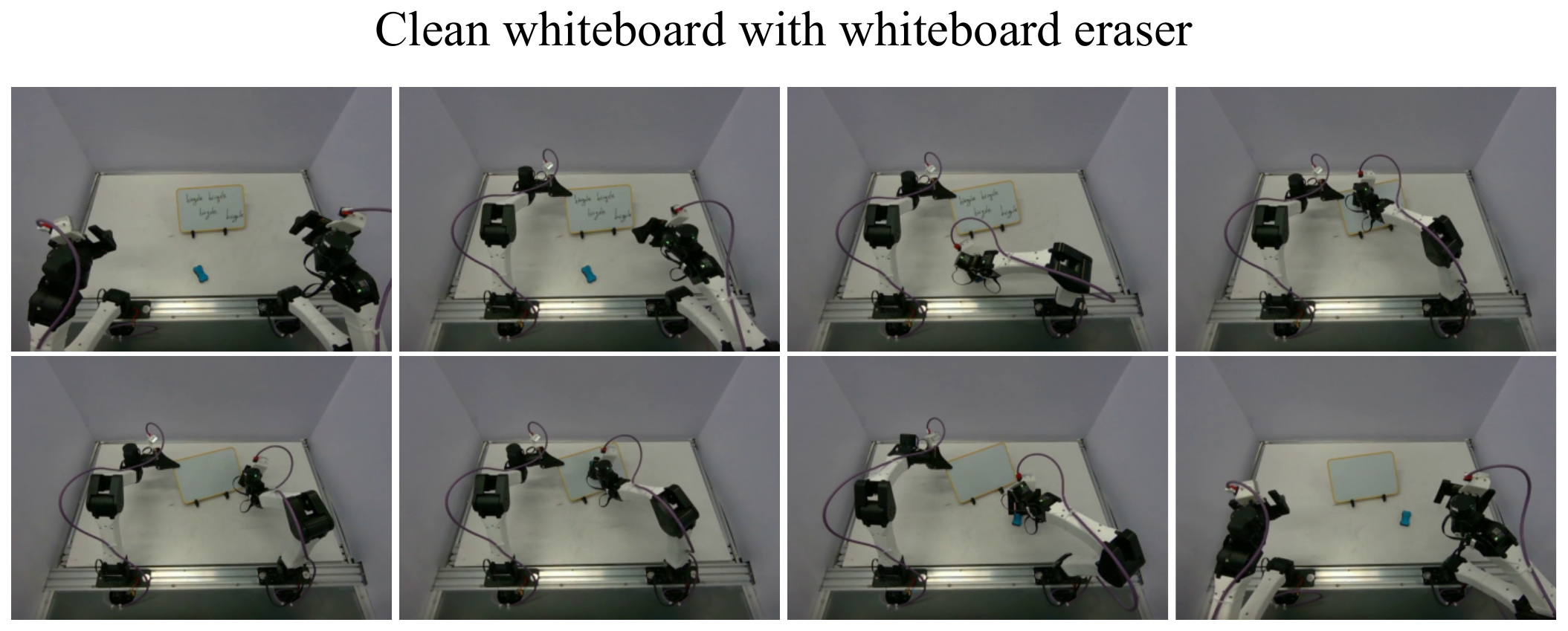}\\[0.6em]
    \includegraphics[width=0.92\linewidth]{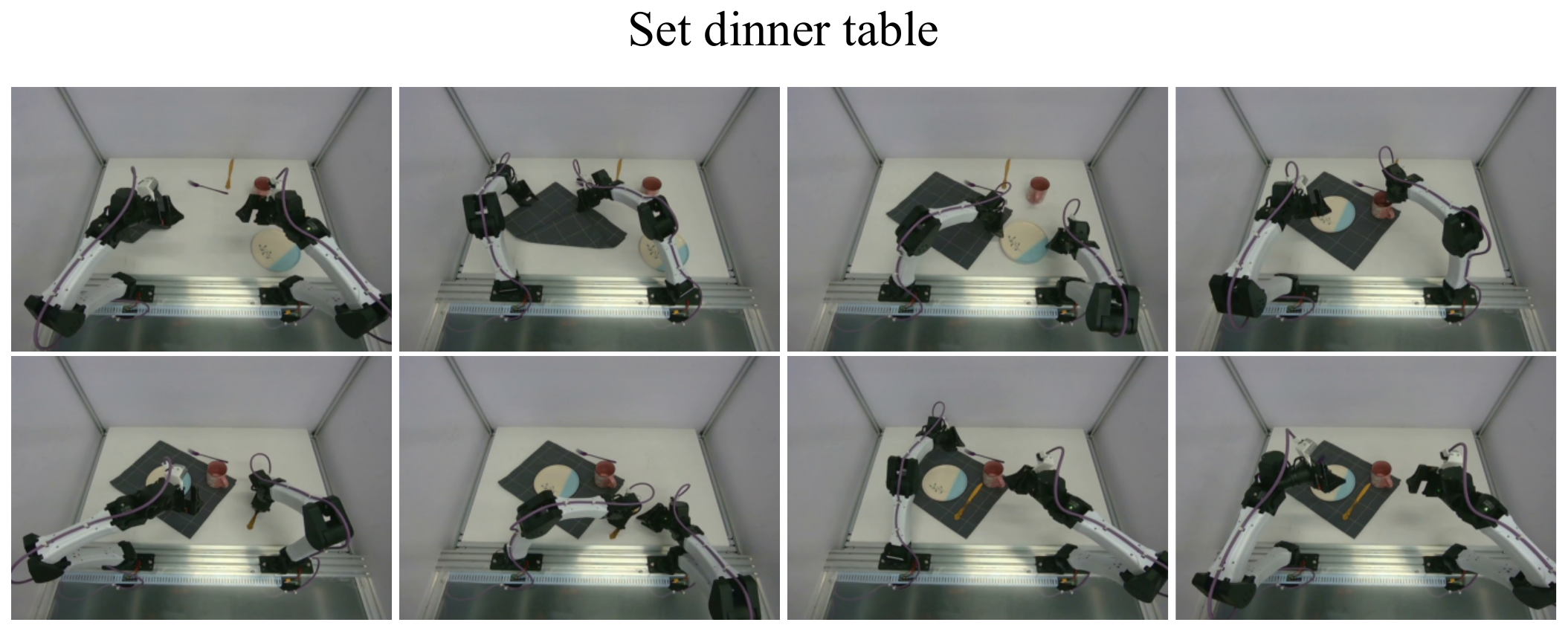}\\[0.6em]
    \includegraphics[width=0.92\linewidth]{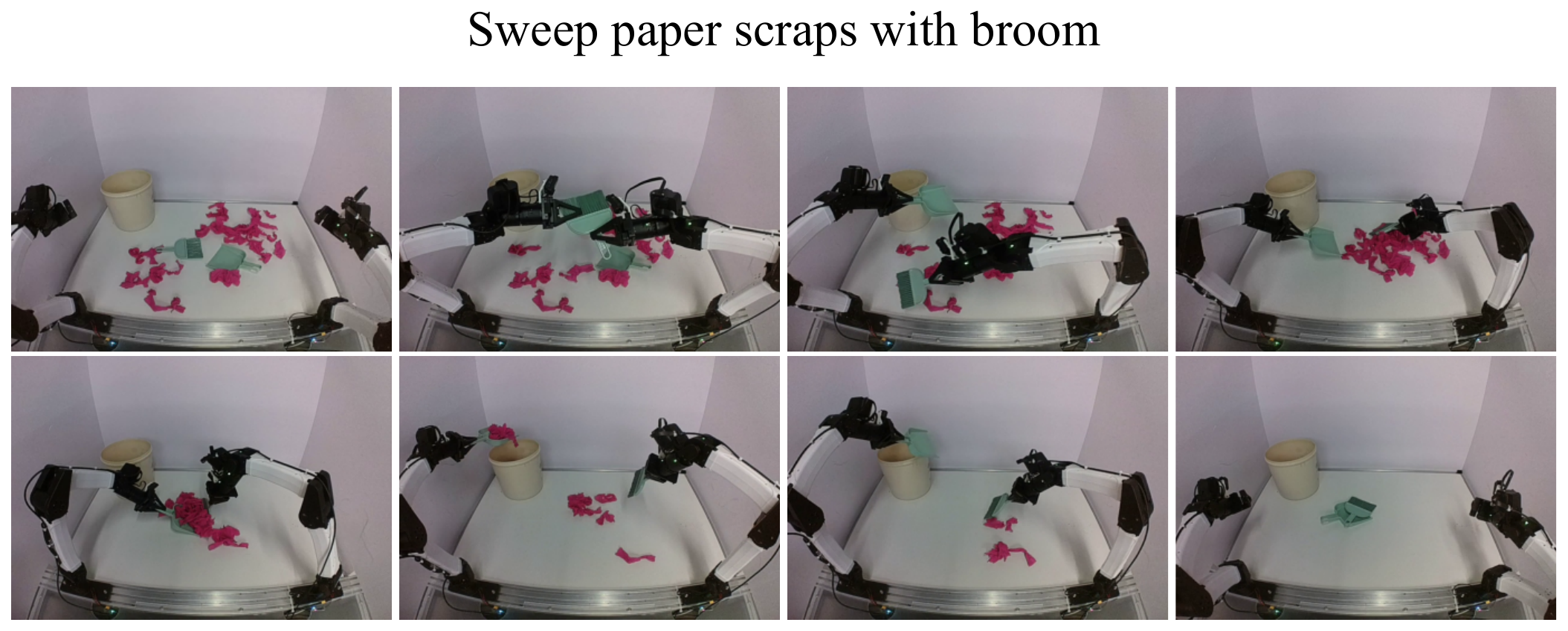}
\end{figure}

\clearpage
\subsection{Hyperparameter Table}
\label{appendix:training}

Table~\ref{tab:hyperparams} consolidates all hyperparameters introduced in the main text, covering the action-primitive stage estimator, the multi-gate MoE value decoder, and the SPIRAL self-improvement loop.

\begin{table}[h!]
\centering
\caption{\textbf{Hyperparameter summary.} Symbols, descriptions, and values for all hyperparameters referenced in the main text.}
\label{tab:hyperparams}
\small
\begin{tabular}{lll}
\toprule
\textbf{Symbol} & \textbf{Description} & \textbf{Value} \\
\midrule
\multicolumn{3}{l}{\emph{Action-primitive stage estimator}} \\
$K$ & Number of action primitives & 21 \\
$N$ & Number of recent context frames per timestep & 6 \\
$\Delta$ & Frame sampling stride & 30 \\
$L_{\text{AP}}$ & Transformer encoder layers & 4 \\
$d$ & Transformer hidden dimension & 512 \\
$t_k$ & Hours of data per primitive in $\mathcal{D}_{\text{AP}}$ & 3 \\
$\text{lr}_{\text{AP}}$ & Learning rate & $5 \times 10^{-5}$ \\
$B_{\text{AP}}$ & Batch size & 96 \\
$T_{\text{AP}}$ & Training steps & $10{,}000$ \\
$\text{opt}_{\text{AP}}$ & Optimizer & AdamW \\
\midrule
\multicolumn{3}{l}{\emph{Multi-gate MoE value decoder}} \\
$L_{\text{VD}}$ & Transformer encoder layers & 6 \\
$M$ & Number of primitive groups (gates) & 8 \\
$E$ & Number of experts in shared pool & 10 \\
$k$ & Top-$k$ routing per timestep & 2 \\
$d_{\text{exp}}$ & Expert hidden width & 256 \\
$n_{\mathrm{h}} \times d_{\text{exp}}$  & Expert MLP shape (depth $\times$ width)  & $3 \times 256$ \\
$\lambda_{\text{bal}}$ & Load-balance auxiliary loss weight & 2.5 \\
$\lambda_{\text{ent}}$ & Entropy auxiliary loss weight & 0.5 \\
$N_{\text{rw}}$ & Number of rewinding frames  & 3 \\
$\text{lr}_{\text{VD}}$ & Learning rate & $5 \times 10^{-5}$ \\
$B_{\text{VD}}$ & Batch size & 64 \\
$T_{\text{VD}}$ & Training steps & $10{,}000$ \\
$\text{opt}_{\text{VD}}$ & Optimizer & AdamW \\
\midrule
\multicolumn{3}{l}{\emph{Behavior Cloning}} \\
$h$ & Action chunk horizon & 50 \\
$\text{lr}_{\text{BC}}$ & Learning rate & $5 \times 10^{-5}$ \\
$B_{\text{BC}}$ & Batch size & 32 \\
$T_{\text{BC}}$ & Training steps & $30{,}000$ \\
$\text{wd}_{\text{BC}}$ & Weight decay & Cosine \\
$\text{opt}_{\text{BC}}$ & Optimizer & AdamW \\
\midrule
\multicolumn{3}{l}{\emph{SPIRAL self-improvement}} \\
$\gamma$ & Discount factor & 0.9995 \\
$\alpha$ & MC objective weight in $J_{\text{critic}}$ & 0.5 \\
$\beta$ & BC regularizer weight in residual actor loss & 30 \\
$N_{\text{critic}}$ & Critic ensemble size & 5 \\
$\sigma$ & Target-policy smoothing noise std & 0.2 \\
$c$ & Target-policy smoothing clip range & 0.5 \\
$\kappa$ & DICE multi-sample latent candidates per state (and best-of-$\kappa$ at deployment) & 4 \\
$n_{\mathrm{RL}} \times d_{\text{RL}}$  & Actor/Critic MLP shape (depth $\times$ width)  & $3 \times 1024$ \\
$\tau_{\text{tgt}}$ & Target network soft-update rate & 0.005 \\
$\text{lr}_{\text{RL}}$ & Learning rate (actor / critic) & $1 \times 10^{-4}$ \\
$B_{\text{RL}}$ & Batch size & 16 \\
$T_{\text{RL}}$ & Training steps per round & $5{,}000$ \\
$\text{opt}_{\text{RL}}$ & Optimizer & AdamW \\
\bottomrule
\end{tabular}
\end{table}

\subsection{Action Primitive Grouping Explaination}
\label{sec:action_primitive_groups}
To assign each of the $K{=}21$ action primitives to an MMoE gate, we cluster them into $M{+}1{=}8$ semantic groups according to shared visual appearance and end-effector motion patterns, so that primitives within a group exhibit similar progress signatures and can plausibly share the same gate's routing distribution. Table~\ref{tab:action_primitive_groups} lists the resulting groups, their member primitives, and a short description of the motion pattern each group captures.

\begin{table}[h!]
\centering
\caption{\textbf{Action-primitive grouping for the multi-gate MoE decoder.} The $K{+}1{=}22$ primitives are clustered into $M{+}1{=}8$ semantic groups by shared visual/motion characteristics; each group maps to one gate of the MMoE decoder while the expert pool is shared. \texttt{Other} is a catch-all for undefined or task-specific actions.}
\label{tab:action_primitive_groups}
\small
\begin{tabular}{
p{0.18\linewidth}
>{\centering\arraybackslash}p{0.26\linewidth}
>{\centering\arraybackslash}p{0.48\linewidth}
}
\toprule
\textbf{Group Name} & \textbf{Action Primitives} & \textbf{Description} \\
\midrule
Acquire
& scoop, pull, pick/grab
& Lift or move a target object up from its initial support surface into robot control. \\
Release
& place, pour/dump
& Move a controlled object down to a target location, container, or support surface. \\
Translate
& erase, zip/unzip, sweep, pass
& Move the end-effector or object along a primarily linear trajectory. \\
Insert
& install, insert
& Guide an object into a constrained target region with spatial alignment. \\
Shape Clothes
& hang, fold/flatten
& Reconfigure cloth-like objects into a desired geometric state or arrangement. \\
Rotate
& wrap, roll, tie/untie, rotate/turn
& Change an object's orientation or configuration through angular motion. \\
Force
& clamp, press
& Apply controlled contact force through the gripper or end-effector. \\
Other
& dummy
& Represent undefined or task-specific actions outside the main primitive taxonomy. \\
\bottomrule
\end{tabular}
\end{table}

\subsection{Algorithm of Reward-Guided Robot Self-Improvement Loop}
\begin{algorithm}[h!]
\caption{Reward-Guided Robot Self-Improvement Loop}
\label{alg:self_improvement}
\begin{algorithmic}[1]
\Require Human demonstrations $\mathcal{D}_{\text{demo}}$; pretrained VLA model $\pi_{\text{VLA}}$; pretrained multi-task reward model $\mathrm{RM}_1$; reward-model pretraining set $\mathcal{D}_{\text{RM}}$; max iterations $N$
\Ensure Improved policy $\pi_{N+1}$

\State \textbf{// Stage 1: Behavior cloning}
\State $\pi_1 \gets \textsc{Finetune}(\pi_{\text{VLA}}, \mathcal{D}_{\text{demo}})$ \Comment{supervised BC fine-tuning}
\State $\mathcal{R}_1 \gets \textsc{Rollout}(\pi_1, \text{n=50--100})$ \Comment{collect rollouts from weakest policy}

\State \textbf{// Stage 2a: One-time reward model adaptation to rollout distribution}
\State $\mathcal{R}_1^{\text{labeled}} \gets \textsc{HumanAnnotate}(\mathcal{R}_1)$ \Comment{segment-level labels + final progress}
\State $\mathcal{D}_{\text{adapt}} \gets \mathcal{R}_1^{\text{labeled}} \cup \textsc{Subsample}(\mathcal{D}_{\text{RM}}, 0.5)$ \Comment{mix in 50\% of pretraining data}
\State $\mathrm{RM}_2 \gets \textsc{Finetune}(\mathrm{RM}_1, \mathcal{D}_{\text{adapt}})$ \Comment{$\sim$1/10 of pretraining cost}

\State \textbf{// Stage 2b: Initial residual RL with pretrained reward model}
\State $\widetilde{\mathcal{D}}_{\text{demo}} \gets \textsc{LabelRewards}(\mathcal{D}_{\text{demo}}, \mathrm{RM}_1)$
\State $\pi_2 \gets \textsc{SPIRAL}(\pi_1, \widetilde{\mathcal{D}}_{\text{demo}})$
\State $\mathcal{R}_2 \gets \textsc{Rollout}(\pi_2, \text{n=100})$

\State \textbf{// Stage 4: On-policy self-improvement loop}
\For{$i = 2, 3, \ldots, N$}
    \State $\widetilde{\mathcal{R}}_i \gets \textsc{LabelRewards}(\mathcal{R}_i, \mathrm{RM}_2)$
    \State $\pi_{i+1} \gets \textsc{SPIRAL}(\pi_i, \widetilde{\mathcal{R}}_i)$
    \State $\mathcal{R}_{i+1} \gets \textsc{Rollout}(\pi_{i+1}, \text{n=100})$
    \If{$\textsc{SuccessRate}(\mathcal{R}_{i+1}) \geq \tau_{\text{success}}$}
        \State \textbf{break}
    \EndIf
\EndFor
\State \Return $\pi_{i+1}$
\end{algorithmic}
\end{algorithm}

\subsection{Reward Model Evaluation Protocol}
\label{sec:rm_protocol}

\paragraph{Demonstration evaluation.} We sample 10 held-out episodes per task and report the per-frame mean squared error (MSE) between the predicted and ground-truth progress. For clarity of presentation, we report MSE separately on $\mathcal{S}_1$ and $\mathcal{S}_2$, alongside a combined score. We note that the combined MSE is computed as a \textit{micro-average} over all evaluated frames; it is therefore not the arithmetic mean of the two per-subset values.

\paragraph{Rollout evaluation.} 
\label{sec:rollout_eval}
Following the protocol of SARM, we assess \textbf{robot rollout progress estimation} on a real robotic platform. We fine-tune a $\pi_{0.5}$~\citep{intelligence2025pi_} policy on each target task and deploy intermediate checkpoints from different training stages, yielding a balanced rollout set of 36 trajectories per task: 12 successful episodes (\texttt{SE}), 12 partially successful episodes (\texttt{PSE}), and 12 failed episodes (\texttt{FE}). Each rollout is classified by the reward model according to:
\begin{equation}
\texttt{Label} =
\begin{cases}
    \texttt{SE},  & \text{if } P_{\text{final}} > 0.8 \;\land\; \tfrac{3}{T} \sum_{t=2T/3}^{T} P_t > 0.6, \\[6pt]
    \texttt{PSE}, & \text{if } \tfrac{1}{T} \sum_{t=1}^{T} P_t \geq \xi, \\[6pt]
    \texttt{FE},  & \text{otherwise},
\end{cases}
\end{equation}
where $P_t$ denotes the predicted progress at frame $t$, $T$ is the trajectory length, and $\xi$ is set to the median of the average progress across all non-successful rollouts. This adaptive threshold yields an equal split between \textbf{PSE} and \textbf{FE} predictions and avoids the bias introduced by manually tuning a fixed cutoff.

We further summarize rollout classification accuracy via the score
\begin{equation}
    \rho = \frac{\#\text{correct} - \#\text{incorrect}}{36} \in [-1, 1],
\end{equation}
which assigns $+1$ for each correctly classified rollout and $-1$ otherwise, normalized by the total number of rollouts. A perfect classifier achieves $\rho = 1$, while random ternary guessing yields $\rho \approx -1/3$.

\paragraph{MoE routing diagnostics.} For all MoE-based reward models, we additionally report the \textbf{top-$k$ routing score}, defined as
\begin{equation}
\label{eqn:moe_load_score}
    \mathcal{S}_{\text{route}} \;=\; \frac{r - 1}{N/k - 1}, \qquad
    r \;=\; \frac{N}{k}\sum_{i=1}^{k} \bar{p}_{(i)},
\end{equation}
where $N$ is the total number of experts, $k$ is the number of activated experts per token, and $\bar{p}_{(i)}$ is the $i$-th largest entry of the routing distribution averaged over the evaluation set. The raw concentration $r$ measures the mass placed on the top-$k$ experts relative to a uniform baseline $1/N$, ranging from $1$ (perfectly balanced) to $N/k$ (fully collapsed); the affine normalization rescales $\mathcal{S}_{\text{route}}$ to $[0, 1]$, removing the dependence on $N$ and $k$ so expert utilization can be compared directly across configurations. A healthy router yields $\mathcal{S}_{\text{route}}$ well below $1$.

\subsection{Policy Evaluation Protocol}

\paragraph{Cleaning Whiteboard -- five-tier scoring.}
\label{sec:wb_progress}
Each episode of the Cleaning Whiteboard evaluation is scored using five progress tiers: \textbf{0\%} for no letters cleaned; \textbf{25\%} for successfully grasping the eraser, stabilizing the board, and cleaning at least one letter; \textbf{50\%} for cleaning more than half but not all letters; \textbf{75\%} for cleaning all letters without stopping correctly, or stopping early with letters remaining; and \textbf{100\%} for cleaning all letters and terminating correctly by releasing the eraser on the table and board. The reported \textbf{Avg. Prog.} metric is the mean tier value across all 20 evaluation episodes, while \textbf{SR} counts only episodes that reach the 100\% tier.

\subsection{MoE Design: Comparison with Dense Model}
\label{sec:moe_dense_budget}

To put the architectural comparison on equal footing, we contrast our decoder-level Mixture-of-Experts model (henceforth \textbf{MoE-Decoder}) against a parameter-comparable dense baseline. Both models share an identical encoder backbone and differ only in the reward-prediction head, which is applied once per timestep to the fused feature vector of dimension $(N{+}3)\,d = 3072$, where the $N{=}3$ camera streams together with the language, task, and state tokens give $(N{+}3){=}6$ modality streams of length $T$. The \textbf{Dense} baseline computes the scalar reward with a 3-hidden-layer MLP of width $512$. The \textbf{MoE-Decoder} variant replaces this MLP with a sparse top-$k$ mixture of $E$ deep experts, each itself a 3-hidden-layer MLP of width $256$, applied to the same fused vector. Because the MoE activates only $k{=}2$ of $E{=}10$ experts per timestep, the relevant comparison quantity is its \emph{activated} parameter count, not its total. With this configuration, MoE-Decoder is matched to the Dense baseline within $\sim\!10\%$ on both activated parameters and FLOPs (Tables~\ref{tab:params} and~\ref{tab:flops}), so any quality difference is attributable to the sparse routing structure rather than to capacity or compute headroom.

\begin{table}[h!]
\centering
\caption{Activated-parameter breakdown of the MoE-Decoder and Dense reward heads. Configuration: fused input dimension $3072$, $E{=}10$ experts of width $256$ with $n_{\mathrm{h}}{=}3$ hidden layers, $k{=}2$.}
\label{tab:params}
\begin{tabular}{lr}
\toprule
\textbf{Component} & \textbf{Params} \\
\midrule
\multicolumn{2}{l}{\emph{MoE-Decoder (activated)}} \\
\quad Active gate (LN, Linear $3072{\to}10$) & $0.04$~M \\
\quad Active expert (LN, Linear $3072{\to}256$, Linear $256{\to}256{\times}2$, Linear $256{\to}1$) & $0.92$~M \\
\quad $\times\,k{=}2$ active experts & $1.85$~M \\
\textbf{MoE-Decoder total (activated)} & \textbf{1.89~M} \\
\midrule
\multicolumn{2}{l}{\emph{Dense baseline}} \\
\quad MLP head: LN, Linear $3072{\to}512$ & $1.58$~M \\
\quad \phantom{MLP head: }Linear $512{\to}512$ ($\times 2$) & $0.53$~M \\
\quad \phantom{MLP head: }Linear $512{\to}1$ & $<\!0.01$~M \\
\textbf{Dense total} & \textbf{2.11~M} \\
\midrule
MoE-Decoder $/$ Dense (activated) & $0.90\times$ \\
\bottomrule
\end{tabular}
\end{table}

\begin{table}[h!]
\centering
\caption{Per-timestep compute of the reward head, in FLOPs (one multiply-accumulate counted as $2$~FLOPs). Both heads run once per timestep on the fused vector.}
\label{tab:flops}
\begin{tabular}{lrrr}
\toprule
\textbf{Component} & \textbf{FLOPs / call} & \textbf{Calls / timestep} & \textbf{FLOPs / timestep} \\
\midrule
\multicolumn{4}{l}{\emph{MoE-Decoder}} \\
\quad Active gate & $0.08$~M & $1$ & $0.08$~M \\
\quad Active expert ($\times k{=}2$) & $1.85$~M & $2$ & $3.70$~M \\
\textbf{MoE-Decoder total} & & & \textbf{3.78~M} \\
\midrule
\multicolumn{4}{l}{\emph{Dense baseline}} \\
\quad MLP head & $4.20$~M & $1$ & $4.20$~M \\
\textbf{Dense total} & & & \textbf{4.20~M} \\
\midrule
MoE-Decoder $/$ Dense & & & $0.90\times$ \\
\bottomrule
\end{tabular}
\end{table}

\subsection{MoE Settings Ablation study}

\begin{table}[h!]
\small
\centering
\caption{\textbf{Ablation over MoE routing configurations of \RM.} \textbf{T$a$E$b$}: top-$k{=}a$ over $b$ experts; \textbf{T2E10} is our default. We sweep $k \in \{1,2,4\}$ at fixed $E{=}10$ and $E \in \{5,10,20\}$ at fixed $k{=}2$; all other components are held identical. Lower demo MSE is better.}
\label{tab:moe_ablation}
\begin{tabular}{lccccc}
\toprule
\textbf{Metrics} & \textbf{T1E10} & \textbf{T4E10} & \textbf{T2E5} & \textbf{T2E20} & \textbf{T2E10 (ours)} \\
\midrule
\multicolumn{6}{l}{\emph{Demo $\mathcal{L}$ $\downarrow$}} \\
$\mathcal{S}_1$         & 0.015 & 0.018 & 0.016 & 0.011 & \textbf{0.006} \\
$\mathcal{S}_2$         & 0.033 & 0.037 & 0.032 & \textbf{0.031} & \textbf{0.031} \\
Overall                 & 0.026 & 0.029 & 0.024 & 0.022 & \textbf{0.020} \\
\midrule
\multicolumn{6}{l}{\emph{Rollout $\rho$ $\uparrow$}} \\
T1              & 0.667    & 0.722    & 0.611    & 0.778  & \textbf{0.833} \\
T2              & 0.556    & 0.5      & 0.5      & \textbf{0.722}  & 0.667 \\
\midrule
\multicolumn{6}{l}{\emph{MoE Density}} \\
 & 0.02 & 0.13 & 0.20 & 0.12 & 0.10 \\
\bottomrule
\end{tabular}
\end{table}

Our default configuration \textbf{T2E10} attains the best overall demo loss (0.020) and the strongest $T_1$ rollout success (0.833), indicating that $k{=}2$ over $E{=}10$ experts essentially reaches the optimum within this sweep. Thanks to the multi-gate design and the load-balance auxiliary loss, all variants maintain non-trivial load scores (0.02--0.20) and remain away from routing collapse, so the observed differences reflect specialization quality rather than degenerate gating. Empirically, varying top-$k$ (T1E10: 0.026, T4E10: 0.029) degrades the overall demo loss more than varying expert count (T2E5: 0.024, T2E20: 0.022); we attribute this to the fact that $k$ directly controls the effective capacity and gradient sparsity per token, where $k{=}1$ removes the redundancy needed for stable specialization and $k{=}4$ over-mixes experts and dilutes the routing signal, whereas changing $E$ only re-partitions an otherwise well-balanced expert pool. This suggests that the routing sparsity level is a more sensitive hyperparameter than the raw expert budget in our per-timestep MoE decoder.

\subsection{MoE Design: MoE-Decoder vs.\ MoE-FFN}
\label{sec:decoder_ffn}

\paragraph{Setup.}
The dominant MoE recipe in large language models places experts inside the Transformer Feed-Forward Network (FFN) sublayer~\citep{fedus2022switch, lepikhin2020gshard, jiang2024mixtral, dai2024deepseekmoe, zheng2026kern}. We ask whether the same placement transfers to multimodal reward modeling by comparing two variants of \RM's MMoE value decoder that are identical in every respect except where the inner MoE is inserted:
\begin{itemize}\itemsep0pt
    \item \textbf{MoE-FFN} replaces the feed-forward sublayer of the last Transformer block of the value model backbone with a sparse MoE FFN; the per-timestep output head is a plain MLP.
    \item \textbf{MoE-Decoder (ours)} keeps the value model backbone fully dense and places the sparse MoE in the output head, acting on the fused-per-timestep representation produced by the encoder.
\end{itemize}
Both variants share the same SigLIP-2 features, the same 6-layer causal Transformer backbone, the same expert count $E{=}10$ and $k{=}2$, and the same primitive-group-conditioned MMoE outer gate (selected by the stage estimator's predicted primitive $\tilde{y}_t$, as in Section~\ref{sec:stage_estimation}). The dense FFN sublayer of the value-model backbone follows the standard $4{\times}$ expansion, giving hidden width $4 d_{\text{model}} = 4{\times}512 = 2048$. For MoE-FFN, each expert is a feed-forward block of hidden width $1024$, and with $k{=}2$ activated per token the active hidden capacity is $1024{\times}2 = 2048$, which matches the dense FFN exactly, so the comparison is compute- and capacity-matched at the FFN sublayer. The inner top-$k$ router is content-driven on whatever vector the placement hands it: for MoE-FFN that is one (modality, timestep) hidden state; for MoE-Decoder that is the fused timestep vector concatenating all modalities. Primitive-group specialization is therefore not a differentiator; the only thing that changes between the two is the sublayer in which the experts live.

We report demo MSE on the $\mathcal{S}_1$ and $\mathcal{S}_2$ subsets (lower is better), rollout score $\rho$ on T1 (Folding Shorts) and T2 (Cleaning Whiteboard) (higher is better), and the normalized top-$k$ routing score $\mathcal{S}_{\text{route}}$ of Equation~\ref{eqn:moe_load_score} as a routing-health / compute-parity check (below $0.4$ is generally healthy).

\begin{table}[h!]
\small
\centering
\caption{\textbf{MoE-Decoder vs.\ MoE-FFN on \RM.} The only difference is the sublayer in which the inner MoE is placed; all other components, including the primitive-group-conditioned MMoE outer gate, are held identical. MoE-Decoder matches MoE-FFN's routing sparsity (MoE Density) yet halves demo MSE overall and roughly triples rollout score on both tasks.}
\label{tab:ffn_ablation}
\begin{tabular}{lcc}
\toprule
\textbf{Metrics} & \textbf{MoE-FFN} & \textbf{MoE-Decoder (ours)} \\
\midrule
\multicolumn{3}{l}{\emph{Demo $\mathcal{L}$ $\downarrow$}} \\
$\mathcal{S}_1$ & 0.026 & \textbf{0.006} \\
$\mathcal{S}_2$ & 0.038 & \textbf{0.031} \\
Overall         & 0.033 & \textbf{0.020} \\
\midrule
\multicolumn{3}{l}{\emph{Rollout $\rho$ $\uparrow$}} \\
T1 & 0.222 & \textbf{0.833} \\
T2 & 0.278 & \textbf{0.667} \\
\midrule
\multicolumn{3}{l}{\emph{MoE Density}} \\
 & 0.09 & 0.10 \\
\bottomrule
\end{tabular}
\end{table}

\paragraph{Findings.}
MoE-Decoder more than halves demo MSE overall ($0.020$ vs.\ $0.033$) and roughly triples rollout score on T1 ($0.833$ vs.\ $0.222$) and T2 ($0.667$ vs.\ $0.278$) at essentially identical routing sparsity. The result inverts the LLM intuition that MoE belongs in the FFN. The rest of this section explains why, structured around what each sublayer is for.

\paragraph{What each sublayer is for.}
A Transformer \emph{FFN sublayer} performs position-wise representation refinement: attention mixes information across positions, and the FFN re-encodes each position in isolation. Stacking attention and FFN is the mechanism by which the value model backbone builds a single coherent fused-per-timestep representation from all modalities. The \emph{output head} (which we call the decoder in this paper) consumes that fused representation and maps it to the predicted progress $\hat{r}_t$. Placing the inner MoE in the FFN therefore specializes \emph{representation building}; placing it in the output head specializes \emph{the prediction mapping}. These are very different things.

\paragraph{(i) MoE-FFN specializes the wrong stage.}
Per-position experts in the encoder FFN route each (modality, timestep) hidden state through a different MLP, and the following attention layer must re-fuse those divergently-processed positions back into a coherent representation. This fragments the backbone's job. Worse, the per-position vector the FFN router sees inherits its identity primarily from the modality projection it came through, not from scene content. Even when content-driven specialization does emerge, it acts on intermediate per-position representations that the prediction does not consume directly; $\hat{r}_t$ depends only on the final fused timestep representation, which is several attention layers downstream of the MoE.

\paragraph{(ii) MoE-Decoder specializes the right stage.}
With the value model backbone dense, all attention$+$FFN layers cooperate to produce one unified fused-per-timestep representation, as the backbone is designed to. The inner MoE then routes that fused vector to one of several scalar output paths, conditioned on its content and primitive-group gate. This is the standard ``shared backbone, specialized head'' pattern made soft and top-$k$: no representation is fragmented, and specialization lands on the axis that genuinely benefits from it. This makes different primitive groups (e.g., \texttt{pick}/\texttt{place}, \texttt{rotate}, \texttt{fold}) call for different progress computations on top of similar fused features. Critically, every routing decision in the output head is supervised directly by $\hat{r}_t$, so the routing gradient is high signal-to-noise.

\paragraph{(iii) Why this flips the LLM intuition.}
In language modeling, the per-position token \emph{is} the prediction unit: each position predicts the next token from its own representation, so per-position MoE-FFN specialization aligns with the prediction axis, and the long context length ($L \!\sim\! 10^{3}\!\!-\!10^{4}$) supplies enough routing decisions per sample to stabilize specialization. In multimodal reward modeling neither holds. The prediction unit is the fused timestep, not the per-modality position; the per-sample token budget is small ($L=(N{+}3)T$); and modality typing biases per-position routing toward modality-identity shortcuts. MoE-Decoder places specialization at the level where prediction actually happens, paying the sparsity cost where it buys differentiation rather than where it fragments shared representation.

\paragraph{Summary.}
The Transformer FFN is for building a unified representation; the output head is for mapping that representation to the prediction. Sparse routing helps when it specializes the right thing. In LLMs the FFN \emph{is} the right thing because per-position prediction makes per-position specialization useful. In \RM's MMoE value decoder the right thing is the output head: the value model backbone must stay coherent, and the head is where primitive-conditioned specialization pays off. MoE-Decoder matches placement to prediction; MoE-FFN does not. The numbers in Table~\ref{tab:ffn_ablation} follow from this mismatch.

\subsection{Reward Model Estimation Visualization Results}
\begin{figure}[h!]
    \centering
    \includegraphics[width=0.48\linewidth]{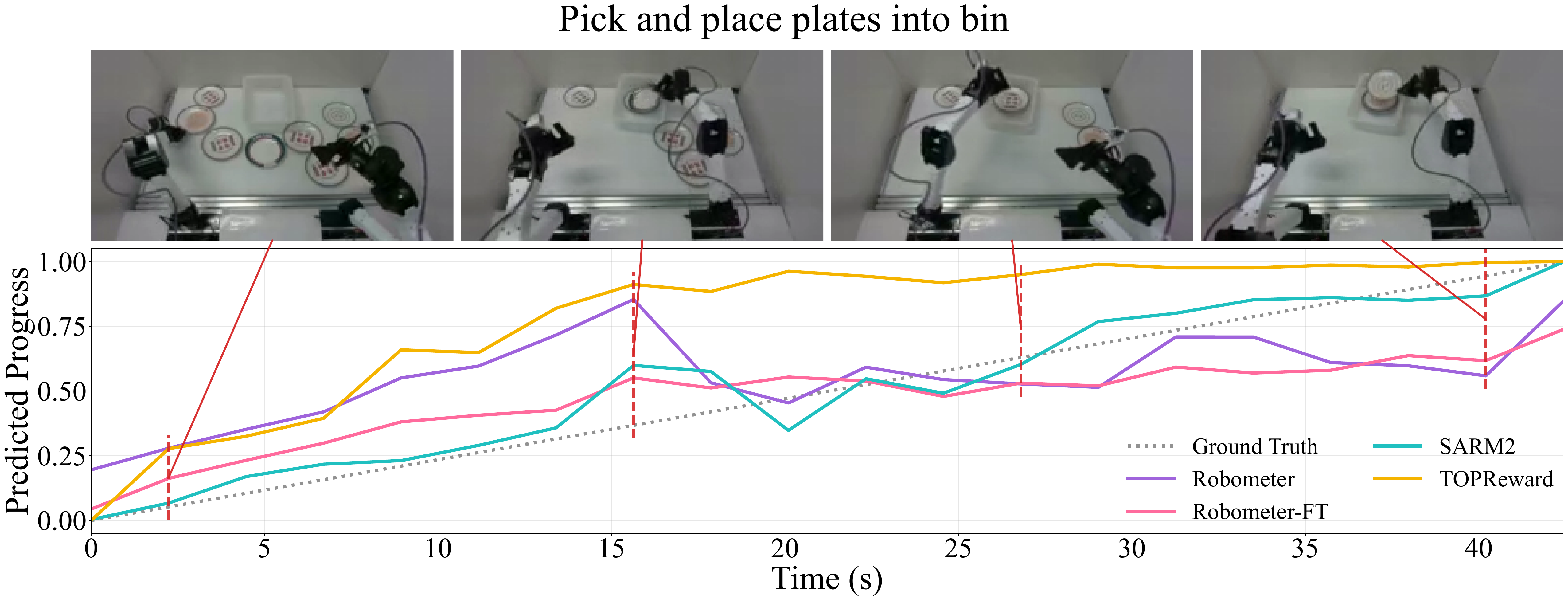}
    \hfill
    \includegraphics[width=0.48\linewidth]{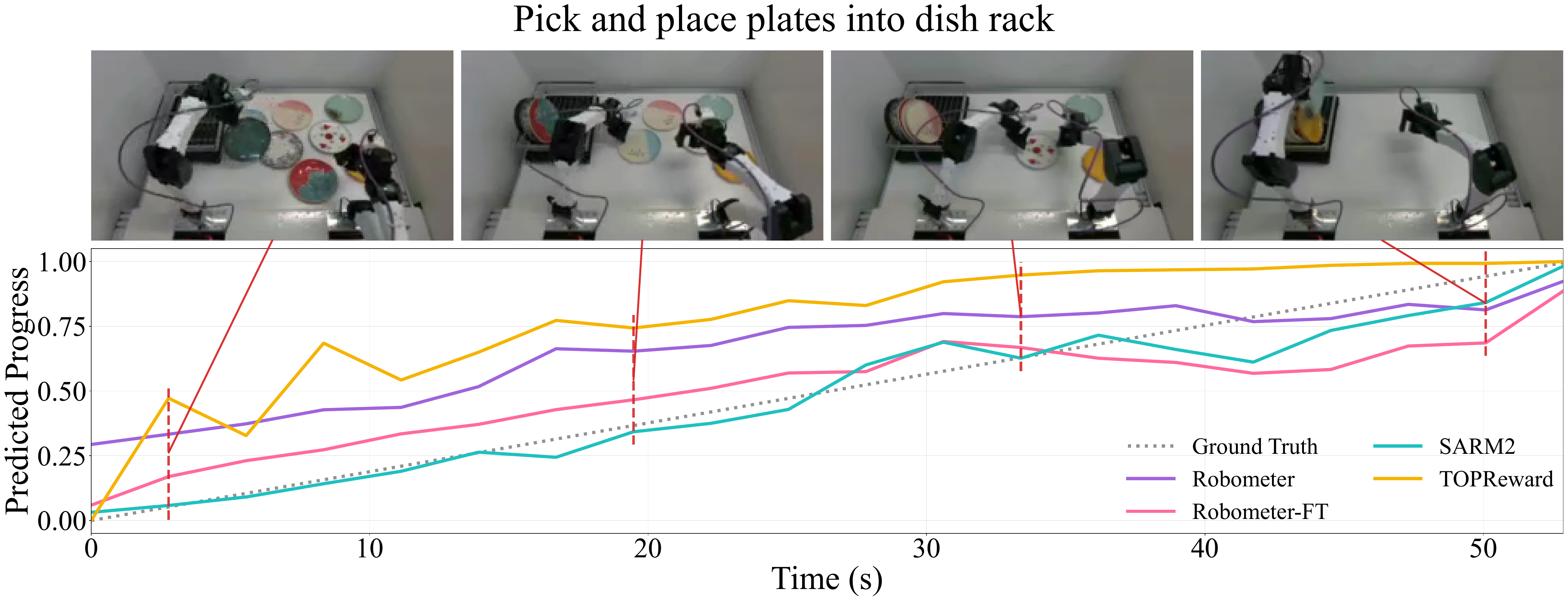}

    \includegraphics[width=0.48\linewidth]{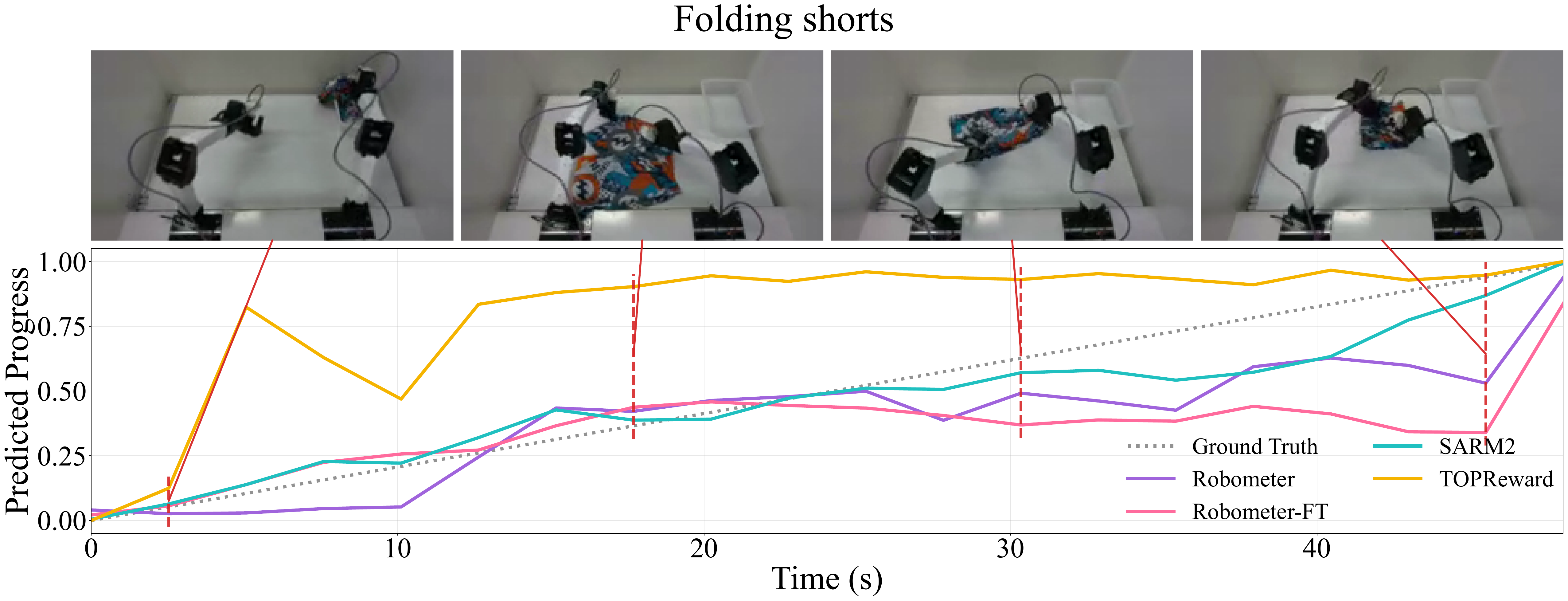}
    \hfill
    \includegraphics[width=0.48\linewidth]{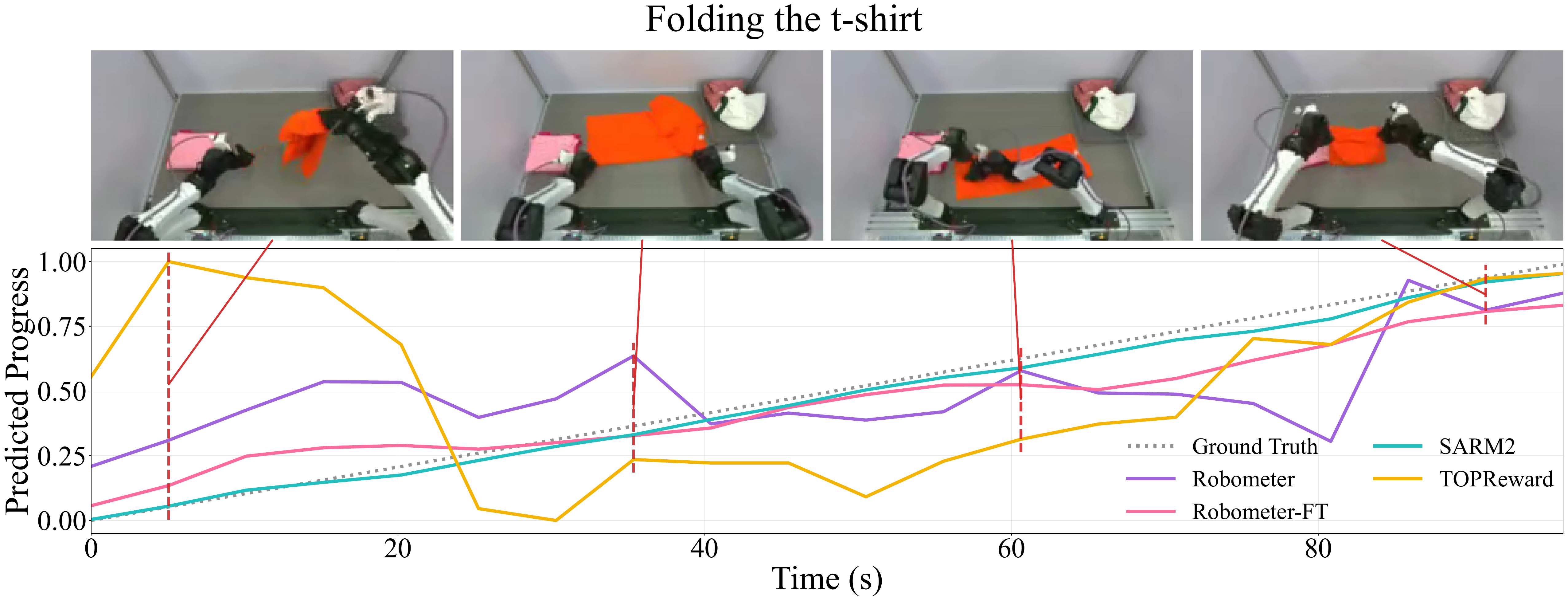}

    \includegraphics[width=0.48\linewidth]{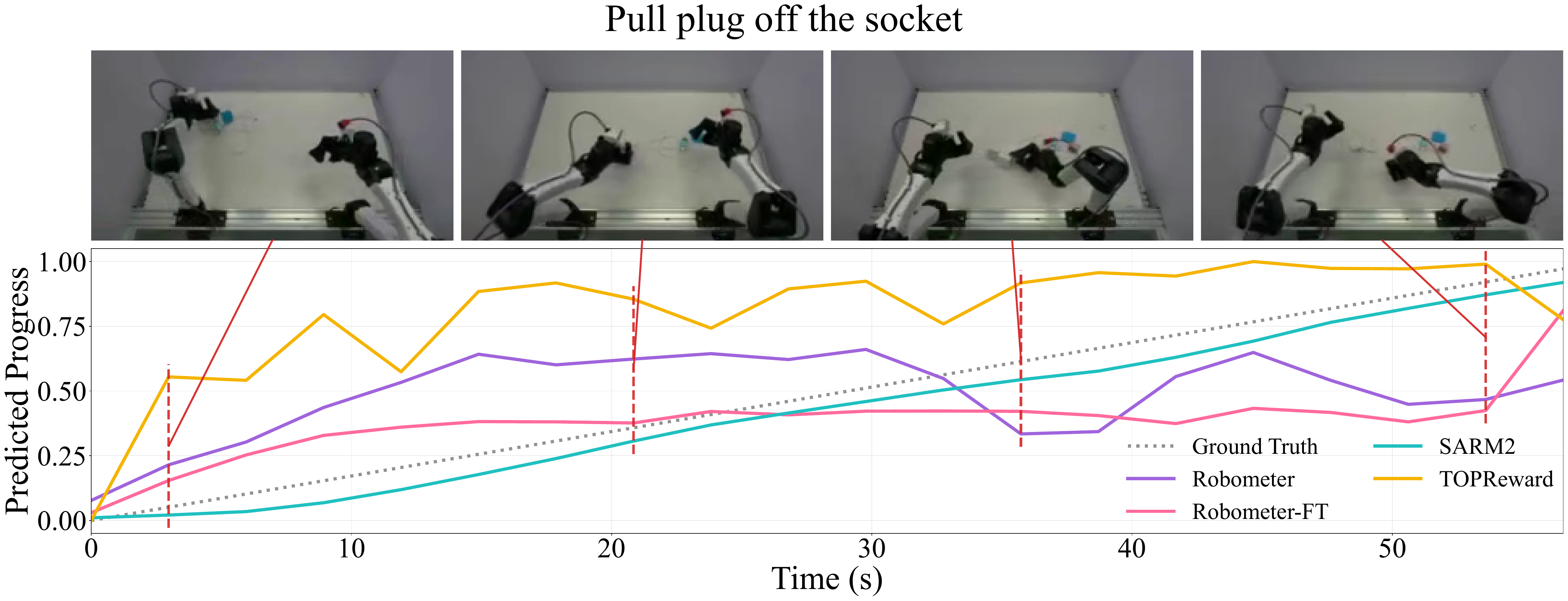}
    \hfill
    \includegraphics[width=0.48\linewidth]{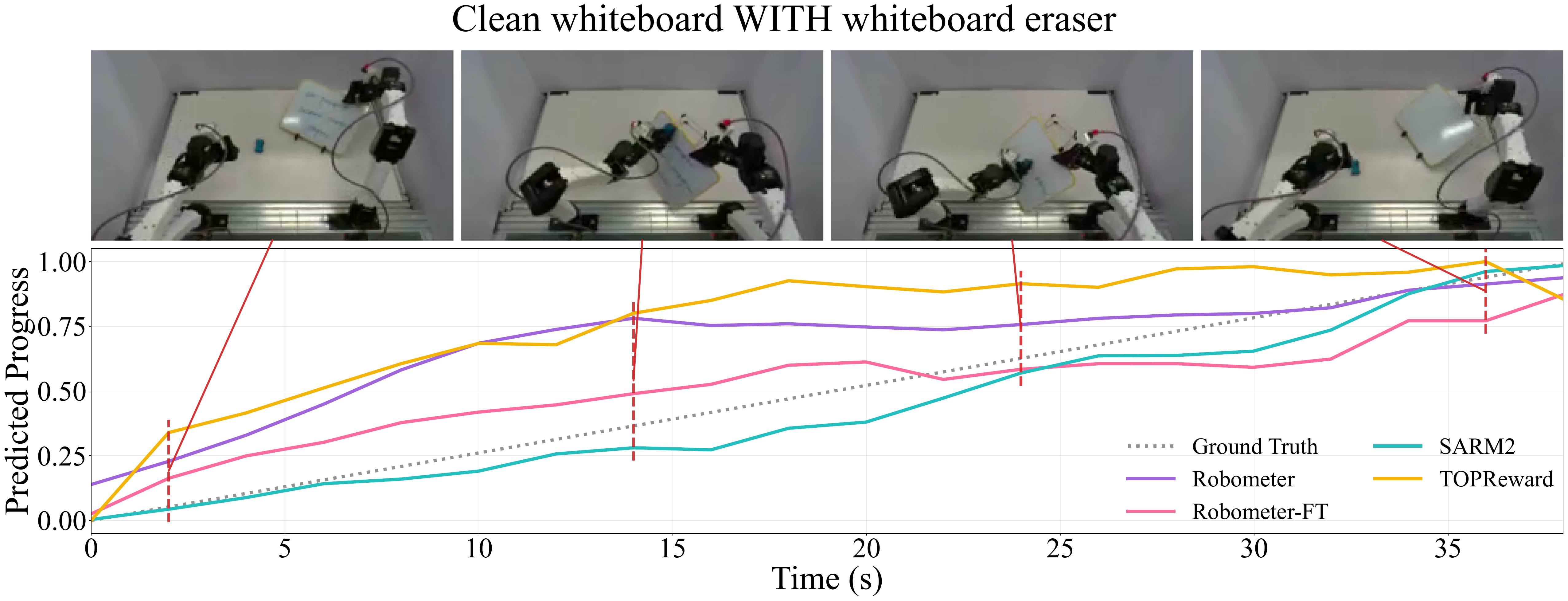}

    \includegraphics[width=0.48\linewidth]{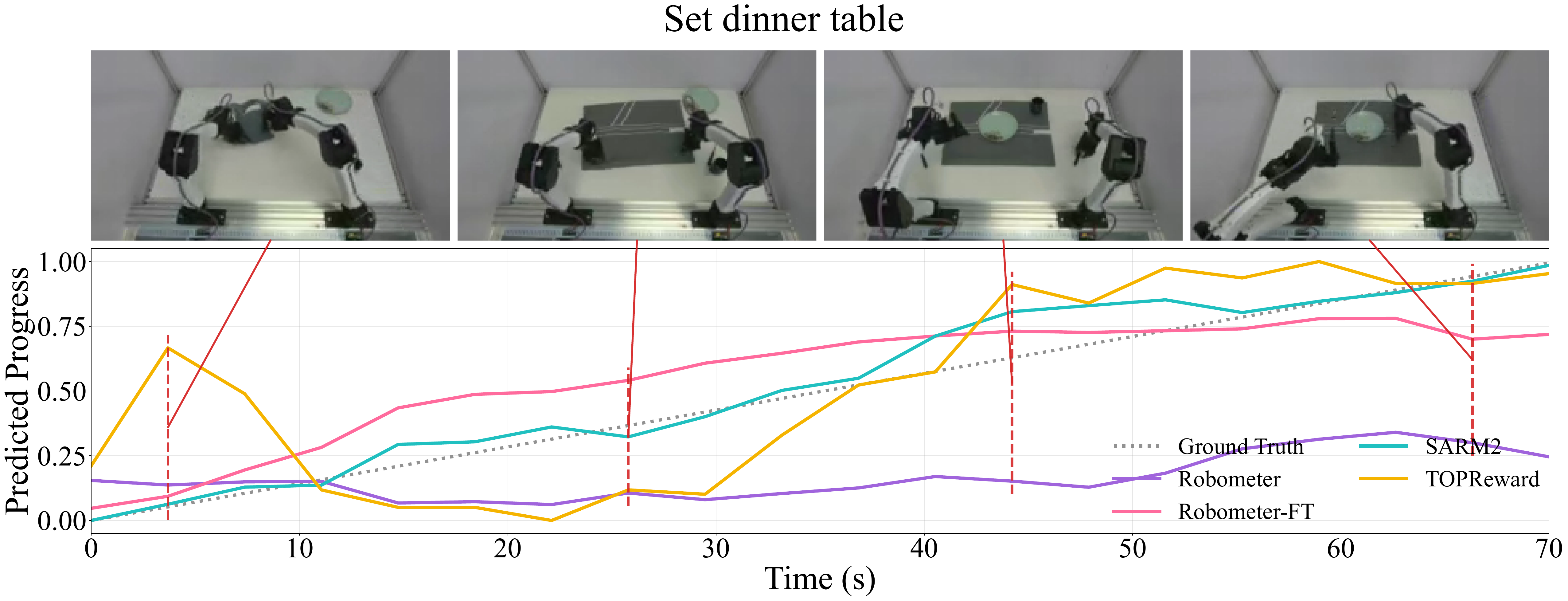}
    \hfill
    \includegraphics[width=0.48\linewidth]{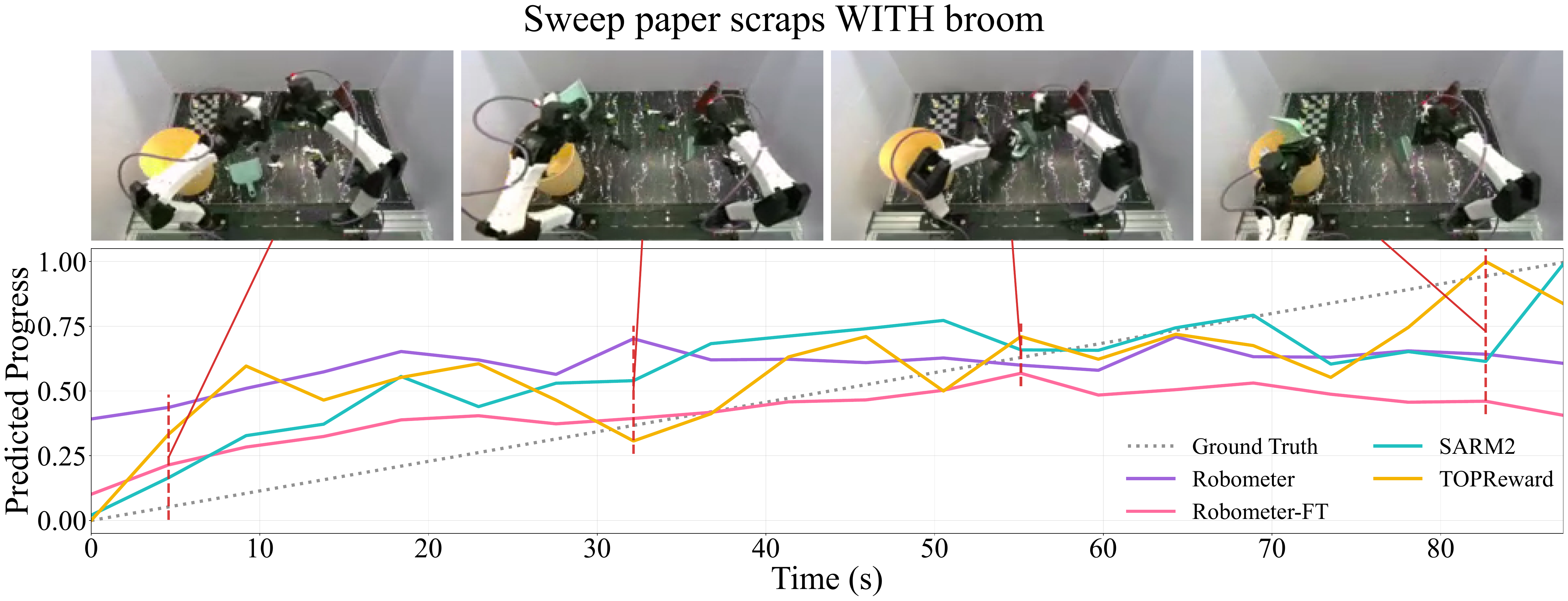}

    \includegraphics[width=0.48\linewidth]{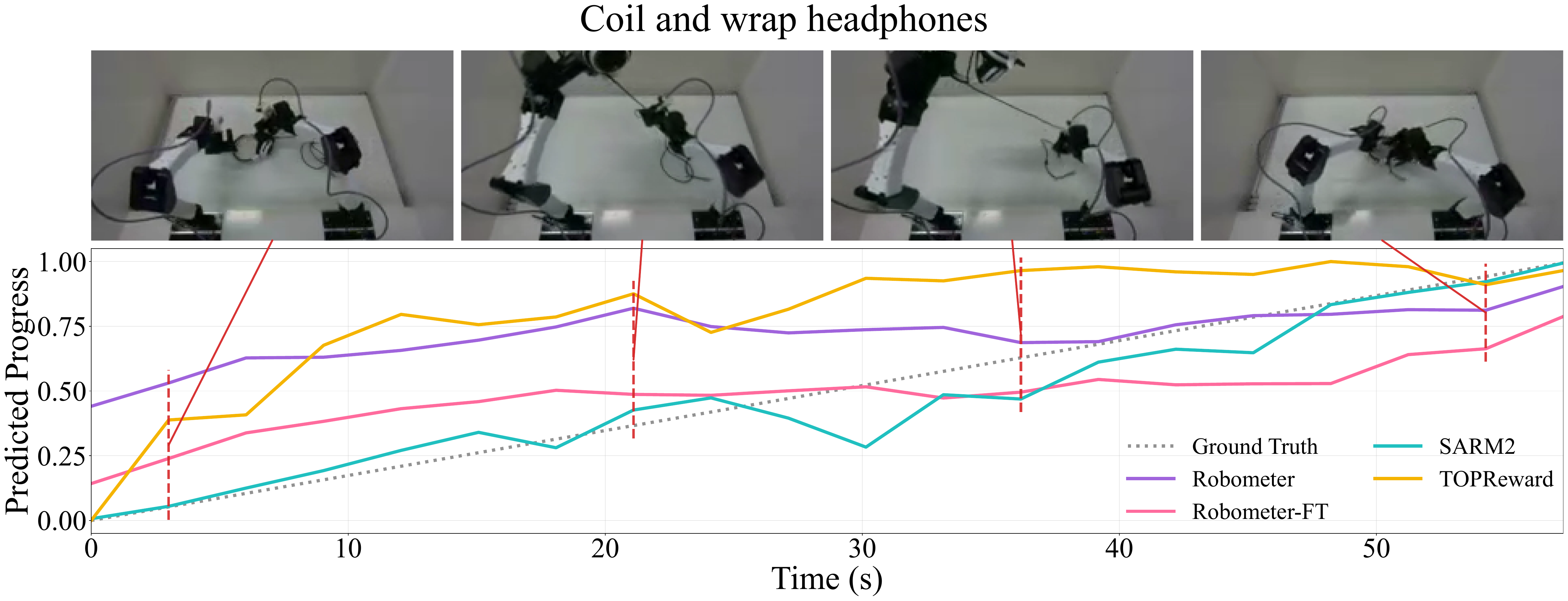}
    \hfill
    \includegraphics[width=0.48\linewidth]{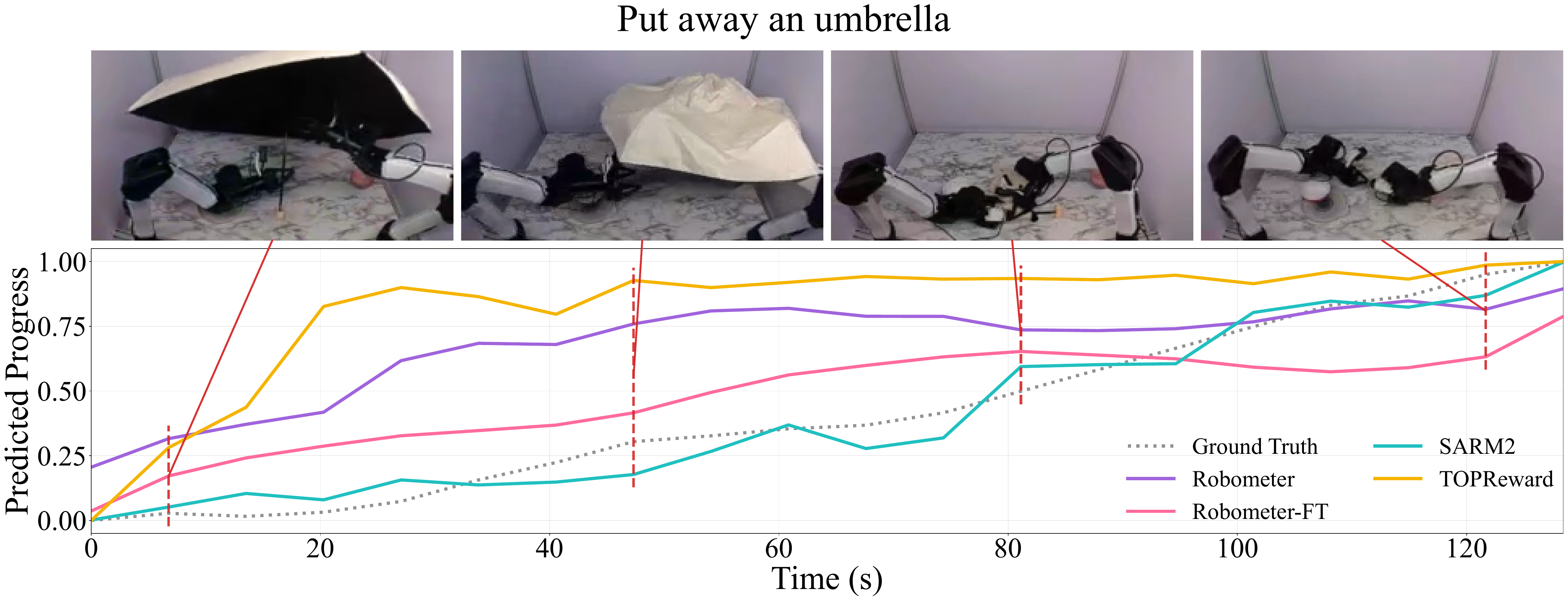}
    
    \caption{\textbf{Per-task progress estimates on held-out demos across all 10 benchmark tasks.} Each panel overlays the ground truth with predictions from TOPReward, Robometer, Robometer-FT, ReWiND, and \RM. The VLM baselines saturate near $1$ early (the over-optimism flagged in Section~\ref{sec:experiment_rm}); \RM{} closely tracks ground truth on both $\mathcal{S}_1$ and $\mathcal{S}_2$.}
    \label{fig:rm_est}
\end{figure}

We first examine how accurately each reward model estimates dense progress on held-out demonstrations. Figure~\ref{fig:rm_est} visualizes per-timestep progress predictions from TOPReward, Robometer, Robometer-FT, ReWiND, and \RM{} against the ground truth across all 10 benchmark tasks, spanning both the in-distribution split $\mathcal{S}_1$ and the held-out split $\mathcal{S}_2$. The two VLM-based baselines (TOPReward and Robometer) saturate near $1$ early in the trajectory, exhibiting the over-optimistic estimation discussed in Section~\ref{sec:experiment_rm}, while ReWiND and Robometer-FT track the ground truth more closely but still drift on longer-horizon tasks. \RM{} most faithfully follows the ground-truth progress curve across all 10 tasks, with tight agreement on both $\mathcal{S}_1$ and $\mathcal{S}_2$.





    

\begin{figure}[h!]
    \centering
    \includegraphics[width=\linewidth]{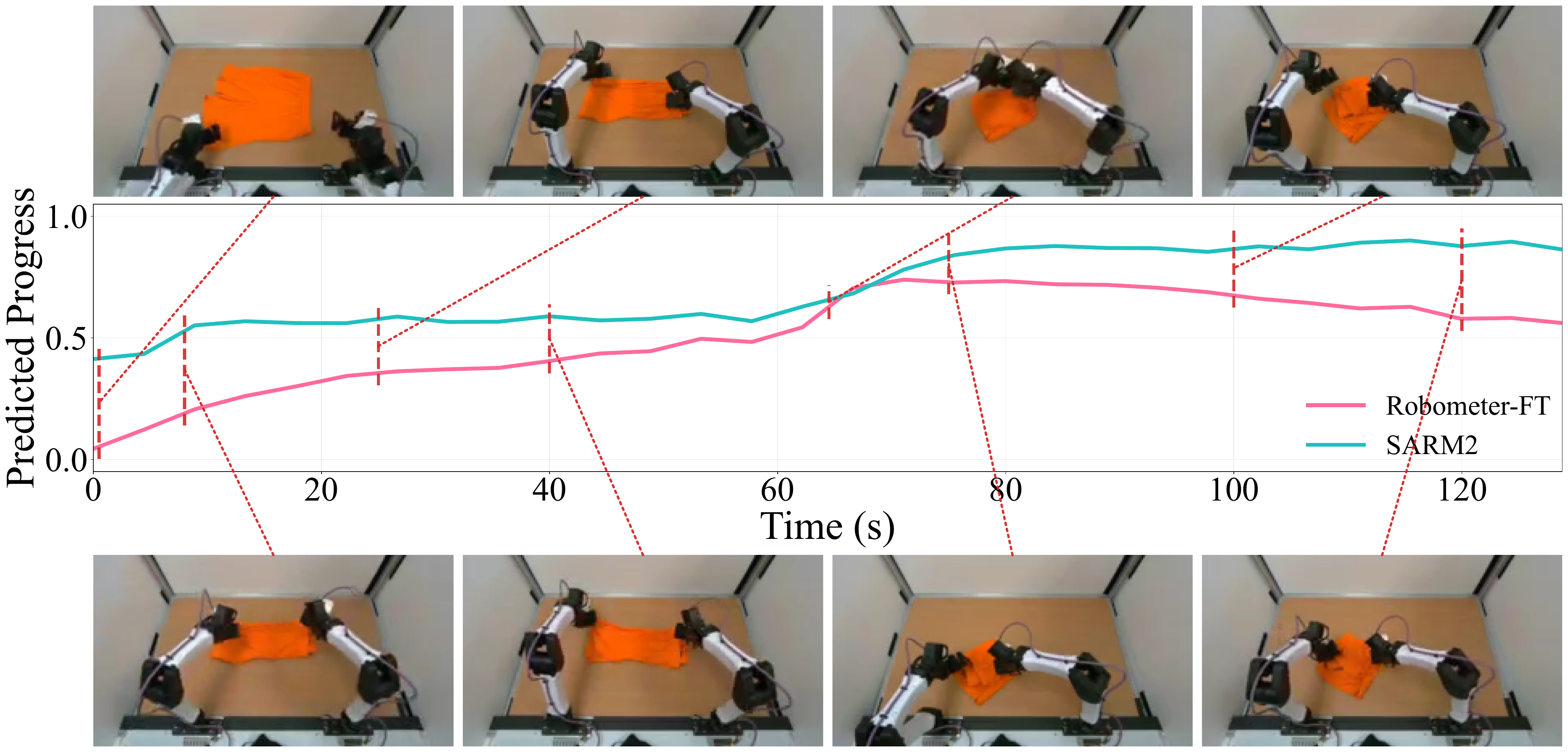}
    
    \vspace{0.5em}
    
    \includegraphics[width=\linewidth]{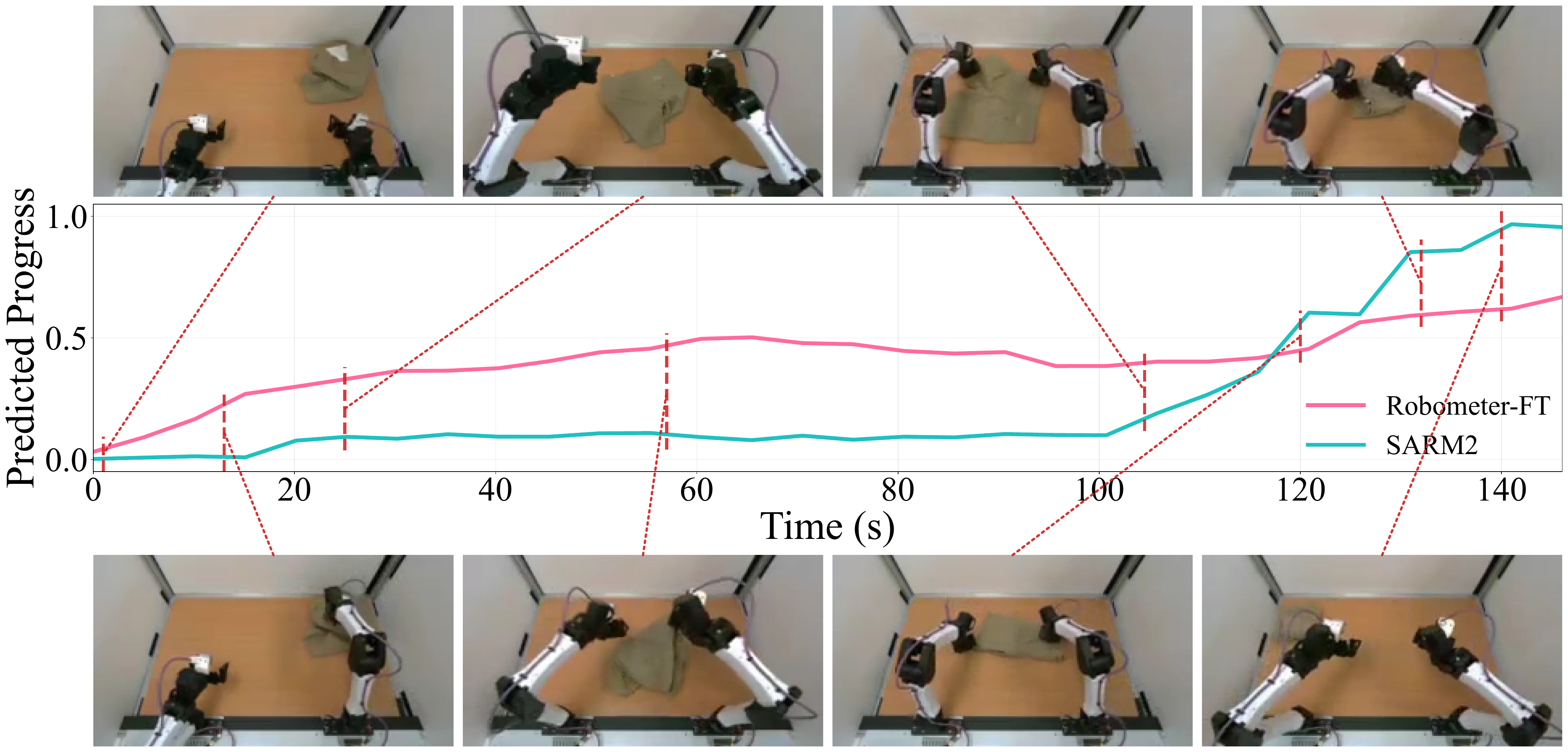}
    \caption{\textbf{Reward-model progress estimation on two Folding Shorts rollouts.} Each panel plots predicted progress vs.\ time for two reward models used in policy training. 8 key frames along the trajectory are demonstrated around the progress figure. \RM{} faithfully track the moments when policy making progress or struggling, whereas finetuned Robometer baseline did not catch those details.}
    \label{fig:rm_rollout_shorts}
\end{figure}

\begin{figure}[h!]
    \centering
    \includegraphics[width=0.85\linewidth]{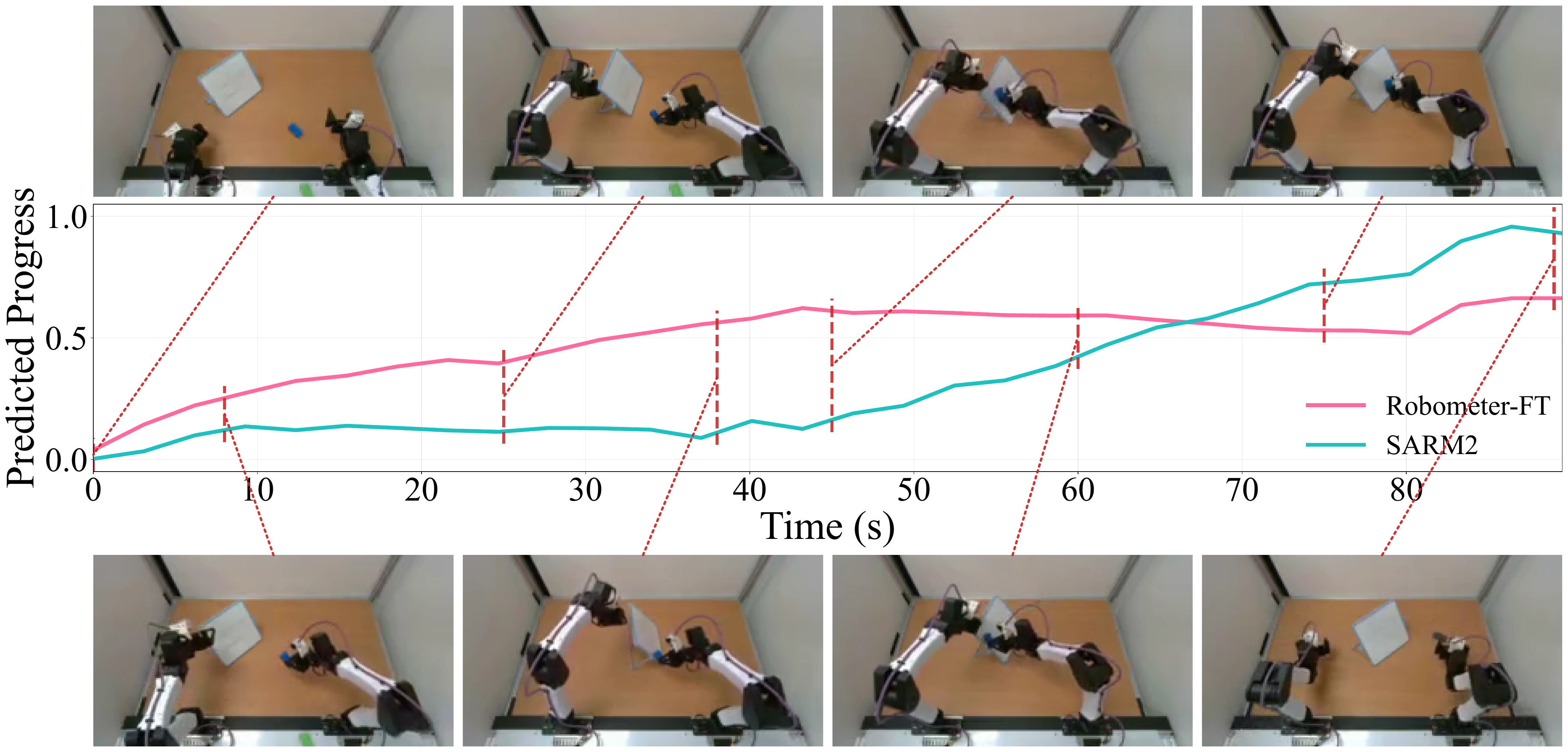}
    
    \vspace{0.5em}
    
    \includegraphics[width=0.85\linewidth]{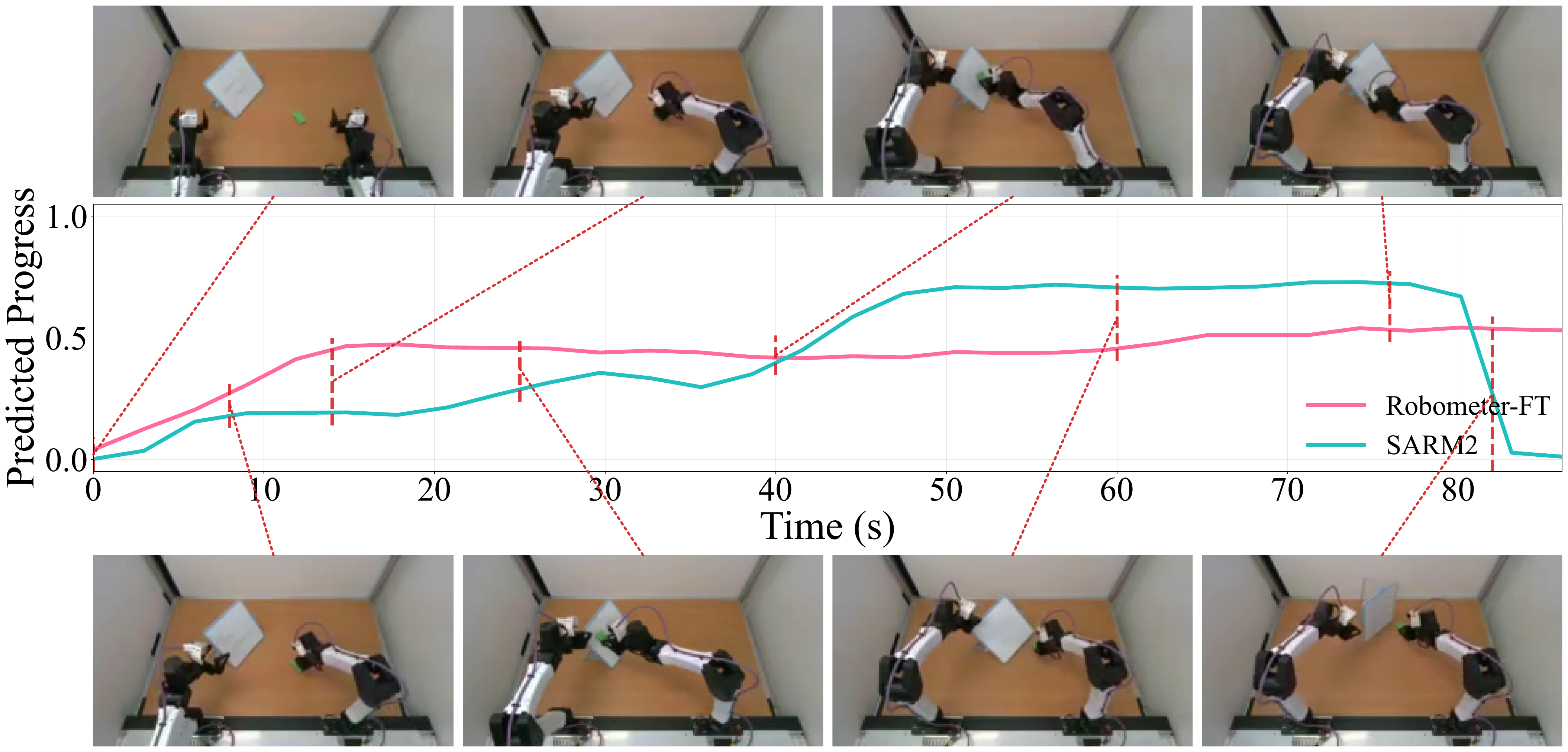}
    \caption{\textbf{Reward-model progress estimation on two Cleaning Whiteboard rollouts.} Each panel plots predicted progress vs.\ time for two reward models used in policy training. 8 key frames along the trajectory are demonstrated around the progress figure. \RM{} closely followed the situation of the robot station, including progress, adjusting, and even catastrophic failures, whereas finetuned Robometer baseline did not follow those details.}
    \label{fig:rm_rollout_white}
\end{figure}

Moreover, to probe how faithfully each reward model behaves under real policy training conditions, we visualize the predicted progress signal alongside key frames from on-policy rollouts on two qualitatively different tasks, \textit{Folding Shorts} (Figure~\ref{fig:rm_rollout_shorts}) and \textit{Cleaning Whiteboard} (Figure~\ref{fig:rm_rollout_white}). Across both tasks, \RM{} captures segment-level detail of the rollout: its predicted progress rises when the robot is making real progress, plateaus or dips when the robot is adjusting or struggling, and drops sharply at catastrophic failures, in agreement with the visualized key frames. In contrast, the finetuned Robometer baseline (also LoRA finetuned on labeled rollouts as \RM{}) misses these transitions, often emitting a rising progress signal while the robot is visibly struggling, or remaining flat through segments where the policy is in fact advancing the task. These gaps directly translate into downstream policy-training gaps: an incorrectly shaped value estimate produces wrong reward and advantage assignments, which feed noisy and sometimes sign-flipped gradients into the policy update and slow or destabilize self-improvement.

\vspace{-1.0em}

\subsection{Action Primitive Estimation and MoE Experts Selection Visualization Results}
\label{sec:act_pri_moe_plot}

\begin{figure}[h!]
    \centering
    \includegraphics[width=0.48\linewidth]{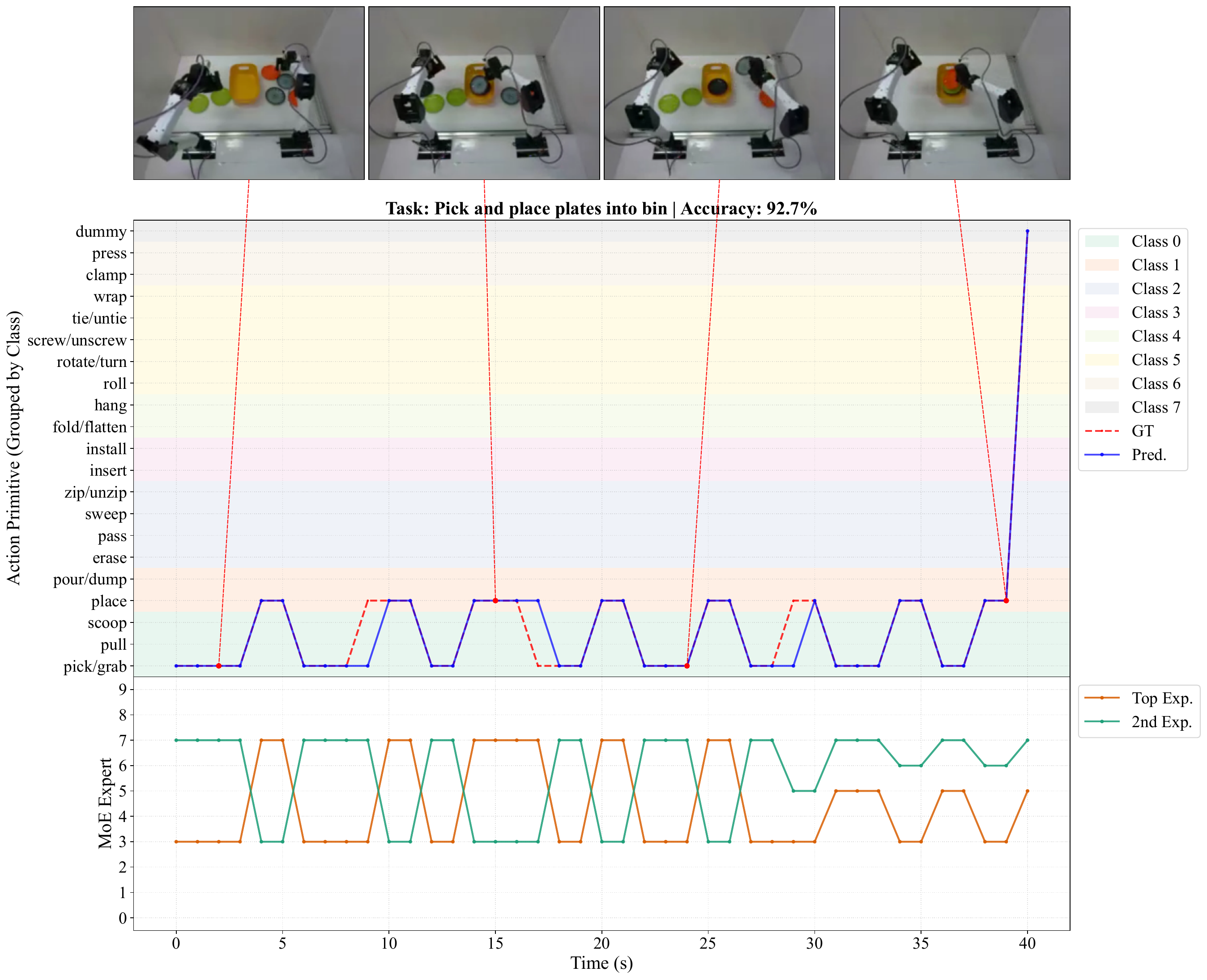}
    \hfill
    \includegraphics[width=0.48\linewidth]{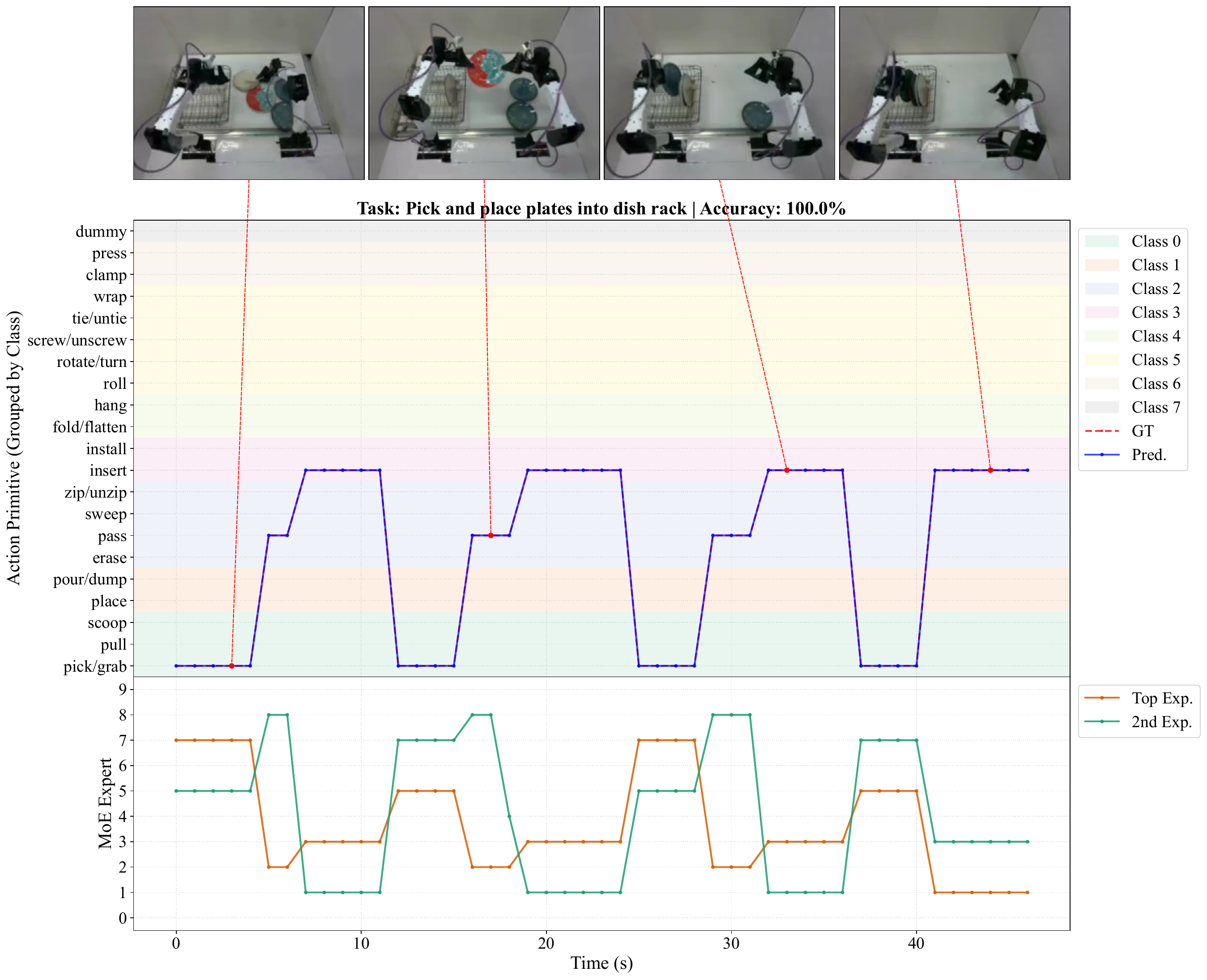}

    \includegraphics[width=0.48\linewidth]{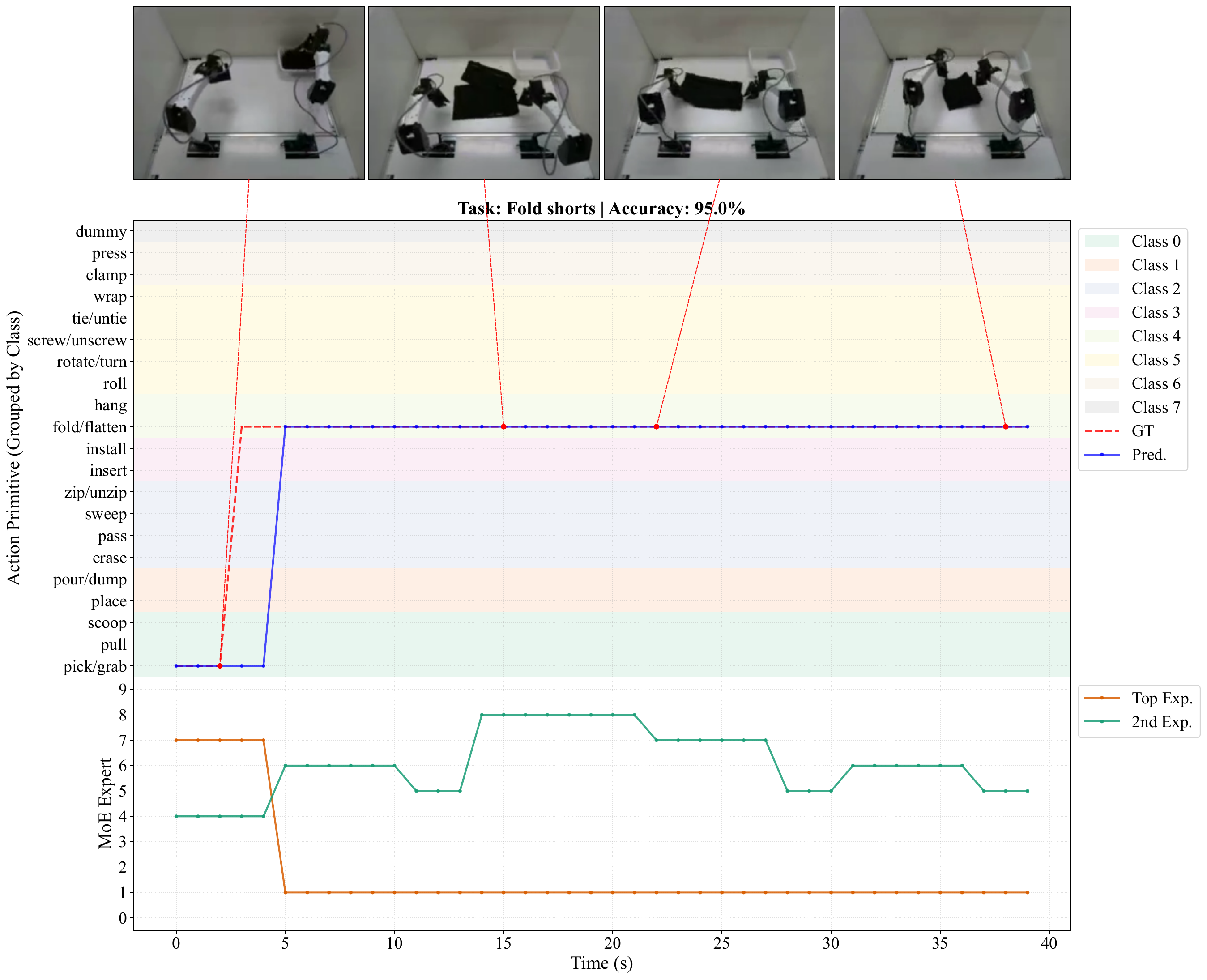}
    \hfill
    \includegraphics[width=0.48\linewidth]{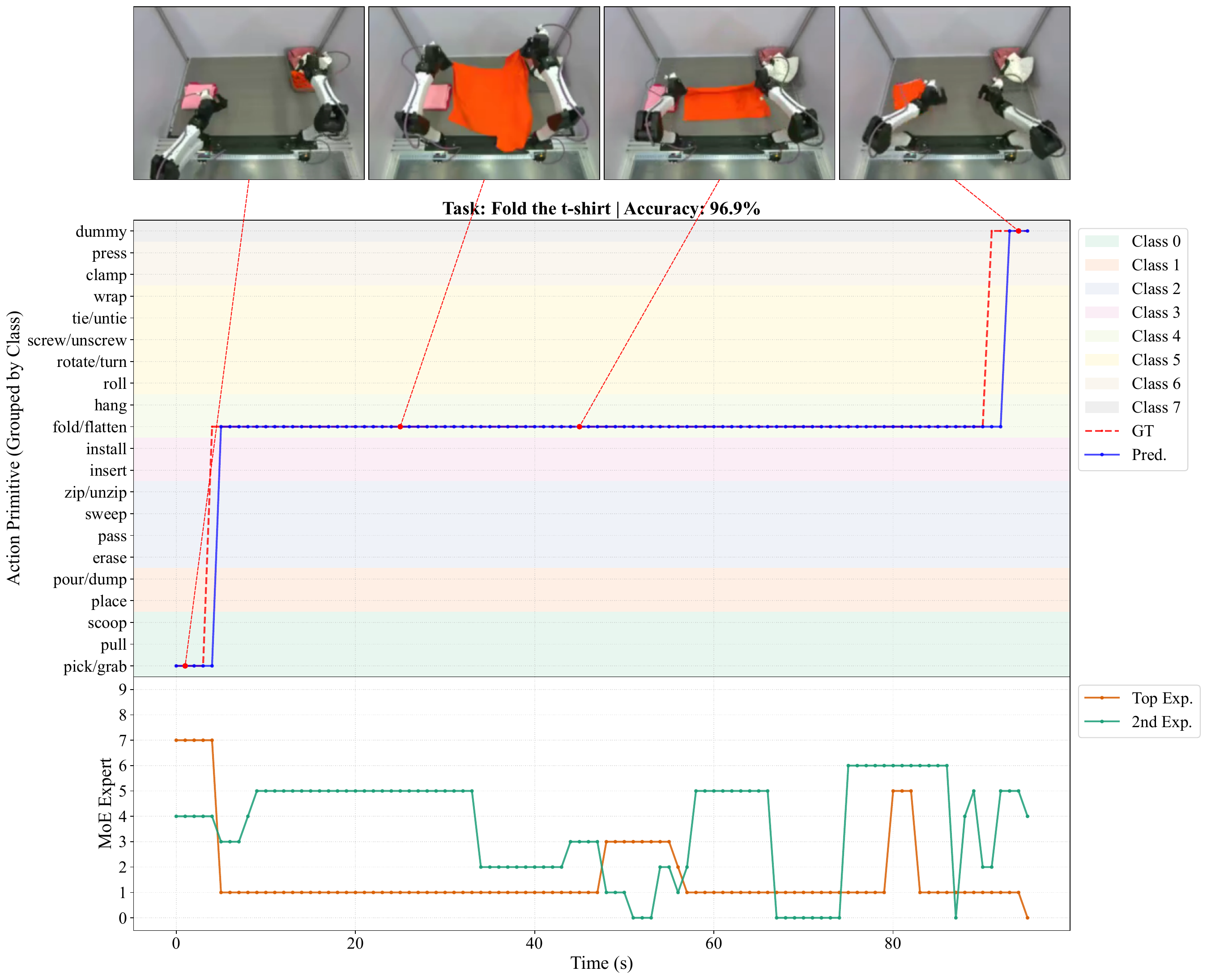}
    
    \includegraphics[width=0.48\linewidth]{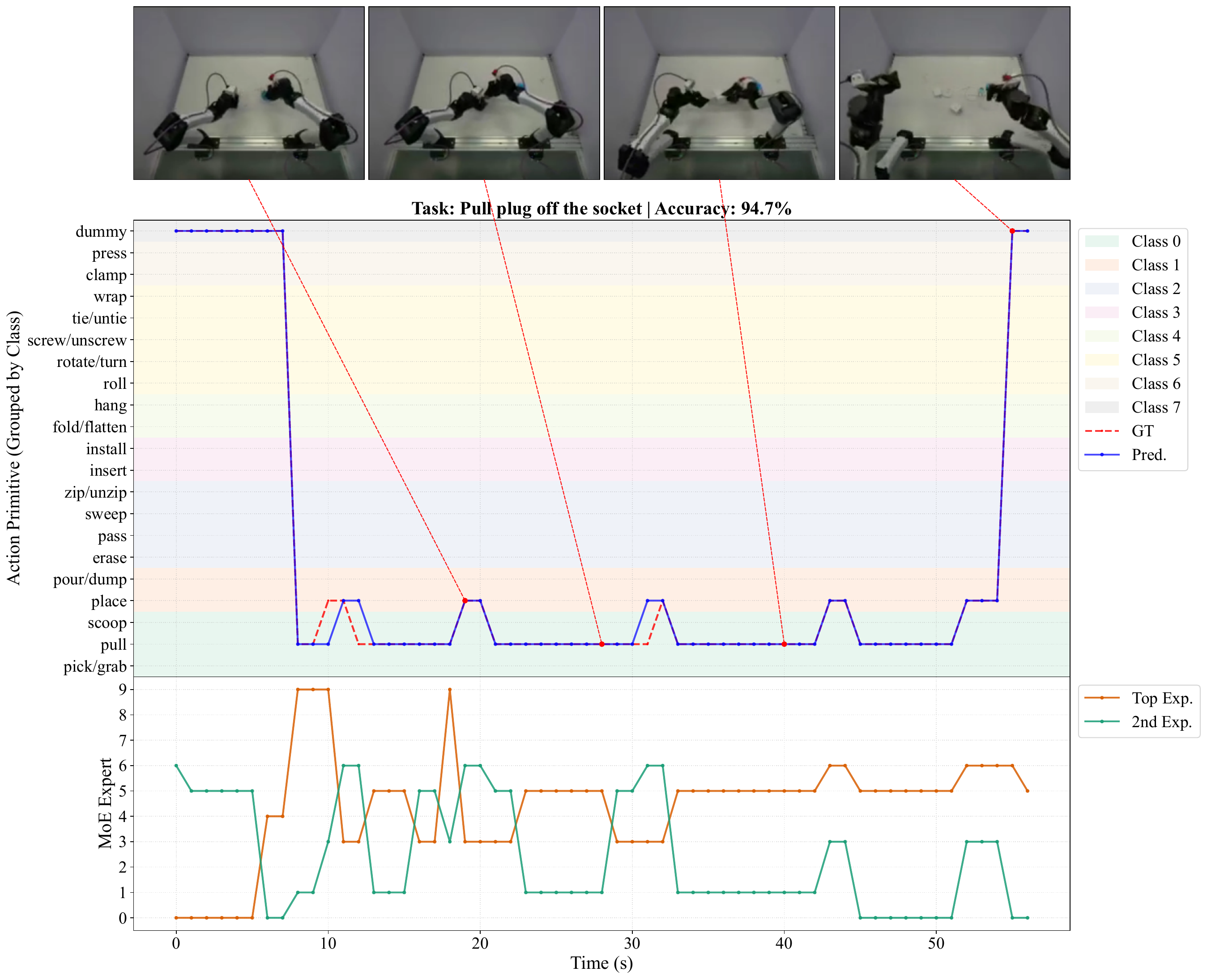}
    \hfill
    \includegraphics[width=0.48\linewidth]{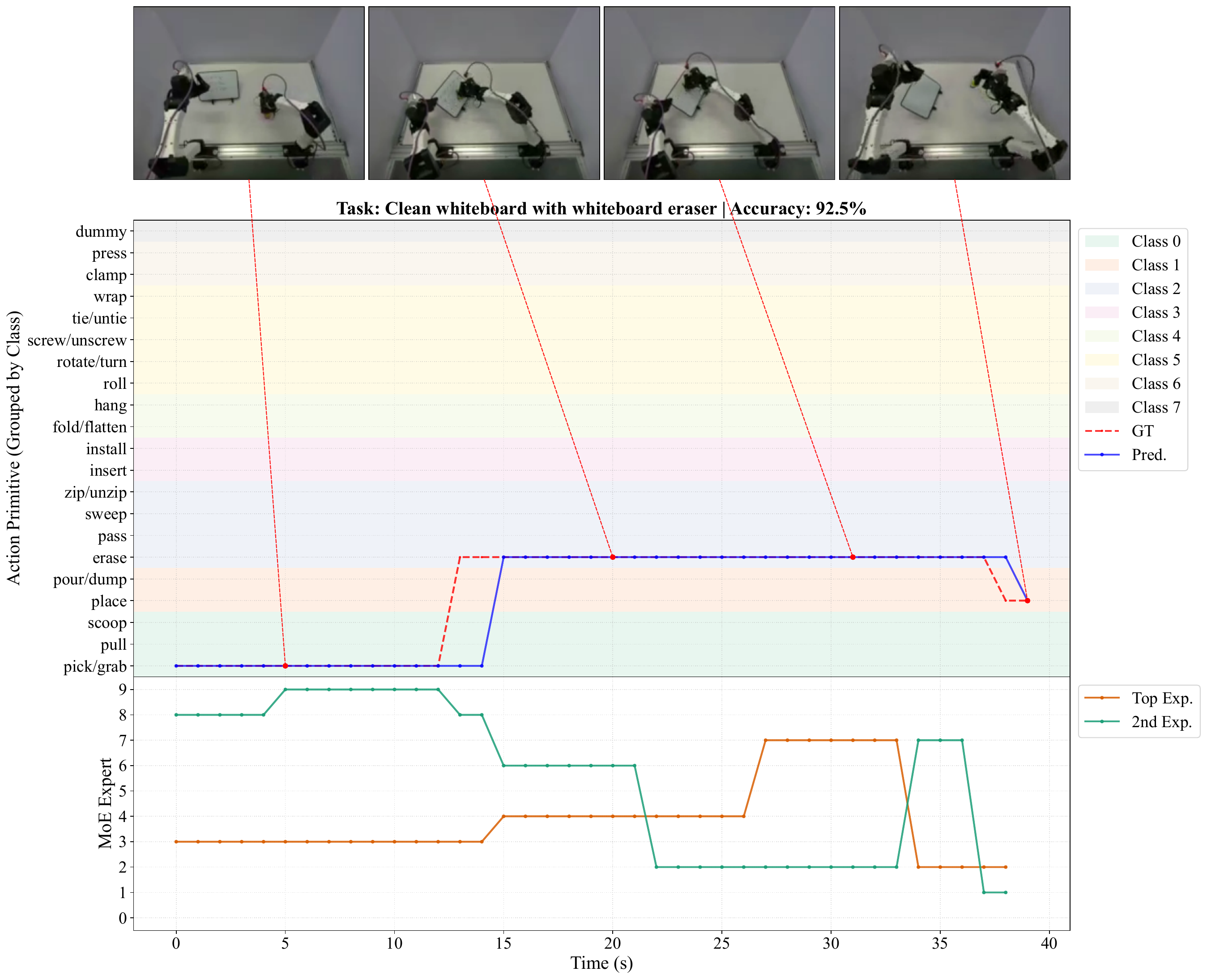}

    \includegraphics[width=0.48\linewidth]{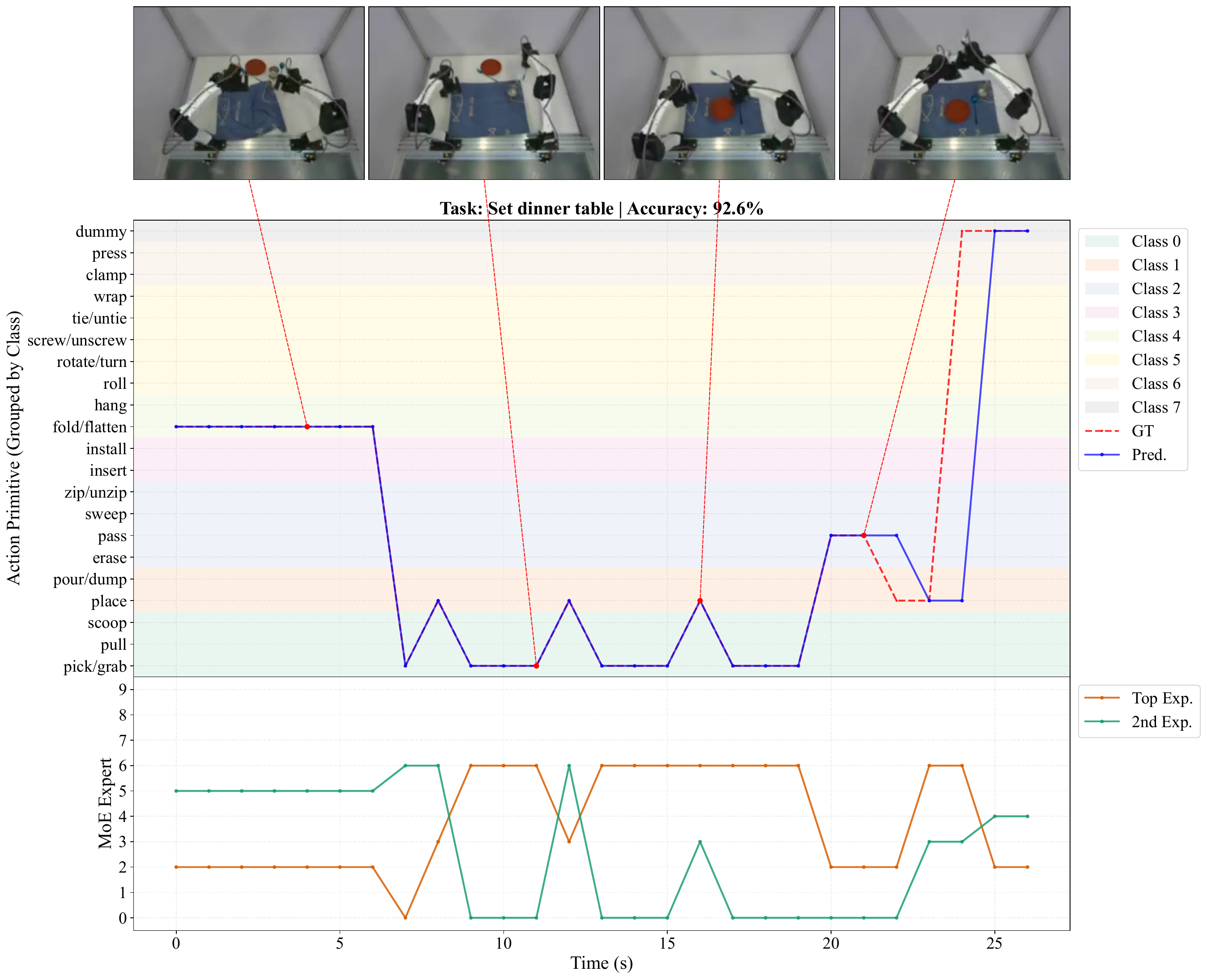}
    \hfill
    \includegraphics[width=0.48\linewidth]{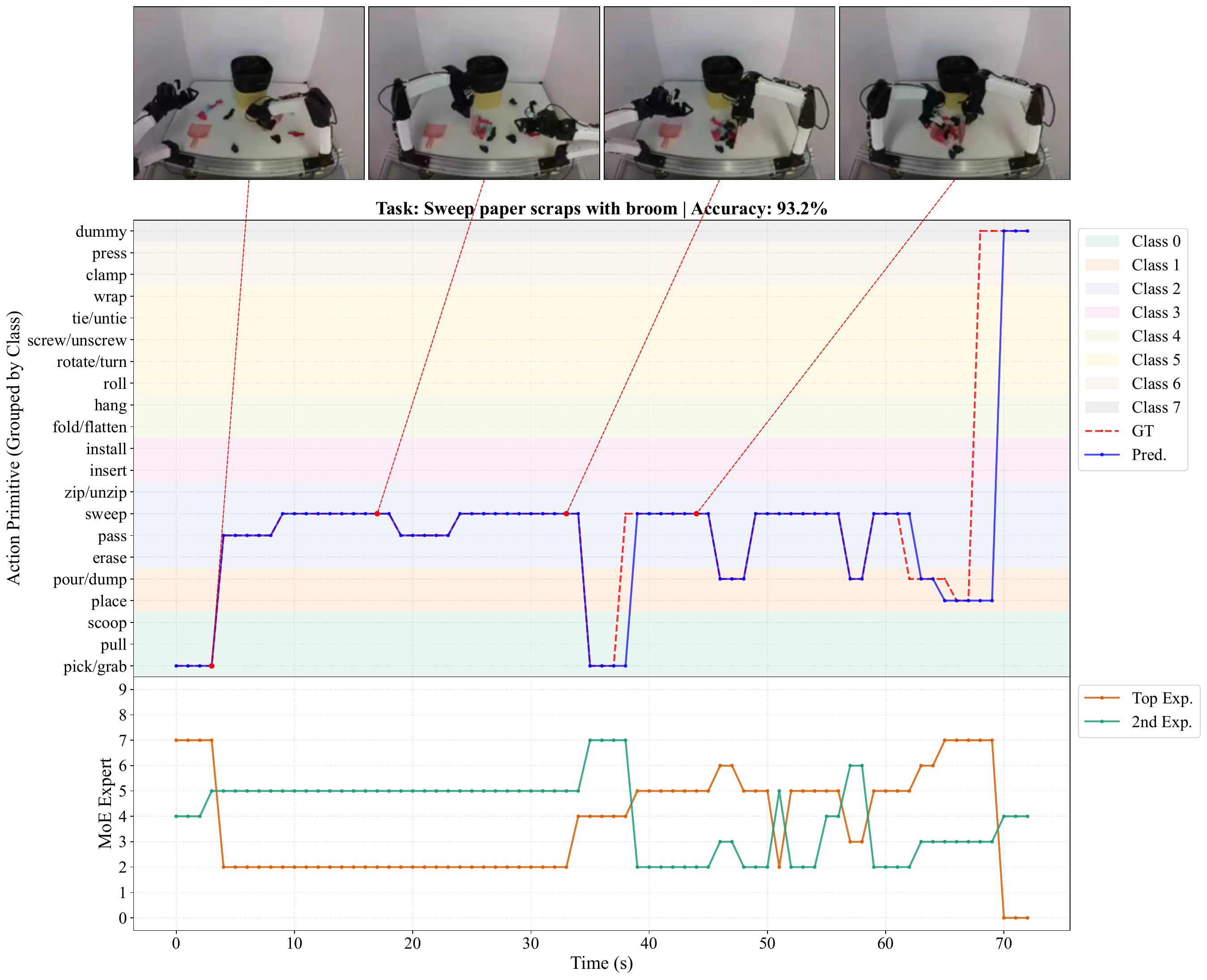}
    
    \caption{\textbf{Action-primitive predictions and MoE experts selection figures (part 1)} }
    \label{fig:action_primitive_main}
\end{figure}

\begin{figure}[h!]
    \centering
    \includegraphics[width=0.48\linewidth]{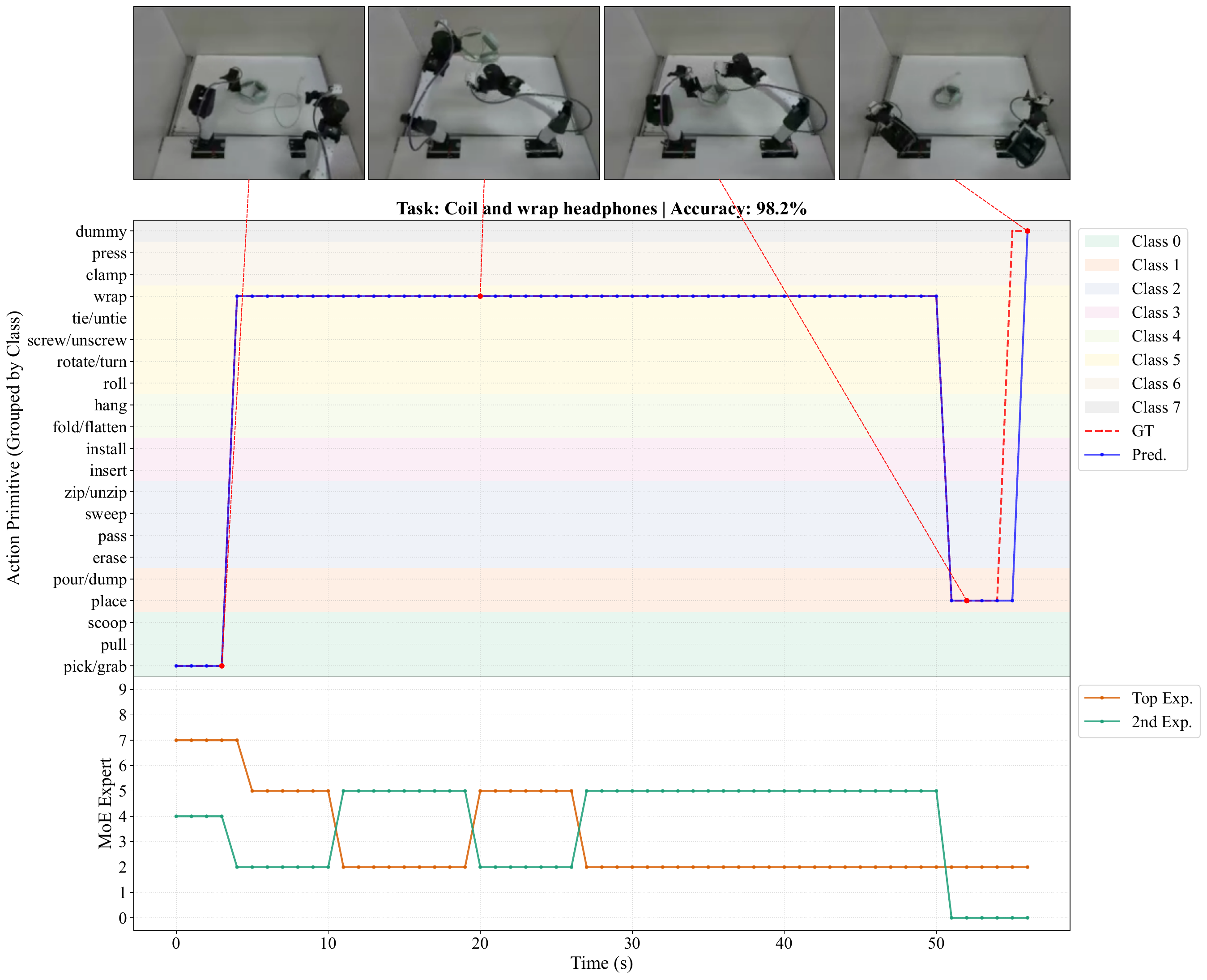}    
    \hfill
    \includegraphics[width=0.48\linewidth]{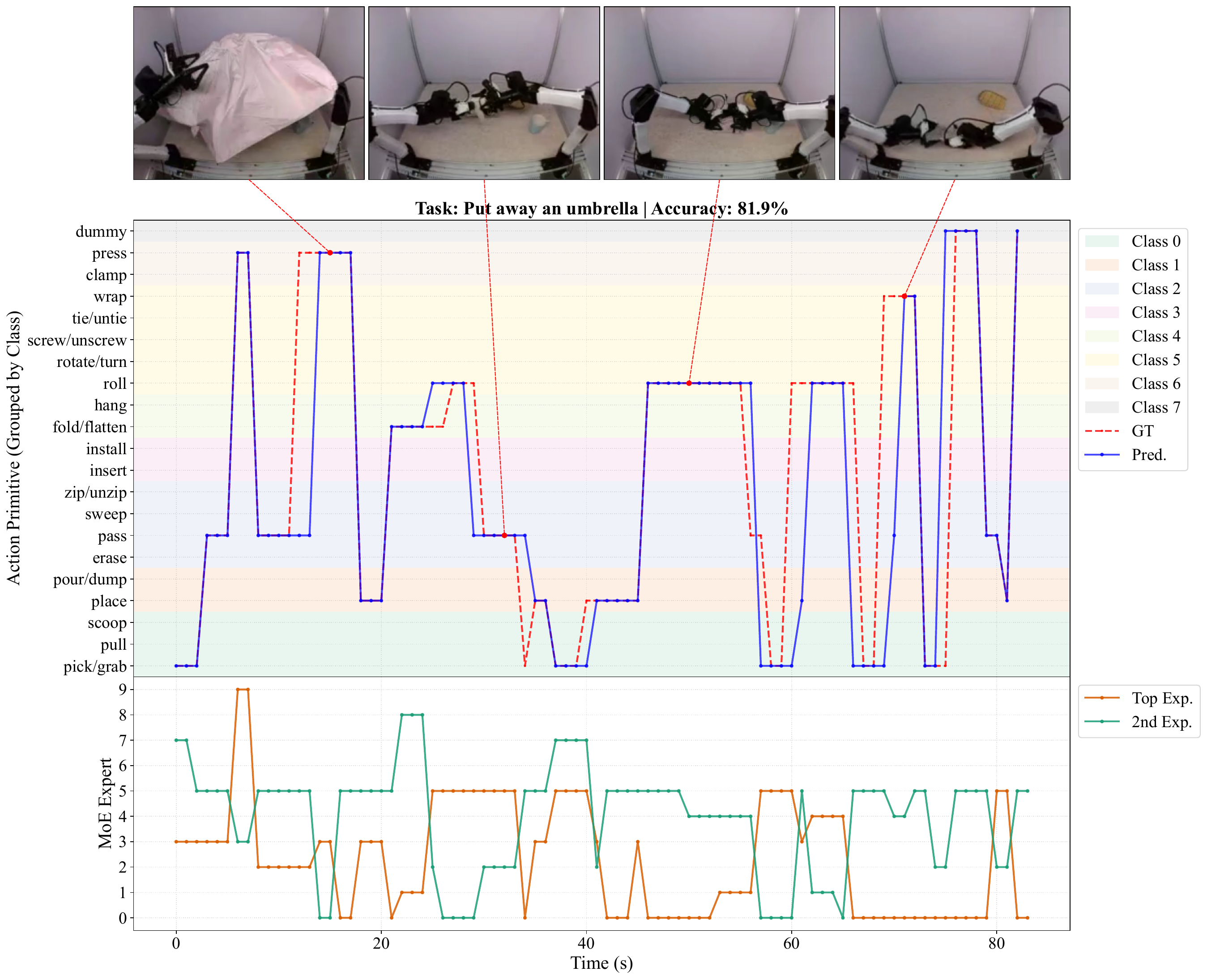}
    \caption{\textbf{Action-primitive predictions and MoE experts selection figures (part 2)} Each panel has four key frames along the trajectory (top), action primitive based stage estimator predictions v.s. ground truth (middle), and MoE experts selection (below); mid panel colors indicate primitive grouping as discussed in Appendix~\ref{tab:action_primitive_groups}.}
    \label{fig:action_primitive_extra}
\end{figure}

Our action-primitive-based stage estimator attains a micro-level average accuracy of $85.22\%$ aggregated over all timesteps; the per-task breakdown on each task's held-out test set (10 episodes per task) is reported in Table~\ref{tab:stage_estimator_accuracy}. Figures~\ref{fig:action_primitive_main} and~\ref{fig:action_primitive_extra} additionally show, for each task, one representative example comparing the predicted action primitive against the ground truth, together with the resulting MoE expert selection along the trajectory.

\vspace{-0.8em}

\begin{table}[h!]
    \small
    \centering
    \caption{\textbf{Stage estimator accuracy across the 10 benchmark tasks.} Tasks are listed in the same order as the per-task description in Appendix~\ref{sec:task_description}.}
    \label{tab:stage_estimator_accuracy}
    \begin{tabular}{lc}
        \toprule
        Task Name & Stage Estimator Accuracy $(\%)$  \\
        \midrule
        Pick and place plates into bin & 89.32 \\
        Pick and place plates into dish rack & 79.01 \\
        Folding shorts & 97.52 \\
        Folding the t-shirt & 98.27 \\
        Pull plug off the socket & 79.40 \\
        Clean whiteboard with whiteboard eraser & 96.22 \\
        Set dinner table & 70.85 \\
        Sweep paper scraps with broom & 91.15 \\
        Coil and wrap headphones & 97.72 \\
        Put away an umbrella & 57.74 \\
        \bottomrule
    \end{tabular}
\end{table}

\clearpage

\subsection{Auxiliary Formulae}
\label{sec:td_mc_derivation}


Let $\bar{p}_e^{(m)}$ denote the average routing probability assigned to expert $e$ by gate $m$ over a batch, and $f_e^{(m)}$ the fraction of tokens for which expert $e$ is selected in the top-$k$. Following~\citep{huang2024mentor},
\begin{equation}
    \mathcal{L}_{\text{balance}} = \sum_{m=0}^{M} E \cdot \sum_{e=1}^{E} f_e^{(m)} \cdot \bar{p}_e^{(m)},
    \qquad
    \mathcal{L}_{\text{entropy}} = -\sum_{m=0}^{M} \sum_{e=1}^{E} \bar{p}_e^{(m)} \log \bar{p}_e^{(m)}.
\end{equation}
Balance encourages uniform expert utilization within each gate; entropy prevents routing distributions from becoming overly peaked early in training.

\subsection{Additional Discussion on SPIRAL}
\label{sec:rl_design_choice}

\paragraph{Reward model reusability.} A single reward model can label demonstrations across multiple downstream tasks and serve as the starting point for rollout-based fine-tuning, amortizing the cost of reward model training across the task suite.

\paragraph{Mixed objective for RL} To motivate this hybrid design, we briefly contrast the two objectives. Given a critic $Q_\phi(s, a)$ trained on transitions from replay buffer $\mathcal{D}$, the TD objective bootstraps from the next-state value, while the MC objective regresses onto the empirical discounted return:
\begin{equation}
\begin{aligned}
    y_t^{\text{TD}} &= r_t + \gamma \, Q_{\phi'}(s_{t+1}, a_{t+1}), \qquad
    & J_{\text{TD}}(\phi) &= \mathbb{E}_{\mathcal{D}}\!\left[ \big(Q_\phi(s_t, a_t) - y_t^{\text{TD}}\big)^2 \right], \\
    y_t^{\text{MC}} &= \sum_{k=t}^{T} \gamma^{k-t} r_k, \qquad
    & J_{\text{MC}}(\phi) &= \mathbb{E}_{\mathcal{D}}\!\left[ \big(Q_\phi(s_t, a_t) - y_t^{\text{MC}}\big)^2 \right].
\end{aligned}
\end{equation}
The TD target depends only on the immediate reward $r_t$ and a one-step bootstrap, so when $r_t$ is dense and locally informative the critic receives a precise learning signal at every step. This makes TD particularly well-suited to long-horizon tasks: each transition contributes gradient, and credit assignment does not need to propagate across hundreds of steps from a distant terminal signal. The MC target, in contrast, regresses the critic onto the empirical discounted return. Under sparse terminal reward ($r_k = 0$ for $k < T$, $r_T = R_{\text{sparse}}$), this collapses to $y_t^{\text{MC}} = \gamma^{T - t} R_{\text{sparse}}$: the discount $\gamma^{T-t}$ exponentially attenuates terminal credit with episode length, so successful-and-fast episodes produce strictly larger targets than successful-but-slow ones. The MC objective therefore expresses a built-in, steady preference for short successful trajectories. The two objectives are complementary: TD drives fine-grained shaping of subtle action choices throughout the trajectory, while MC imposes a stable global preference for fast successful episodes and lets the policy inherit their motion style. We combine them as
\begin{equation}
    J_{\text{critic}}(\phi) = J_{\text{TD}}(\phi) + \alpha \cdot J_{\text{MC}}(\phi),
\end{equation}
where $\alpha$ balances local shaping against global preference.

To validate this hybrid design, we ablate the two reward components in the offline RL setting, comparing the full mixed objective against variants that use either the dense-reward TD term or the sparse-reward MC term alone on both tasks (Table~\ref{tab:dense_only_ablation}).

\begin{table}[t]
\small
\centering
\caption{\textbf{Ablation of the critic objective in the offline RL stage.} We compare behavior cloning (BC) against offline RL trained with the sparse-reward MC term only, the dense-reward TD term only, and our full mixed objective ($J_{\text{TD}} + \alpha\,J_{\text{MC}}$), on Folding Shorts (from flat and from crumble) and Cleaning Whiteboard. We report success rate (SR), average progress (Avg. Prog.), and progress gain over BC (Prog. Gain). The mixed objective achieves the best results on both tasks, confirming that the dense TD and sparse MC terms are complementary.}
\label{tab:dense_only_ablation}
\begin{tabular}{lcccccc}
\toprule
{\multirow{2}{*}{\textbf{Training Methods}}}
& \multicolumn{2}{c}{\textbf{Folding Shorts (SR$\uparrow$)}} 
& \multicolumn{3}{c}{\textbf{Cleaning Whiteboard}} \\
\cmidrule(lr){2-3} \cmidrule(lr){4-6}
{}
& \textbf{Flat} 
& \textbf{Crumble} 
& \textbf{SR$\uparrow$} 
& \textbf{Avg. Prog. $\uparrow$}
& \textbf{Prog. Gain $\uparrow$} \\
\midrule
BC         & 1/12  & 0/12 & 4/20  & 0.475 & N/A \\
RL-Sparse (MC Only)    & 4/12  & 2/12 & 6/20  & 0.625 & 0.150\\
RL-Dense (TD Only)    & 5/12  & 4/12 & 9/20  & 0.745 & 0.270\\
RL-Dense (mixed obj.)     & \textbf{7/12}  & 4/12 & \textbf{10/20} & \textbf{0.813} & \textbf{0.338} \\
\bottomrule
\end{tabular}
\end{table}

\begin{figure}
    \centering
    \includegraphics[width=\linewidth]{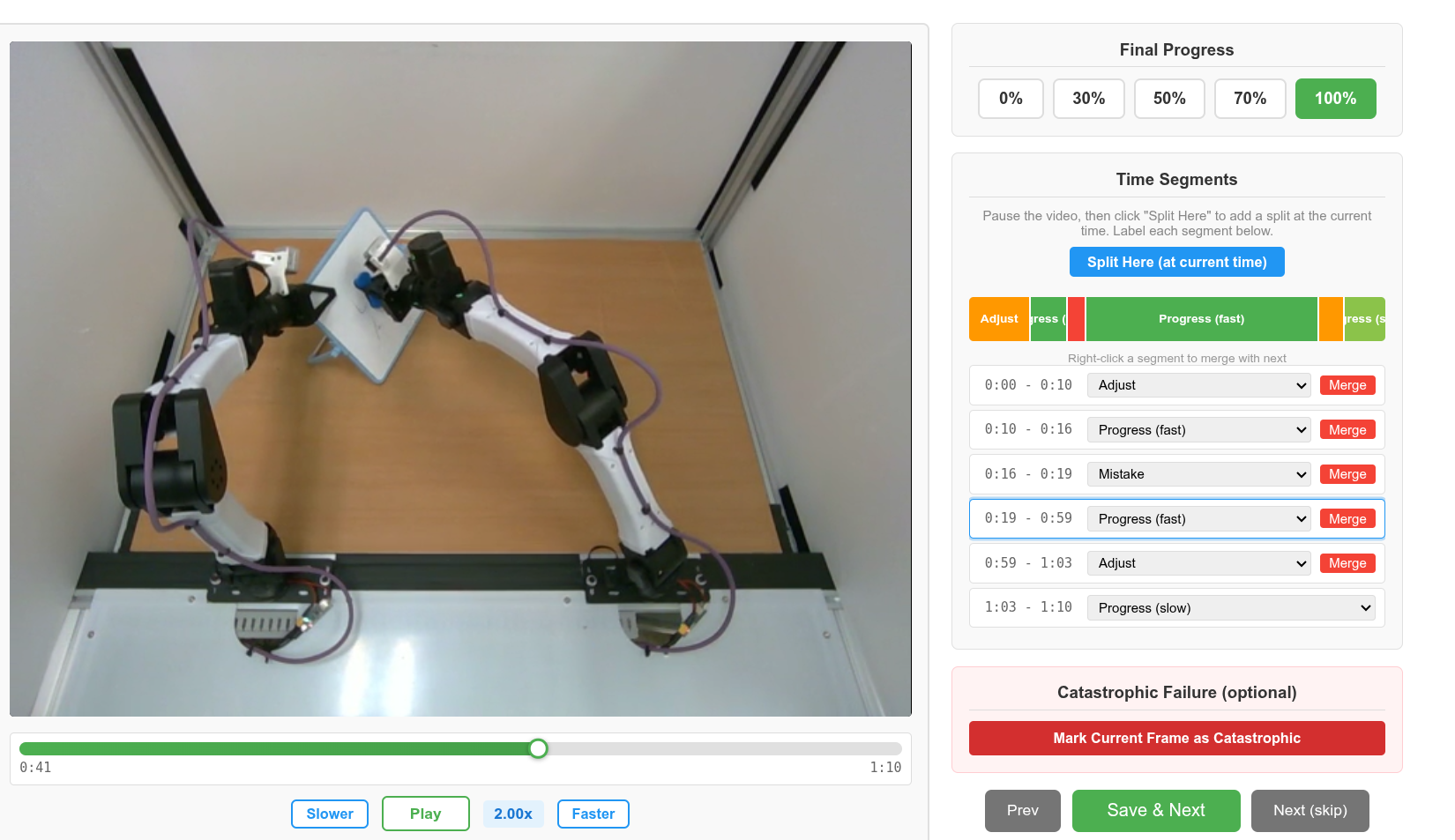}
    \caption{\textbf{Annotation interface for one-time reward-model adaptation (Stage~3).} The annotator segments a rollout into chunks and labels each with one of \{\texttt{fast progress}, \texttt{slow progress}, \texttt{adjust}, \texttt{mistake}\} plus a final progress value. Annotating $\sim$100 rollouts of $\pi_1$ takes 2--3 hours per task.}
    \label{fig:rollout_label}
\end{figure}

\paragraph{One-time human labeling effort.} Reward model adaptation (Stage 2) is a one-time effort. We deliberately collect $\mathcal{R}_1$ from the BC policy $\pi_1$ — the weakest policy in the pipeline — because its rollouts surface the most diverse out-of-distribution (OOD) failure modes, which are precisely the states the reward model must learn to score correctly. We collect around 100 rollouts for $\mathcal{R}_1$, segment them, and label each segment with one of \{\texttt{fast progress}, \texttt{slow progress}, \texttt{adjust}, \texttt{mistake}\}, together with a final progress value derived from our evaluation protocol. The annotation interface is shown in Figure~\ref{fig:rollout_label}. Based on our experience, it took 2 to 3 hours labeling 100 rollouts for both tasks. To prevent overfitting to the small adaptation set, $\mathcal{R}_1^{\text{labeled}}$ is mixed with a 50\% subsample of the original task-specific reward model training data; the resulting fine-tuning run requires roughly 1/10 of the pretraining compute. 

\paragraph{Minimal human involvement.} Among self-improvement pipelines, ours imposes the least demanding human role. Once $\mathrm{RM}_2$ is obtained, the SPIRAL loop runs without any further human labeling: an operator is needed only for environment resets and safety supervision, and we collect just 50--100 rollouts per iteration, with 2--3 iterations sufficient to robustly solve the target tasks (see Figure~\ref{fig:self_improve_trend}). This contrasts sharply with intervention-based methods such as $\pi_{0.6}$~\citep{intelligence2025pi}, where human-in-the-loop corrections must be supplied \textit{online}, at exactly the right moment, with precise and decisive motions, demanding annotators who are simultaneously expert teleoperators and intimately familiar with the policy's failure modes. Our framework instead requires only \textit{offline}, researcher-level annotation of an initial rollout set, a one-time effort that imposes no real-time skill requirements on the human and is far easier to scale.

\begin{figure}
    \centering
    \includegraphics[width=\linewidth]{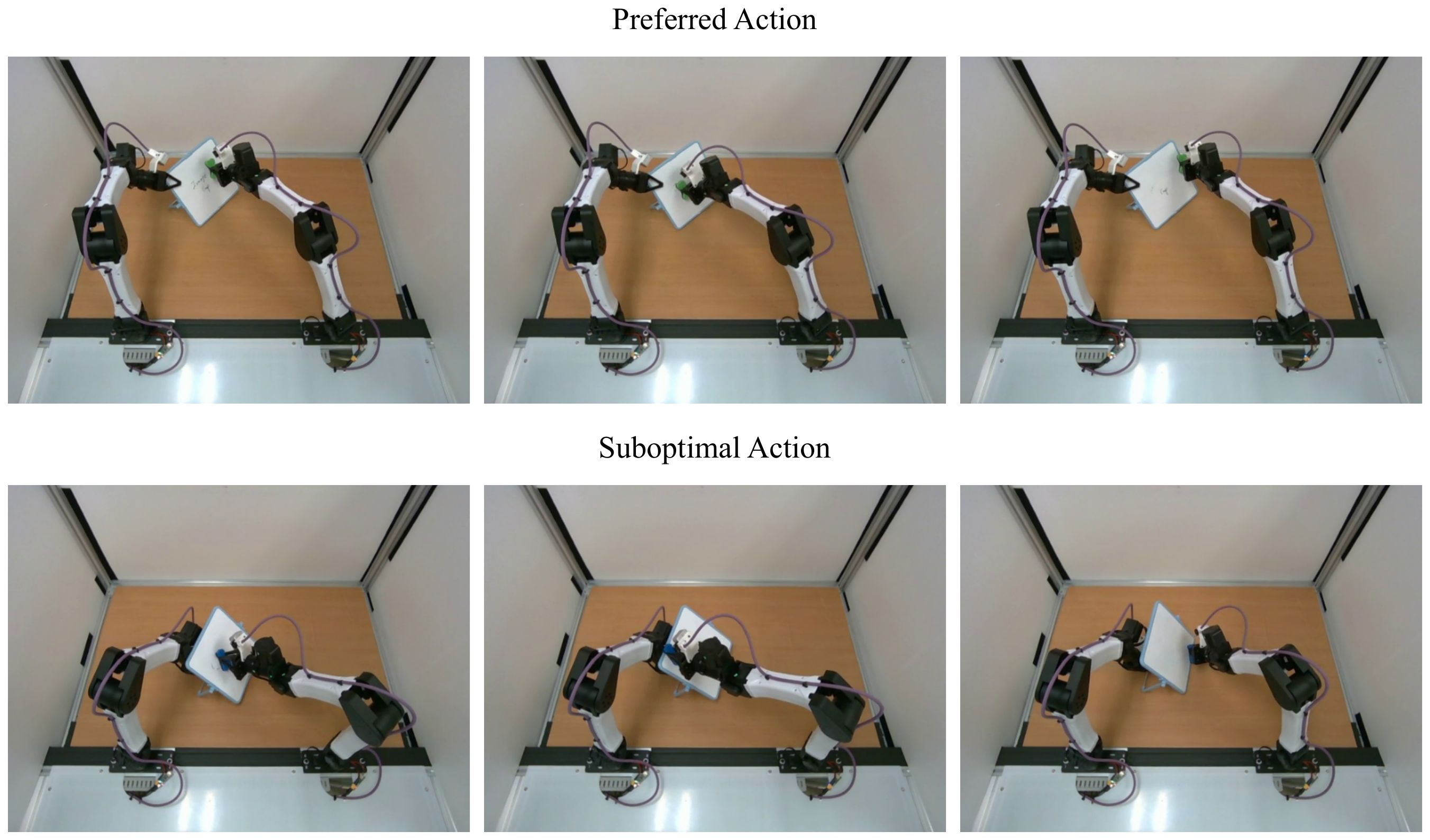}
    \caption{Comparison of high efficiency action from policy after SPIRAL loop (top) and suboptimal action from init BC policy (below).}
    \label{fig:act_pref}
\end{figure}

\paragraph{Beyond error correction.} Labeling rollouts does more than fix erroneous behaviors and recover from OOD states: it implicitly steers the policy toward the most efficient and stable solution strategies, aligning execution style with human preferences. Figure~\ref{fig:act_pref} illustrates this on Cleaning Whiteboard. The top frames show the canonical strategy: the left arm grasps the top corner of the board to hold it rigid, while the right arm wipes with the eraser under a stable contact. The bottom frames show a failure mode of the initial BC policy: the left arm instead grips the lower bar of the stand, an unstable support such that every wiping stroke tilts the board and lets it spring back, making cleaning slow and error-prone. This grasp does not appear in the demonstration set; it is an OOD behavior. Strictly speaking, it is not a mistake. Progress can still increase (the board does get cleaner), but the motion is inefficient and risks toppling the whiteboard entirely. We therefore label such segments as ``adjust'' with zero reward, injecting human preference and high-level prior knowledge into $\mathrm{RM}_2$ via fine-tuning. After the SPIRAL loop, the policy distills this preference and no longer adopts the unstable grasp. 

\paragraph{Imperfect results on Folding Shorts from crumble.}
After three rounds of SPIRAL, our policy achieves a substantially higher success rate than vanilla BC and offline RL ($8/12$ vs.\ $0/12$ vs.\ $4/12$). However, this result still shows a clear gap compared with the nearly saturated performance on Folding Shorts from flat ($12/12$) and Cleaning Whiteboard ($18/20$), and is also lower than recent SOTA cloth-folding VLA policies such as X-VLA~\citep{wen2025dexvla}. We attribute this limitation to several factors. First, our policy is trained with only 20 hours of demonstration data, which is a low-data regime for cloth folding. Moreover, the demonstrations are not filtered and have varied quality; as a rough proxy, the time required to fold one pair of shorts varies from 40s to 100s across episodes. The data also contains subtle modality differences because it is collected by different operators, who may have different action preferences in some steps, such as the grasping position used during folding. Second, compared with Folding Shorts from flat or Cleaning Whiteboard, Folding Shorts from crumble contains a more challenging and less structured stage: flattening the shorts. During this stage, the robot often needs multiple attempts to make the shorts fully flat. Due to the deformability of soft cloth and the diversity of cloth textures, the shorts can appear in a large number of possible states. As a result, the current reward model does not have sufficient granularity and accuracy to reliably judge whether the cloth becomes flatter after each attempt,  especially for the first few attempts when the shorts are still highly crumpled. This causes the reward signal to remain flat or become noisy during this stage. Furthermore, many subtle human folding skills are difficult to label with our current method, as shown in Figure~\ref{fig:rollout_label}. These skills are therefore hard to distill into the downstream policy through SPIRAL. We recognize this as an important limitation of our method: for highly complex tasks with large state variation, the most effective post-training strategy may still require learning from a large amount of high-quality demonstration data collected under a unified protocol and modality.

\vspace{-1.0em}

\subsection{Policy Rollout Examples}
\label{sec:policy_rollout}
\begin{figure}[h!]
    \centering
    \includegraphics[width=0.9\linewidth]{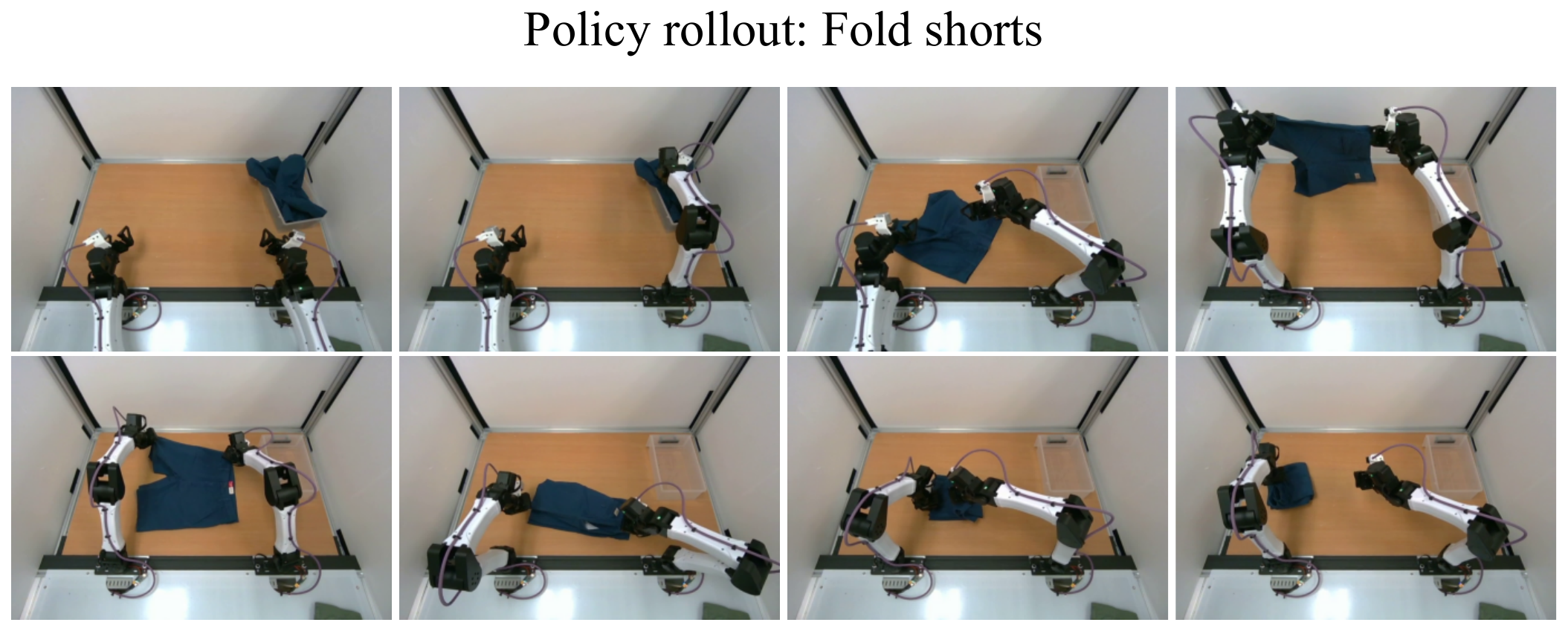}

    \vspace{0.5em}
    
    \includegraphics[width=0.9\linewidth]{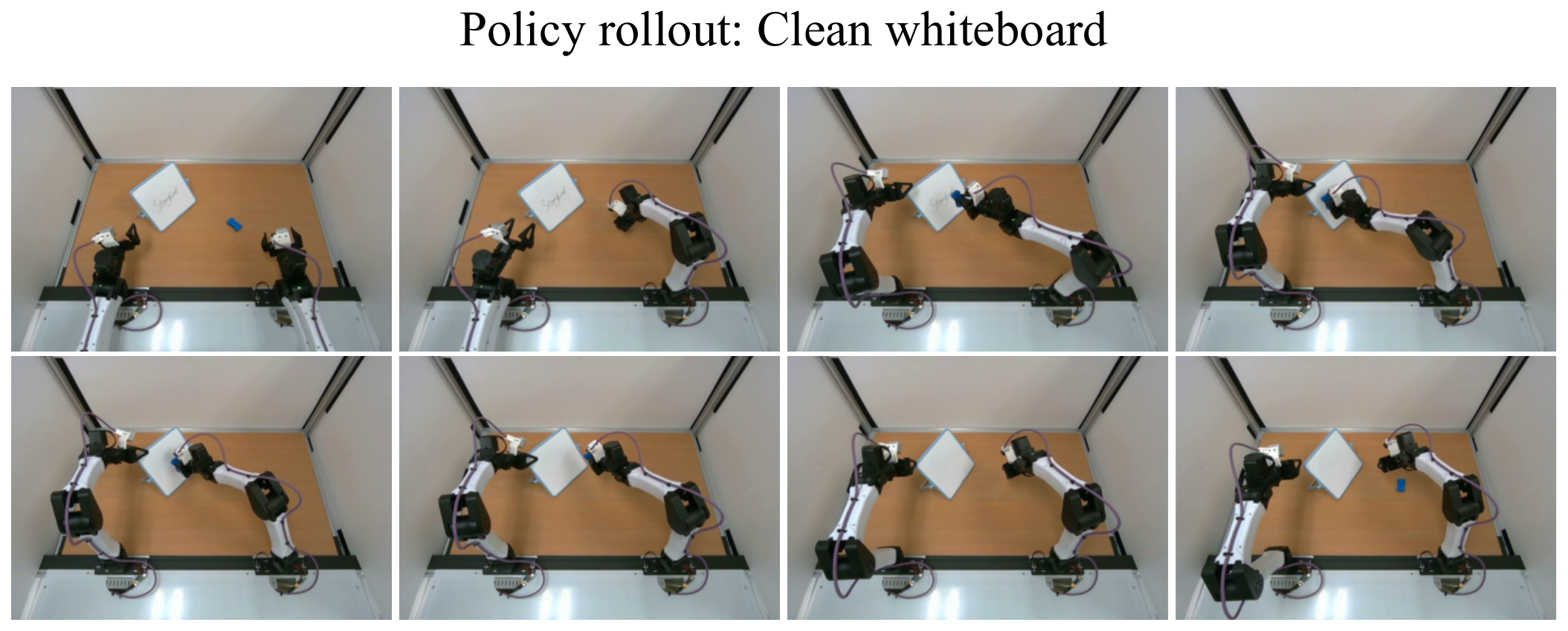}
    \caption{Example policy rollout trajectories for (1) fold shorts (top), (2) clean whiteboard (below).}
    \label{fig:policy_rollout}
\end{figure}

\end{document}